\documentclass[10pt,journal,compsoc]{IEEEtran}
\ifCLASSOPTIONcompsoc
  \usepackage[nocompress]{cite}
\else
  \usepackage{cite}
\fi

\usepackage{graphicx}
\usepackage{amsmath,amssymb} 
\usepackage{eucal}
\usepackage{algpseudocode}
\usepackage{algorithm}
\usepackage{multirow}
\usepackage{amsthm}
\usepackage{mathrsfs}  
\usepackage{array}
\usepackage{makecell}
\usepackage{pifont}
\usepackage[numbers,sort]{natbib}
\usepackage{color}
\usepackage{xcolor}
\usepackage{colortbl}
\usepackage{wrapfig}
\newcommand{\xmark}{\ding{55}}%
\newcommand{\eg}{\textit{e}.\textit{g}.}
\newcommand{\etal}{\textit{et al}.}

\DeclareMathOperator{\E}{\mathbb{E}}
\newtheorem{defn}{Definition}

\usepackage{hyperref}
\hypersetup{colorlinks,linkcolor={red},citecolor={blue},urlcolor={red}}  

\ifCLASSINFOpdf
\else
\fi

\begin{document}
%
\title{Knowledge Distillation and Student-Teacher Learning for Visual Intelligence: A Review and New Outlooks}

\author{Lin Wang,~\IEEEmembership{Student Member,~IEEE,}
        and~Kuk-Jin Yoon,~\IEEEmembership{Member,~IEEE}

\IEEEcompsocitemizethanks{\IEEEcompsocthanksitem L. Wang and K.-J. Yoon are with the Visual Intelligence Lab., Department of Mechanical Engineering, Korea Advanced Institute of Science and Technology, 291 Daehak-ro, Guseong-dong, Yuseong-gu, Daejeon 34141, Republic of Korea. 
E-mail: \{wanglin, kjyoon \}@kaist.ac.kr.}
\thanks{Manuscript received April 19, 2020; revised August 26, 2020.\hfil\break(Corresponding author: Kuk-Jin Yoon)}}

%
%

\markboth{Journal of \LaTeX\ Class Files,~Vol.~14, No.~8, April~2020}%
{Shell \MakeLowercase{\textit{et al.}}: Bare Demo of IEEEtran.cls for Computer Society Journals}
%



\IEEEtitleabstractindextext{%
\begin{abstract}
Deep neural models, in recent years, have been successful in almost every field, even solving the most complex problem statements. However, these models are huge in size with millions (and even billions) of parameters, demanding heavy computation power and failing to be deployed on edge devices. Besides, the performance boost is highly dependent on redundant labeled data. 
To achieve faster speeds and to handle the problems caused by the lack of labeled data, knowledge distillation (KD) has been proposed to transfer information learned from one model to another. KD is often characterized by the so-called `Student-Teacher' (S-T) learning framework and has been broadly applied in model compression and knowledge transfer.
This paper is about KD and S-T learning, which are being actively studied in recent years. First, we aim to provide explanations of what KD is and how/why it works. 
Then, we provide a comprehensive survey on the recent progress of KD methods together with S-T frameworks typically used for vision tasks. In general, we investigate some fundamental questions that have been driving this research area and thoroughly generalize the research progress and technical details. Additionally, we systematically analyze the research status of KD in vision applications. Finally, we discuss the potentials and open challenges of existing methods and prospect the future directions of KD and S-T learning. 
\end{abstract}

}

\maketitle

\IEEEdisplaynontitleabstractindextext

%
\IEEEpeerreviewmaketitle

\IEEEraisesectionheading{\section{Introduction}\label{sec:introduction}}

%
%
%
%

\begin{figure*}[t!]
    \centering
    \includegraphics[width=\textwidth]{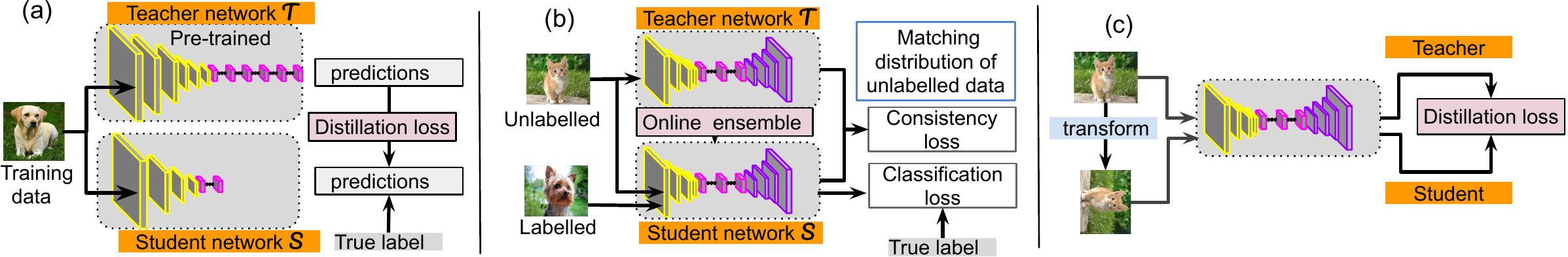}
    \caption{Illustrations of KD methods with S-T frameworks. (a) for model compression and for knowledge transfer, \eg, (b) semi-supervised learning and (c) self-supervised learning. 
    }
    \label{fig:overview_fig}
\end{figure*}

\IEEEPARstart{T}{he} success of deep neural networks (DNNs) generally depends on the elaborate design of DNN architectures. In large-scale machine learning, especially for tasks such as image and speech recognition, most DNN-based models are over-parameterized to extract the most salient features and to ensure generalization. Such cumbersome models are usually very deep and wide, which require a considerable amount of computation for training and are difficult to be operated in real-time. Thus, to achieve faster speeds, many researchers have been trying to utilize the cumbersome models that are trained to obtain lightweight DNN models, which can be deployed in edge devices.  That is, when the cumbersome model has been trained, it can be used to learn a small model that is more suitable for real-time applications or deployment \cite{hinton2015distilling} as depicted in Fig.~\ref{fig:overview_fig}(a). 

On the other hand, the performance of DNNs is also heavily dependent on very large and high-quality labels to training datasets. For such a reason, many endeavours have been taken to reduce the amount of labeled training data without hurting too much the performance of DNNs. A popular approach for handling such a lack of data is to \textit{transfer knowledge} from one source task to facilitate the learning on the target task. One typical example is semi-supervised learning in which a model is trained with only a small set of labeled data and a large set of unlabeled data. 
Since the supervised cost is undefined for the unlabeled examples, it is crucial to apply consistency costs or regularization methods to match the predictions from both labeled and unlabeled data. In this case, knowledge is transferred within the model that assumes a dual role as \textit{teacher} and \textit{student} \cite{tarvainen2017mean}. For the unlabeled data, the student learns as before; however, the teacher generates targets, which are then used by the student for learning.  The common goal of such a learning metric is to form a better teacher model from the student without additional training, as shown in Fig.~\ref{fig:overview_fig}(b). Another typical example is self-supervised learning, where the model is trained with artificial labels constructed by the input transformations (\eg, rotation, flipping, color change, cropping). In such a situation, the knowledge from the input transformations is transferred to supervise the model itself to improve its performance as illustrated in Fig.~\ref{fig:overview_fig}(c). 

This paper is about \textit{knowledge distillation (KD) and student-teacher (S-T) learning}, a topic that has been actively studied in recent years. Generally speaking, KD is widely regarded as a primary mechanism that enables humans to quickly learn new complex concepts when given only small training sets with the same or different categories \cite{gutstein2008knowledge}. In deep learning, KD is an effective technique that has been widely used to transfer information from one network to another network whilst training constructively. KD was first defined by \cite{bucilua2006model} and generalized by Hinton \etal ~\cite{hinton2015distilling}.  KD has been broadly applied to two distinct fields:  model compression (refer to Fig.~\ref{fig:overview_fig}(a)) and knowledge transfer (refer to Fig.~\ref{fig:overview_fig} (b) and (c)). For model compression, a smaller student model is trained to mimic a pretrained larger model or an ensemble of models.  Although various forms of knowledge are defined based on the purpose, one common characteristic of KD is symbolized by its\textit{ S-T framework}, where the model providing knowledge is called the teacher and the model learning the knowledge is called the student. 

In this work, we focus on analyzing and categorizing existing KD methods accompanied by various types of S-T structures for model compression and knowledge transfer. We review and survey this rapidly developing area with particular emphasis on the recent progress.
Although KD has been applied to various fields, such as visual intelligence, speech recognition, natural language processing (NLP), etc.,
this paper mainly focuses on the KD methods in the vision field, as most demonstrations have been done on computer vision tasks. KD methods used in NLP and speech recognition can be conveniently explained using the KD prototypes in vision. 
As the most studied KD methods are for model compression, we systematically discuss the technical details, challenges, and potentials. Meanwhile, we also concentrate on the KD methods for knowledge transfer in semi-supervised learning, self-supervised learning, etc., and we highlight the techniques that take S-T learning as a way of learning metric. 

We explore some fundamental questions that have been driving this research area. Specifically, what is the theoretical principle for KD and S-T learning? What makes one distillation method better than others? Is using multiple teachers better than using one teacher? Do larger models always make better teachers and teach more robust students? Can a student learn knowledge only if a teacher model exists? Is the student able to learn by itself? Is off-line KD always better than online learning? 

With these questions being discussed, we incorporate the potentials of existing KD methods and prospect the future directions of the KD methods together with S-T frameworks. We especially stress the importance of recently developed technologies, such as neural architecture search (NAS), graph neural networks (GNNs), and gating mechanisms for empowering KD. Furthermore, we also emphasize the potential of KD methods for tackling challenging problems in particular vision fields such as 
 360$^\circ$ vision and event-based vision. 

\begin{figure*}[t!]
    \centering
    \includegraphics[width=0.98\textwidth]{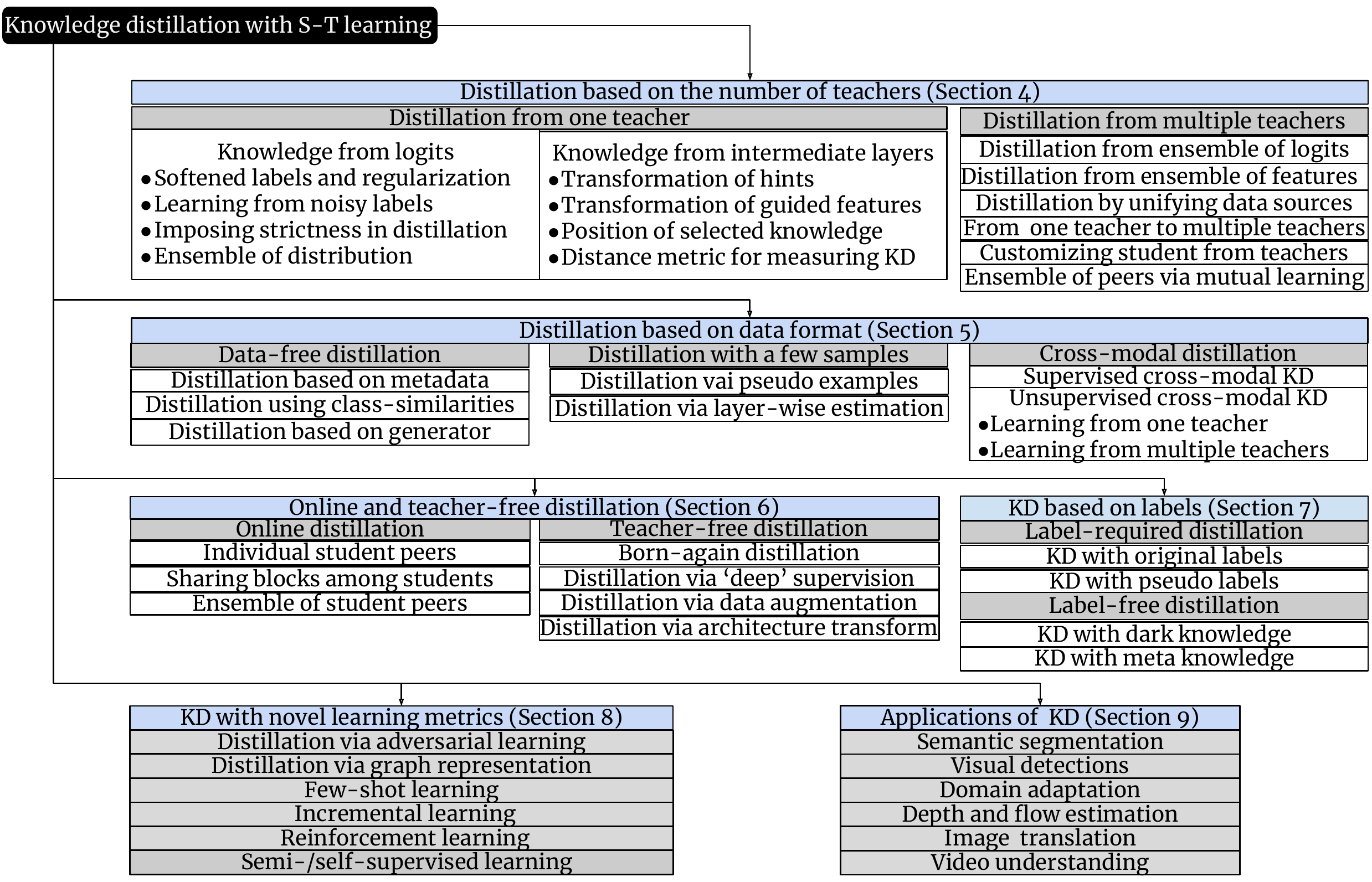}
    \caption{Hierarchically-structured taxonomy of knowledge distillation with S-T learning.}

    \label{fig:taxonomy_KD}
\end{figure*}

The main contributions of this paper are three-fold:
\begin{itemize}
   \item We give a comprehensive overview of KD and S-T learning methods, including problem definition, theoretical analysis, a family of KD methods with deep learning, and vision applications. 
    \item We provide a systematic overview and analysis of recent advances of KD methods and S-T frameworks hierarchically and structurally and offer insights and summaries for the potentials and challenges of each category. 
    \item We discuss the problems and open issues and identify new trends and future direction to provide insightful guidance in this research area.
\end{itemize}
The organization of this paper is as follows. First, we explain why we need to care about KD and S-T learning in Sec.2. Then, we provide a theoretical analysis of KD in Sec.3. Section 3 is followed by Sec.4 to Sec.8, where we categorize the existing methods and analyze their challenges and potential. Fig.~\ref{fig:taxonomy_KD} shows the taxonomy of KD with S-T learning to be covered in this survey in a hierarchically-structured way. In Sec.9, based on the taxonomy, we will discuss the answers to the questions raised in Sec.1. Section 10 will present the future potentials of KD and S-T learning, followed by a conclusion in Sec.11. 

\section{What is KD and Why Concern it? }
\noindent \textbf{What's KD?} Knowledge Distillation (KD) was first proposed by \cite{bucilua2006model} and expanded by \cite{hinton2015distilling}. KD refers to the method that helps the training process of a smaller student network under the supervision of a larger teacher network. Unlike other compression methods, KD can downsize a network regardless of the structural difference between the teacher and the student network. 
 In \cite{hinton2015distilling}, the knowledge is transferred from the teacher model to the student by minimizing the difference between the logits (the inputs to the final softmax) produced by the teacher model and those produced by the student model. 
 
 However, in many situations, the output of softmax function on the teacher's logits has the correct class at a very high probability, with all other class probabilities very close to zero. In such a circumstance, it does not provide much information beyond the ground truth labels already provided in the dataset. To tackle such a problem, \cite{hinton2015distilling,intellabs_github} introduced the concept of 'softmax temperature', which can make the target to be 'soft.' Given the logits $z$ from a network, the class probability $p_i$ of an image is calculated as \cite{intellabs_github}:
\begin{equation}
    p_i = \frac{\exp (\frac{z_i}{\rho})}{\sum_j\exp(\frac{z_i}{\rho})}
    \label{soft_label}
\end{equation}
where $\rho$ is the temperature parameter. When $\rho=1$, we get the standard softmax function. As $\rho$ increases, the probability distribution produced by the softmax function becomes softer, providing more information as to which classes the teacher found more similar to the predicted class. The information provided in the teacher model is called \textit{dark knowledge} \cite{hinton2015distilling}. It is this dark knowledge that 
affects the overall flow of information to be distilled.  When computing the distillation loss, the same $\rho$ used in the teacher is used to compute the logits of the student. 

For the images with ground truth, \cite{hinton2015distilling} stated that it is beneficial to train the student model together with the ground truth labels in addition to the teacher's soft labels. Therefore, we also calculate the 'student loss' between the student's predicted class probabilities and the ground truth labels. The overall loss function, composed of the student loss and the distillation loss, is calculated as:
\begin{equation}
\begin{split}
       \mathcal{L}_{KD} = \alpha * H(y, \sigma(z_s)) + 
       \beta * H(\sigma(z_t;\rho),  \sigma(z_s; \rho)) \\ 
       = \alpha * H(y, \sigma(z_s) + 
       \beta * [KL(\sigma(z_t; \rho), 
       \sigma (z_s,\rho))  
       + H(\sigma(z_t))] 
\end{split}
\label{loss_student}
\end{equation}
where 
$H$ is the loss function, $y$ is the ground truth label, $\sigma$ is the softmax function parameterized by the temperature $\rho$ ($\rho \ne 1$ for distillation loss), and $\alpha$ and $\beta$ are coefficients. $z_s$ and $z_t$ are the logits of the student and teacher respectively. 

\noindent \textbf{Why concern KD? } KD has become a field in itself in the machine learning community, with broad applications to computer vision, speech recognition, NLP, etc. From 2014 to now, many research papers  \cite{frunk_github,dkozlov_github} have been presented in the major conferences, such as CVPR, ICCV, ECCV, NIPS, ICML, ICLR, etc., and the power of KD has been extended to many learning processes (\eg, few-shot learning) except to model compression. The trend in recent years is that KD with S-T frameworks has become a crucial tool for knowledge transfer, along with model compression. The rapid increase in scientific activity on KD has been accompanied and nourished by a remarkable string of empirical successes both in academia and industry. The particular highlights on some representative applications are given in Sec.9, and 
in the following Sec.\ref{theoretical_analysis}, we provide a systematic theoretical analysis.





\section{A theoretical analysis of KD}
\label{theoretical_analysis}
Many KD methods have been proposed with various intuitions. However, there is no commonly agreed theory as to how knowledge is transferred, thus making it difficult to effectively evaluate the empirical results and less actionable to design new methods in a more disciplined way. 
Recently, Ahn \etal~\cite{Ahn_2019_CVPR}, Hegde \etal ~\cite{hegde2019variational} and Tian \etal ~\cite{tian2019contrastive} formulate KD as a maximization of mutual information between the representations of the teacher and the student networks. Note that the representations here can refer to either the logits information or the intermediate features.
From the perspective of representation learning and information theory, the mutual information reflects the joint distribution or mutual dependence between the teacher and the student and quantifies how much information is transferred. We do agree that maximizing the mutual information between the teacher and the student is crucial for learning constructive knowledge from the teacher.  We now give a more detailed explanation regarding this.

Based on Bayes's rule, the mutual information between two paired representations can be defined as:
\begin{equation}
\centering
\begin{split}
      I(T;S) =  H(R(T)) - H(R(T)|R(S))   \\
     = -\mathbb{E}_T[\log p(R(T))]  + \mathbb{E}_{T,S}[\log p((R(T)|R(S))]
\end{split}
\label{mutual_info}
\end{equation}
where $R(T)$ and $R(S)$ are the representations from both the teacher and the student, and $H(\cdot)$ is the entropy function. Intuitively, the mutual information is to increase degree of certainty in the information provided in $R(T)$ when $R(S)$ is known. Therefore, maximizing $\mathbb{E}_{T,S}[\log p((R(T)|R(S))]$ w.r.t. the parameters of the student network $S$ increases the lower bound on mutual information. However, the true distribution of $p((R(T)|R(S))$ is unknown, instead it is desirable to estimate $p((R(T)|R(S))$ by fitting a variations distribution $q((R(T)|R(S))$ to approximate the true distribution $p((R(T)|R(S))$. Then Eqn.~\ref{mutual_info} can be rewritten  as:
\begin{equation}
\centering
\begin{split}
      I(T;S) = H(R(T)) + \mathbb{E}_{T,S}[\log p(R(T)|R(S))] \\
      = H(R(T)) + \mathbb{E}_{T,S}[\log q((R(T)|R(S))] +  \\ \mathbb{E}_S[KL(p(R(T)|R(S))||q(R(T)|R(S))] \\
\end{split}
\label{vid_mutual}
\end{equation}
Assuming there is sufficiently expressive way of modeling $q$, Eqn.~\ref{vid_mutual} can be updated as:
\textbf{\begin{equation}
\centering
\begin{split}
      I(T;S) \ge H(R(T)) + \mathbb{E}_{T,S}[\log q((R(T)|R(S))]
\end{split}
\label{lower_bound}
\end{equation}}
Note that the last term in Eqn.~\ref{vid_mutual} is non-negative since $KL(\cdot)$ function is non-negative and $H(R(T))$ is constant w.r.t the parameters to be optimized. 
By modeling $q$, it is easy to quantify the amount of knowledge being learned by the student.  In general, $q$ can be modeled by a Gaussian distribution, Monte Carlo approximation, or noise contrastive estimation (NCE).  We do believe that theoretically explaining how KD works is connected to representation learning, where the correlations and higher-order output dependencies between the teacher and the student are captured. The critical challenge is increasing the lower bounds of information, which is also pointed out in \cite{tian2019contrastive}.

In summary, we have theoretically analyzed how KD works and mentioned that the \textit{representation} of knowledge is crucial for the transfer of knowledge and learning of the student network. Explicitly dealing with the representation of knowledge from the teacher is significant and challenging, because the knowledge from the teacher expresses a more  general learned information (\eg feature information, logits, data usage, etc.) that is helpful for building up a well-performing  student. In the following sections, we will provide a hierarchically-structured taxonomy for the KD methods regarding how the information is transferred for both teacher and student, how knowledge is measured, and how the teacher is defined.

\section{KD based on the number of teachers}
\subsection{Distillation from one teacher}
\label{one_teacher}
\noindent \textbf{Overall insight:} \textit{Transferring knowledge from a large teacher network to a smaller student network can be achieved using either the logits or feature information from the teacher.}

\subsubsection{Knowledge from logits}
\noindent\textbf{Softened labels and regularization.} 
Hinton \etal ~\cite{hinton2015distilling} and Ba and Caruana ~\cite{ba2014deep} propose to shift the knowledge from teacher network to student network by learning the class distribution via softened softmax (also called 'soft labels') given in Eqn. (\ref{soft_label}). The softened labels are in fact achieved by introducing temperature scaling to increase of small probabilities. These KD methods achieved some surprising results on vision and speech recognition tasks. Recently, Mangalam \etal ~\cite{mangalam2018compressing} introduce a special method based on class re-weighting to compress U-net into a smaller version. Re-weighting of the classes, in fact, softens the label distribution by obstructing inherent class imbalance. Compared to \cite{hinton2015distilling}, some works such as Ding \etal ~\cite{ding2019adaptive}, Hegde \etal~\cite{hegde2019variational}, Tian \etal~\cite{tian2019contrastive}, Cho \etal ~\cite{cho2019building} and Wen  \etal ~\cite{wen2019preparing}, point out that the trade-off (see Eqn.~\ref{loss_student}) between the soft label and the hard label is rarely optimal, and since $\alpha$, $\beta$ and $T$ are fixed during training time, it lacks enough flexibility to cope with the situation without the given softened labels.  Ding \etal ~\cite{ding2019adaptive} instead propose \textit{residual label} and \textit{residual loss} to enable the student to use the erroneous experience during the training phase, preventing over-fitting and improving the performance. Similarly, Tian \etal ~\cite{tian2019contrastive} formulate the teacher's knowledge as structured knowledge and train a student to capture significantly more \textit{mutual information} during contrastive  learning. Hegde \etal ~\cite{hegde2019variational} propose to train a \textit{variational} student by adding sparsity regularizer based on variational inference, similar to the method in \cite{Ahn_2019_CVPR}. The sparsification of the student training reduces over-fitting and improves the accuracy of classification. Wen \etal ~\cite{wen2019preparing} notice that the knowledge from the teacher is useful, but uncertain supervision also influences the result. Therefore, they propose to fix the incorrect predictions (knowledge) of the teacher via smooth regularization and avoid overly uncertain supervision using dynamic temperature.

On the other hand, Cho \etal ~\cite{cho2019efficacy}, Yang \etal ~\cite{yang2019snapshot} and Liu \etal ~\cite{liu2019knowledge} focus on different perspectives of regularization to avoid under-/over-fitting. Cho \etal ~\cite{cho2019efficacy} discover that early-stopped teacher makes a better student especially when the capacity of the teacher is larger than the student's. Stopping the training of the teacher early is akin to regularizing the teacher, and stopping knowledge distillation close to convergence allows the student to fit the training better.  Liu \etal ~\cite{liu2019knowledge} focus on modeling the distribution of the parameters as prior knowledge, which is modeled by aggregating the distribution (logits) space from the teacher network. Then the prior knowledge is penalized by a sparse recording penalty for constraining the student to avoid over-regularization. Mishra  \etal ~\cite{mishra2017apprentice} combine network quantization with model compression by training an apprentice using KD techniques and showed that the performance of low-precision networks could be significantly improved by distilling the logits of the teacher network. Yang \etal ~\cite{yang2019snapshot} propose a snapshot distillation method to perform S-T (similar network architecture) optimization in \textit{one generation}. Their method is based on a cycle learning rate policy (refer to Eqn.~\ref{loss_student} and Eqn.~\ref{loss_student_onegenration}) in which the last snapshot of each cycle (\eg,$W_{T}^{l-1}$ in iteration $l-1$) serves as a teacher in the next cycle (\eg,  $W_{T}^{l}$ in iteration $l$).  Thus, the idea of snapshot distillation is to extract supervision signals in earlier epochs in the same generation to make sure the difference between teacher and student is sufficiently large to avoid under-fitting. The snapshot distillation loss can be described as: 
\begin{equation}
\begin{split}
       \mathcal{L}(x, W_{l-1}) = \alpha * H(y, \sigma(z_s^{l-1} ; \rho=1) + \\
       \beta * H(\sigma(z_t^{l}; \rho=\tau), \sigma (z_t^{l-1}, \rho=\tau)) 
\end{split}
\label{loss_student_onegenration}
\end{equation}
where the $W_{l-1}$ is the weights of student at iteration $l-1$. $z_s^{l-1}$ and $z_t^{l-1}$ represent the logits of student and teacher at iteration $l-1$. \textit{More detailed analysis for the methods with mutual information and one generation will be discussed in Sec.~\ref{online_kd}}. 

\begin{table*}[t!]
\caption{A taxonomy of KD methods using logits. The given equations here are the generalized objective functions, and they may vary in individual work.  }
\small
\begin{center}
\begin{tabular}{c|c|c|c|c}
\hline
 Method & Sub-category & Description & \thead{KD objective \\function}&References   \\
 \hline \hline 
    \multirow{8}{*}{\thead{KD from\\ logits}}& \thead{Softened labels and\\ regularization} & \thead{Distillation using soft labels \\ and add regularizatio to \\avoid under-/over-fitting}  & Eqn.~\ref{loss_student_onegenration}& \thead{\cite{hinton2015distilling, ba2014deep,mangalam2018compressing,hegde2019variational, tian2019contrastive, cho2019building, wen2019preparing} \\ \cite{ hegde2019variational,Ahn_2019_CVPR,wen2019preparing,cho2019efficacy,yang2019snapshot,liu2019knowledge,mishra2017apprentice,mishra2017apprentice}} \\   \cline{2-5}
    & \thead{Learning from \\noisy labels} & \thead{Adding noise \\or using noisy data } & \thead{Eqn.~\ref{loss_student_onegenration} \\or Eqn.~\ref{loss_student_noise}} & \cite{li2017learning, xie2019self, xu2019positive,sarfraz2019noisy,srivastava2014dropout}\\  \cline{2-5}
    & Imposing strictness & \thead{Adding optimization methods \\to  teacher or student} & \thead{ Eqn.~\ref{loss_tsd} or \\ Eqn.~\ref{loss_student_onegenration}} & \cite{yang2019training,yu2019learning, arora2019knowledge, park2019relational, peng2019correlation,furlanello2018born, wang2016relational}\\  \cline{2-5}
    & Ensemble of distribution  & \thead{Estimating model or\\ data uncertainty}  & Eqn.~\ref{loss_student_ensemble} & \cite{mirzadeh2019improved, cho2019efficacy, malinin2019ensemble, zhang2018deep,phuong2019distillation}\\ \hline 
\end{tabular}
\end{center}
\label{table:logit_comp}
\end{table*}

\noindent\textbf{Learning from noisy labels} \cite{li2017learning,xie2019self,xu2019positive,sarfraz2019noisy} propose methods that utilize the similar knowledge (softened labels) as in \cite{hinton2015distilling} but focus on data issue. Specifically, \cite{li2017learning} assume that there is a small clean dataset ${D}_c$ and a large noisy dataset ${D}_n$, while \cite{xie2019self} and \cite{xu2019positive} use both labeled and unlabeled data to improve the performance of student. In \cite{li2017learning}, the aim of distillation is to use the large amount of noisy data ${D}_n$ to augment the small clean dataset ${D}_c$ to learn a better visual representation and classifier. That is, the knowledge is distilled from the small clean dataset ${D}_c$ to facilitate a better model from the entire noisy dataset ${D}_n$. The method is essentially different from \cite{hinton2015distilling} focusing on inferior model instead of inferior dataset. The same loss function in Eqn. ~\ref{loss_student} is used, except $z_t= \sigma[f_{{D}_c}(x)]$, where $f_{{D}_c}$ is an auxiliary model trained from the clean dataset ${D}_c$. Furthermore, a risk function on the unreliable label $\bar{y}$ is defined as $R_{ \bar{y}} = \E_{{D}_t}[||\bar{y}- y^*||]^2$, where $y^*$ is the unknown ground truth label and ${D}_t$ is the unseen test dataset. $R_{ \bar{y}}$ is an indicator that measures the level of noise in the distillation process.

Xu \etal~\cite{xu2019positive} probes a positive-unlabeled classifier for addressing the problem of requesting the entire original training data, which can not be easily uploaded to the cloud. \cite{xie2019self} trains a \textit{noisy} student by next three steps: 1) train a teacher model on labeled data, 2) use the teacher to generate pseudo labels on unlabeled images, and 3) train a student model on the combination of labeled images and pseudo labeled images while injecting noise (adversarial perturbation) to the student for better generalization and robustness. This way, the student generalizes better than the teacher. 
Similarly, \cite{sarfraz2019noisy} study adversarial perturbation and consider it as a crucial element in improving both the generalization and the robustness of the student. Based on how humans learn, two learning theories for the S-T model are proposed: fickle teacher and soft randomization. The fickle teacher model is to transfer the teacher's uncertainty to the student using Dropout \cite{srivastava2014dropout} in the teacher model. The soft randomization method is to improve the adversarial robustness of student model by adding Gaussian noise in the knowledge distillation. In this setting, the original distillation objective for the student in Eqn.~\ref{loss_student} can be updated as: 
\begin{equation}
\begin{split}
       \mathcal{L}(x+\delta, W) = \alpha * {H}(y, \sigma(z_s ; \rho=1) + \\
       \beta * {H}(\sigma(z_t; \rho=\tau), \sigma (z_s, \rho=\tau)) 
\end{split}
\label{loss_student_noise}
\end{equation}
where $\delta$ is the variation of adversarial perturbation.  It is shown that using the teacher model trained on clean images to train the student model with adversarial perturbation can retain the adversarial robustness  and mitigate the loss in generalization. 

\noindent\textbf{Imposing strictness in distillation.} In contrast, Yang \etal ~\cite{yang2019training}, Yu \etal ~\cite{yu2019learning}, Arora \etal ~\cite{arora2019knowledge}, RKD \cite{park2019relational} and Peng \etal ~\cite{peng2019correlation} shift to a new perspective focusing more on putting \textit{strictness} to the distillation process via optimization (\eg, distribution embedding, mutual relations, etc). In particular, \cite{yang2019training} initiates to put strictness on the teacher while \cite{yu2019learning} proposes two teaching metrics to impose strictness on the student. Yang \etal~ observe that, except learning \textit{primary class} (namely, the ground truth), learning \textit{secondary class} ( high confidence scores in the \textit{dark knowledge} in \cite{hinton2015distilling}) may help to alleviate the risk of the student over-fitting.  They thus introduce a framework of optimizing neural networks in \textit{generations} (namely, iterations), which requires training a patriarchal model ${M}^0$  only supervised by the dataset. After $m$ generations, the student ${M}^m$ is trained by $m$-th generation with the supervision of a teacher ${M}^{m-1}$. Since the secondary information is crucial for training a robust teacher, a fixed integer $K$ standing for the semantically similar class is chosen for each image, and the gap between the confidence scores of the primary class and other $K-1$ classes with highest scores is computed, This can  be described as:
\begin{equation}
\begin{split}
       \mathcal{L}(x, W^T) = \alpha * {H}(y, \sigma(z_t ; \rho=1) + \\
       \beta * [f_{a_1}^T - \frac{1}{K-1} \sum_{k=2}^{K}f_{a_k}^T] 
\end{split}
\label{loss_tsd}
\end{equation}
where $f_{a_{k}}$ indicates the $k$-th largest elements of the output (logits) $z_t$.  
Note that this S-T optimization is similar to BAN \cite{furlanello2018born}; however, the goal here is to help the student learn inter-class similarity and prevent over-fitting.  Different from the teacher in \cite{furlanello2018born}, the teacher here is deeper and larger than the student.
\cite{yu2019learning} extends \cite{hinton2015distilling} for metric learning by using embedding networks to project the information (logits) learned from images to the embedding space. The embeddings are typically used to perform distance computation between the data pairs of a teacher and a student. From this point of view, the knowledge computed based on the embedding network is the actual knowledge as it represents the data distribution.  They design two different teachers: absolute teacher and relative teacher. For the absolute teacher, the aim is to minimize the distance between the teacher and student embeddings while the aim for the relative teacher is to enforce the student to learn any embedding as long as it results in a similar distance between the data points. They also explore hints \cite{hinton2015distilling} and attention \cite{zagoruyko2016paying} to strengthen the distillation of embedding networks. \textit{We will give more explicit explanations of these two techniques in Sec.~\ref{sec_feamap_dis}.}  

\cite{arora2019knowledge} proposes an embedding module that captures interactions between query and document information for question answering. The embedding of the output representation (logits) includes a simple attention model with a query encoder, a prober history encoder, a responder history encoder, and a document encoder. The attention model minimizes the summation of cross-entropy loss and KL-divergence loss, inspired by \cite{hinton2015distilling}. 
On the other hand, \cite{wang2016relational} and RKD \cite{park2019relational} consider another type of strictness, namely the \textit{mutual relation or relation knowledge} of the two examples in the learned representations for both the teacher and the student. This approach is very similar to the relative teacher in  ~\cite{yu2019learning} since both aim to measure the distance between the teacher's and the student's embeddings. However, RKD \cite{park2019relational} also considers the angle-wise relational measure, similar to persevering secondary information in ~\cite{yang2019training}. 

\noindent\textbf{Ensemble of distribution.}
Although various methods have been proposed to extract knowledge from logits, some works \cite{mirzadeh2019improved, cho2019efficacy, malinin2019ensemble, zhang2018deep} show that KD is not always practical due to knowledge uncertainty. The performance of the student degrades when the gap between the student and the teacher is large. \cite{malinin2019ensemble} points out that estimating the model's uncertainty is crucial since it ensures a more reliable knowledge to be transferred. They stress on the ensemble approaches to estimate the data uncertainty and the distributional uncertainty. To estimate the distributional uncertainty, an ensemble distribution distillation approach anneals the temperature of the softmax to not only capture the mean of ensemble soft labels but also the diversity of the distribution. Meanwhile,~\cite{phuong2019distillation} proposes a similar approach of matching the distribution of distillation-based multi-exit architectures, in which a sequence of feature layers is augmented with early exits at different depths. By doing so, the loss defined in Eqn.~\ref{loss_student} becomes:
\begin{equation}
\begin{split}
       \mathcal{L}(x, W) = \frac{1}{K}\sum_{k=1}^K[ \alpha * {H}(y, \sigma(p_s^k ; \rho=1) + \\
       \beta *{H}(\sigma(p_t^k; \rho=\tau), \sigma (p_s^k, \rho=\tau))] 
\end{split}
\label{loss_student_ensemble}
\end{equation}
where $K$ indicates the total number of exits, and $p_s^k$ and $p_t^k$ represent the $ k$-th probabilistic output at exit $k$. 

Conversely, \cite{you2017learning, zhang2018deep, papernot2016semi,sau2016deep, arora2019knowledge,tarvainen2017mean, furlanello2018born, lan2018knowledge, song2018collaborative, radosavovic2018data, tan2019multilingual, Vongkulbhisal_2019_CVPR,wu2019distilled, dvornik2019diversity, yang2019model, park2019feed, lee2019stochasticity, chen2019online, mirzadeh2019improved} propose to add more teachers or other auxiliaries, such as teaching assistant and small students, to improve the robustness of ensemble distribution. We will explicitly analyze these approaches in the following Sec.~\ref{multi_teach}. 
\begin{figure}[t!]
    \centering
    \includegraphics[width=\columnwidth]{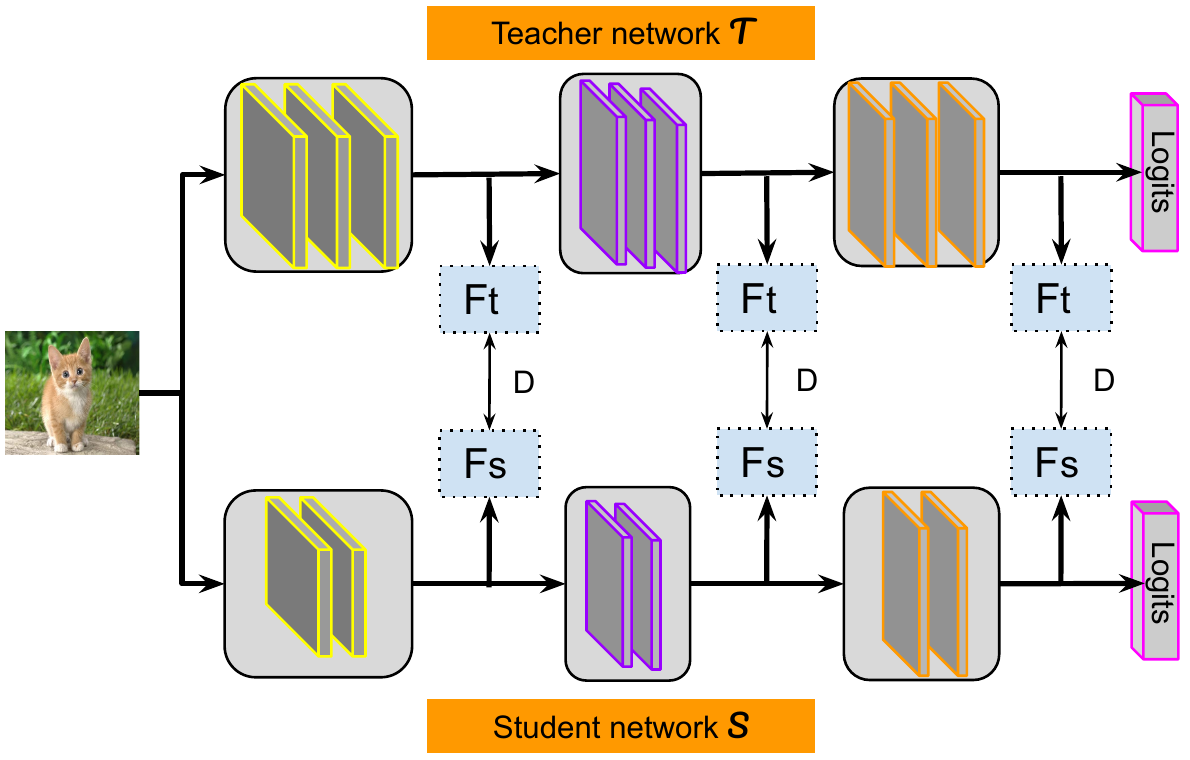}
    \caption{An illustration of general feature-based distillation.  
    }
    \label{fig:fea_dis_fig}
\end{figure}

\noindent\textbf{Summary.} Table.~\ref{table:logit_comp} summarizes the KD methods that use logits or `soft labels'. We divide these methods into four categories. In overall, distillation using logits needs to transfer the dark knowledge to avoid over-/under-fitting. Meanwhile, the gap of model capacity between the teacher and the student is also very crucial for effective distillation. Moreover, the drawbacks of learning from logits are obvious. First, the effectiveness of distillation is limited to softmax loss and relies on the number of classes. Second, it is impossible to apply these methods to the KD problems in which there are no labels (\eg, low-level vision). 

\noindent \textbf{Open challenges:} The original idea in \cite{hinton2015distilling} is in its apparent generality: any student can learn from any teacher; however, it is shown that this promise of generality is hard to be achieved on some datasets \cite{zagoruyko2016paying, cho2019efficacy} (\eg, ImageNet \cite{deng2009imagenet}) even when regularization or strictness techniques are applied. When the capacity of the student is too low, it is hard for the student to incorporate the logits information of the teacher successfully. Therefore, it is expected to improve the generality and provide a better representation of logits information, which can be easily absorbed by the student.  

\subsubsection{Knowledge from the intermediate layers}
\textbf{Overall insight:} \textit{Feature-based distillation enables learning richer information from the teacher and provides more flexibility for performance improvement. }

\label{sec_feamap_dis}
Apart from distilling knowledge from the softened labels, Romero \etal ~\cite{romero2014fitnets} initially introduce \textit{hint} learning rooted from \cite{hinton2015distilling}. A hint is defined as the outputs of a teacher's hidden layer, which helps guide the student's learning process. The goal of student learning is to learn a feature representation that is the optimal prediction of the teacher's intermediate representations. Essentially, the function of hints is a form of regularization; therefore, a pair of hint and guided (a hidden layer of the student) layer has to be carefully chosen such that the student is not over-regularized. 
Inspired by \cite{romero2014fitnets}, many endeavours have been taken to study the methods to choose, transport and match the hint layer (or layers) and the guided layer (or layers) via various layer transform (\eg, transformer \cite{heo2019comprehensive, kim2018paraphrasing}) and distance (\eg, MMD \cite{huang2017like}) metrics. Generally, the hint learning objective can be written as:
\begin{equation}
    \mathcal{L}(F_T, F_S) = D({TF}_t(F_T), {TF}_s(F_S))
\label{fea_dis_loss}
\end{equation}
Where $F_T$ and $F_S$ are the selected hint and guided layers  of teacher and student. ${TF}_t$ and ${TF}_s$ are the transformer and regressor functions for the hint layer of teacher and guided layer of student. $D(\cdot)$ is the distance function(\eg, $l_2$) measuring the similarity of the hint and the guided layers.

Fig.~\ref{fig:fea_dis_fig} depicts the general paradigm of feature-based distillation. It is shown that various intermediate feature representations can be extracted from different positions and are transformed with a certain type of regressor or transformer. The similarity of the transformed representations is finally optimized via an arbitrary distance metrics $D$ (\eg, $L_1$ or $L_2$ distance). In this paper, we carefully scrutinize various design considerations of feature-based KD methods and summarize four key factors that are often considered: \textit{transformation of the hint, transform of the guided layer, position of the selected distillation feature, and distance metric} \cite{heo2019comprehensive}. In the following parts, we will analyze and categorize all existing feature-based KD methods concerning these four aspects.
\begin{table*}[t!]
\caption{A taxonomy of  knowledge distillation from the intermediate layers (feature maps). KP incidates knowledge projection. 
}
\small
\begin{center}
\begin{tabular}{c|c|c|c|c|c}
\hline
 Method & Teacher's ${TF}_t$ & Student's ${TF}_s$& Distillation position & Distance metric & Lost knowledge   \\
\hline\hline
FitNet \cite{romero2014fitnets}  &None& $1 \times 1$ Conv & Middle layer & $L_1$ & None  \\ \hline
AT \cite{zagoruyko2016paying}  &Attention map& Attention map & End of layer group&  $L_2$ & Channel dims  \\ \hline
KP \cite{Zhang2017KnowledgePF}  & Projection matrix & Projection matrix  &Middle layers & $L_1$ +  KP loss &  Spatial dims\\ \hline
FSP \cite{yim2017gift} &FSP matrix & FSP matrix  &End of layer group & $L_2$ & Spatial dims \\ \hline
FT \cite{kim2018paraphrasing}  &Encoder-decoder& Encoder-decoder & End of layer group & $L_1$ & Channel + Spatial dims  \\ \hline
AT \cite{zagoruyko2016paying}  &Attention map& Attention map & End of layer group&  $L_2$ & Channel dimensions  \\ \hline
MINILM \cite{wang2020minilm}  &Self-ttention & Self-attention & End of layer group&  KL & Channel dimensions  \\ \hline
Jacobian \cite{srinivas2018knowledge} & Gradient penalty & Gradient penalty & End of layer group &   $L_2$  &  Channel dims \\ \hline
SVD \cite{yim2017gift} & Truncated SVD  & Truncated SVD  & End of layer group & $L_2$ & Spatial dims \\ \hline 
VID \cite{Ahn_2019_CVPR} &None& $1 \times 1$ Conv  & Middle layers & $KL$ & None \\ \hline 
IRG \cite{liu2019knowledge} &Instance graph & Instance graph  & Middle layers & $L_2$ & Spatial dims \\ \hline 
RCO \cite{jin2019knowledge} &None  & None  & Teacher's train route & $L_2$ & None\\ \hline 
SP \cite{tung2019similarity} &Similarity matrix  & Similarity matrix  & Middle layer & Frobenius norm & None\\ \hline 
MEAL \cite{shen2019meal} &Adaptive pooling  & Adaptive pooling  &End of layer group & $L_{1/2}$/KL/$L_{GAN}$ & None\\ \hline 
Heo \cite{shen2019meal} &Margin ReLU  & $1 \times 1$ Conv & Pre-ReLU & Partial $L_2$ & Negative features \\ \hline 
AB \cite{heo2018knowledge} &Binarization & $1 \times 1$ Conv & Pre-ReLU & Margin $L_2$ & feature values \\ \hline 
Chung \cite{chung2020featuremaplevel} &None & None & End of layer & $L_{GAN}$  & None \\ \hline 
Wang \cite{wang2019distilling} & None & Adaptation layer & Middle layer & Margin $L_1$ & Channel + Spatial dims \\ \hline
KSANC \cite{changyong2019knowledge} & Average pooling & Average pooling & Middle layers & $L_2$ + $L_{GAN}$ & Spatial dims \\ \hline
Kulkarni \cite{kulkarni2019stagewise} & None & None &End of layer group & $L_2$ & None \\ \hline
IR \cite{aguilar2019knowledge} & Attention matrix & Attention matrix &Middle layers & KL+ Cosine & None \\ \hline
Liu \cite{liu2019knowledge} & Transform matrix & Transform matrix &Middle layers & KL & Spatial dims \\ \hline 
NST \cite{huang2017like} & None & None & Intermediate layers & MMD & None \\ \hline
Gao \cite{gao2020residual} & None & None & Intermediate layers & $L_2$ & None \\
\hline
\end{tabular}
\end{center}
\label{table:fea_comp}
\end{table*}

\noindent\textbf{Transformation of hints}
As pointed in \cite{Ahn_2019_CVPR}, the knowledge of teacher should be easy to learn as the student. To do this, teacher's hidden features are usually converted by a transformation function $T_t$. Note that the transformation of teacher's knowledge is a very crucial step for feature-based KD since there is a risk of losing information in the process of transformation. The transformation methods of teacher's knowledge in AT \cite{zagoruyko2016paying}, MINILM \cite{wang2020minilm}, FSP \cite{yim2017gift}, ASL\cite{li2019layer}, Jacobian \cite{srinivas2018knowledge}, KP \cite{Zhang2017KnowledgePF}, SVD \cite{lee2018self}, SP \cite{tung2019similarity}, MEAL \cite{shen2019meal}, KSANC \cite{changyong2019knowledge}, and NST \cite{huang2017like} cause the knowledge to be missing due to the reduction of feature dimension. Specifically, AT \cite{kim2018paraphrasing} and MINILM \cite{wang2020minilm} focus on attention mechanisms (\eg, self-attention \cite{vaswani2017attention}) via an attention transformer $T_t$ to transform the activation tensor $F \in \mathbb{R}^{C \times H \times W}$ to $C$ feature maps $F \in \mathbb{R}^{H \times W}$. FSP \cite{yim2017gift} and ASL \cite{li2019layer} calculate the information flow of the distillation based on the Gramian matrices, through which the tensor $F \in \mathbb{R}^{C \times H \times W}$ is transformed to  $G \in \mathbb{R}^{C \times N}$, where $N$ represents the number of matrices. Jacobian \cite{srinivas2018knowledge} and SVD \cite{lee2018self} map the tensor $F \in \mathbb{R}^{C \times H \times W}$  to  $G \in \mathbb{R}^{C \times N}$ based on Jacobians using first-order Taylor series and truncated SVD, respectively, inducing information loss. KP \cite{Zhang2017KnowledgePF} projects $F \in \mathbb{R}^{C \times H \times W}$ to $M$ feature maps $F \in \mathbb{R}^{M \times H \times W}$, causing loss of knowledge. Similarly, SP \cite{tung2019similarity} proposes a similarity-preserving knowledge distillation method based on the observation that semantically similar inputs tend to elicit similar activation patterns. To achieve this goal, the teacher's feature $F \in \mathbb{R}^{B \times C \times H \times W}$ is transformed to $G \in \mathbb{R}^{B \times B}$, where $B$ is the batch size. The $G$ encodes the similarity of the activations at the teacher layer, but leads to an information loss during the transformation.  MEAL \cite{shen2019meal} and KSANC \cite{changyong2019knowledge} both use \textit{pooling} to align the intermediate map of the teacher and student, leading to an information loss when transforming the teacher's knowledge. NST \cite{huang2017like} and PKT \cite{passalis2018learning} match the distributions of neuron selectivity patterns and the affinity of data samples between the teacher and the student networks. The loss functions are based on minimizing the maximum mean discrepancy (MMD) and Kullback-Leibler (KL) divergence between these distributions respectively, thus causing information loss when selecting neurons. 

On the other hand, FT \cite{kim2018paraphrasing} proposes to extract good \textit{factors} through which transportable features are made. The transformer ${TF}_t$ is called the \textit{paraphraser} and the transformer  ${TF}_s$ is called the \textit{translator}. To extract the teacher factors, an adequately trained paraphraser is needed. Meanwhile, to enable the student to assimilate and digest the knowledge according to its own capacity, a user-defined paraphrase ratio is used in the paraphraser to control the factor of the transfer. Heo \etal ~\cite{heo2018knowledge} use the original teacher's feature in the form of binarized values, namely via a separating hyperplane (activation boundary (AB)) that determines whether neurons are activated  or deactivated. Since AB only considers the activation of neurons and not the magnitude of neuron response, there is information loss in the feature binarization process. Similar information loss happens in IRG \cite{liu2019knowledge}, where the teacher's feature space is transformed to a graph representation with vertices and edges where the relationship matrices are calculated. IR \cite{aguilar2019knowledge} distills the internal representations of the teacher model to the student model. However, since multiple layers in the teacher model are compressed into one layer of the student model, there is information loss when matching the features.  Heo \etal ~\cite{heo2019comprehensive} design ${TF}_t$ with a margin ReLU function to exclude the negative (adverse) information and to allow positive (beneficial) information. The margin $m$ is determined based on batch normalization \cite{ioffe2015batch} after $1 \times 1$ convolution in the student's transformer ${TF}_s$.  

Conversely, FitNet \cite{romero2014fitnets}, RCO \cite{jin2019knowledge},  Chung \etal ~\cite{chung2020featuremaplevel}, Wang \etal ~\cite{wang2019distilling}, Gao \etal ~\cite{gao2020residual} and Kulkarni \etal ~\cite{kulkarni2019stagewise} do not add additional transformation to the teacher's knowledge; this leads to no information loss from teacher's side. However, not all knowledge from the teacher is beneficial for the student. As pointed by \cite{heo2019comprehensive}, features include both adverse and beneficial information. For effective distillation,  it is important to impede the use of adverse information and to avoid missing the beneficial information. 

\noindent\textbf{Transformation of the guided features}
The transformation ${TF}_s$ of the guided features (namely, student transform) of the student is also an important step for effective KD. Interestingly, the SOTA works such as AT \cite{zagoruyko2016paying}, MINILM \cite{wang2020minilm}, FSP \cite{yim2017gift}, Jacobian \cite{srinivas2018knowledge}, FT \cite{kim2018paraphrasing}, SVD \cite{lee2018self}, SP \cite{tung2019similarity}, KP \cite{ Zhang2017KnowledgePF}, IRG \cite{liu2019knowledge}, RCO \cite{jin2019knowledge},MEAL \cite{shen2019meal},  KSANC \cite{changyong2019knowledge}, NST \cite{huang2017like}, Kulkarni \etal ~\cite{kulkarni2019stagewise}, Gao \etal ~\cite{gao2020residual} and Aguilar \etal ~\cite{aguilar2019knowledge} use the same ${TF}_s$ as the ${TF}_t$, which means the same amount of information might be lost in both transformations of the teacher and the student.  

Different from the transformation of teacher, FitNet \cite{romero2014fitnets}, AB \cite{heo2018knowledge}, Heo \etal ~\cite{heo2019comprehensive}, and VID \cite{Ahn_2019_CVPR} change the dimension of teacher's feature representations and design ${TF}_s$ with a `bottleneck' layer ($1\times1$ convolution) to make the student's features match the dimension of the teacher's features. Note that Heo \etal ~\cite{heo2019comprehensive} add a batch normalization layer after a $1\times1$ convolution to calculate the margin of the proposed margin ReLU transformer of the teacher. There are some advantages of using $1 \times 1$ convolution in KD. First, it offers a channel-wise pooling without a reduction of the spatial dimensionality. Second, it can be used to create a one-to-one linear projection of the stack of feature maps. Lastly, the projection created by $1 \times 1$ convolution can also be used to directly increase the number of feature maps in the distillation model. In such a case, the feature representation of student does not decrease but rather increase to match the teacher's representation; this does not cause information loss in the transformation of the student.

Exceptionally, some works focus on a different aspect of the transformation of student's feature representations. Wang \etal ~\cite{wang2019distilling} make the student imitate the fine-grained local feature regions close to object instances of the teacher's representations. This is achieved by designing a particular adaptation function ${TF}_s$ to fulfill the imitation task. IR \cite{aguilar2019knowledge} aims to let the student acquire the abstraction in a hidden layer of the teacher by matching the internal representations. That is, the student is taught to know how to compress the knowledge from multiple layers of the teacher into a single layer. In such a setting, the transformation of the student's guided layer is done by a self-attention transformer.  Chung \etal ~\cite{chung2020featuremaplevel}, on the other hand, propose to impose no transformation to both student and teacher, but rather add a discriminator to distinguish the feature map distributions of different networks (teacher or student). 

\noindent\textbf{Distillation positions of features}
In addition to the transformation of teacher's and student's features, distillation position of the selected features is also very crucial in many cases. Earlier, FitNet \cite{romero2014fitnets}, AB \cite{heo2018knowledge}, and Wang \etal ~\cite{wang2019distilling} use the end of an arbitrary middle layer as the distillation point. However, this method is shown to have poor distillation performance. Based on the definition of layer group \cite{zagoruyko2016wide}, in which a group of layers have same spatial size, AT \cite{zagoruyko2016paying}, FSP \cite{yim2017gift}, Jacobian \cite{srinivas2018knowledge}, MEAL \cite{shen2019meal}, KSANC \cite{changyong2019knowledge}, Gao \etal~\cite{gao2020residual} and Kulkarni \etal~\cite{kulkarni2019stagewise} define the distillation position at the end of each layer group, in contrast to  FT \cite{kim2018paraphrasing} and NST \cite{huang2017like} where the position lies only at the end of last layer group. Compared to FitNet, FT achieves better results since it focuses more on informational knowledge. IRG \cite{liu2019knowledge} considers all the above-mentioned critical positions; namely, the distillation position lies not only in the end of earlier layer group but also in the end of the last layer group. Interestingly, VID \cite{Ahn_2019_CVPR}, RCO \cite{jin2019knowledge}, Chung \etal ~\cite{chung2020featuremaplevel}, SP \cite{tung2019similarity}, IR \cite{aguilar2019knowledge}, and Liu \etal ~\cite{liu2019knowledge} generalize the selection of distillation positions by employing variational information maximization \cite{barber2003algorithm}, curriculum learning \cite{bengio2009curriculum}, adversarial learning \cite{goodfellow2014generative}, similarity-presentation in representation learning \cite{horn2016learning}, muti-task learning \cite{clark2019bam}, and reinforcement learning \cite{mnih2016asynchronous}. We will discuss more for these methods in later sections.

\noindent\textbf{Distance metric for measuring distillation}
The quality of KD from teacher to student is usually measured by various distance metrics. The most commonly used distance function is based on $L_1$ or $L_2$ distance. FitNet \cite{romero2014fitnets}, AT \cite{zagoruyko2016paying}, NST \cite{huang2017like}, FSP \cite{yim2017gift}, SVD \cite{lee2018self}, RCO \cite{jin2019knowledge}, FT \cite{kim2018paraphrasing}, KSANC \cite{changyong2019knowledge}, Gao \etal~\cite{gao2020residual} and Kulkarni \etal~\cite{kulkarni2019stagewise} are mainly based on $L_2$ distance,  whereas MEAL \cite{shen2019meal} and Wang \etal ~\cite{wang2019distilling} mainly use $L_1$ distance.
On the other hand, Liu \etal~\cite{liu2019knowledge} and IR~\cite{aguilar2019knowledge} utilize KL-divergence loss to measure feature similarities. Furthermore, a cosine-similarity loss is adopted by IR \cite{aguilar2019knowledge} and RKD \cite{park2019relational} to regularize the context representation on the feature distributions of teacher and student. 

Some works also resort to the adversarial loss for measuring the quality of KD. MEAL \cite{shen2019meal} shows that the student learning the distilled knowledge with discriminators is better optimized than the original model, and the student can learn distilled knowledge from a teacher model with arbitrary structures. Among the works focusing on feature-based distillation, 
KSANC \cite{jin2019knowledge} adds an adversarial loss at the last layer of both teacher and student networks, while MEAL adds multi-stage discriminators in the position of every extracted feature representation. It is worth mentioning that using adversarial loss has shown considerable potential in improving the performance of KD. We will explicitly discuss the existing KD techniques based on adversarial learning in the following Sec.~\ref{kd_gan}.

\noindent\textbf{Potentials and open challenges}
Table.~\ref{table:fea_comp} summarizes the existing feature-based KD methods. It is shown that most works employ feature transformations for both the teacher and the student. $L1$ or $L2$ loss is the most commonly used loss for measuring KD quality. 
A natural question one may ask is what's wrong with directly matching the features of teacher and student?
If we consider the activation of each spatial position as a feature, the flattened activation map of each filter is a sample of the space of selected neurons with dimension $HW$, which reflects how DNN learns an image \cite{huang2017like}. Thus, when matching distribution, it is less desirable to directly match the samples since the sample density can be lost in the space, as pointed in \cite{romero2014fitnets}. Although \cite{gao2020residual} proposes to distill knowledge by directly matching feature maps, a teaching assistant is introduced to learn the residual errors of between the feature maps of the student and teacher. This approach better mitigates the performance gap between the teacher and student, thus improving generalization capability. 

\noindent \textbf{Potentials:} Feature-based methods show more generalization capabilities and quite promising results. In the next research, more flexible ways of determining the representative knowledge of features are expected. The approaches used in representation learning (\eg, parameter estimation, graph models) might be reasonable solutions for these problems. Additionally, neural architecture search (NAS) techniques may better handle the selection of features. Furthermore, feature-based KD methods are possible for use in cross-domain transfer and low-level vision problems.

\noindent \textbf{Open challenges:} 
Although we have discussed most existing feature-based methods, it is still hard to say which one is best. First, it is difficult to measure the different aspects in which information is lost. Additionally, most works randomly choose intermediate features as knowledge, and yet do not provide a reason as to why they can be the representative knowledge among all layers.  Third, the distillation position of features is manually selected based on the network or the task. Lastly, multiple features may not represent better knowledge than the feature of a single layer. Therefore, better ways to choose knowledge from layers and to represent knowledge could be explored.

\begin{figure*}[t!]
    \centering

    \includegraphics[width=\textwidth]{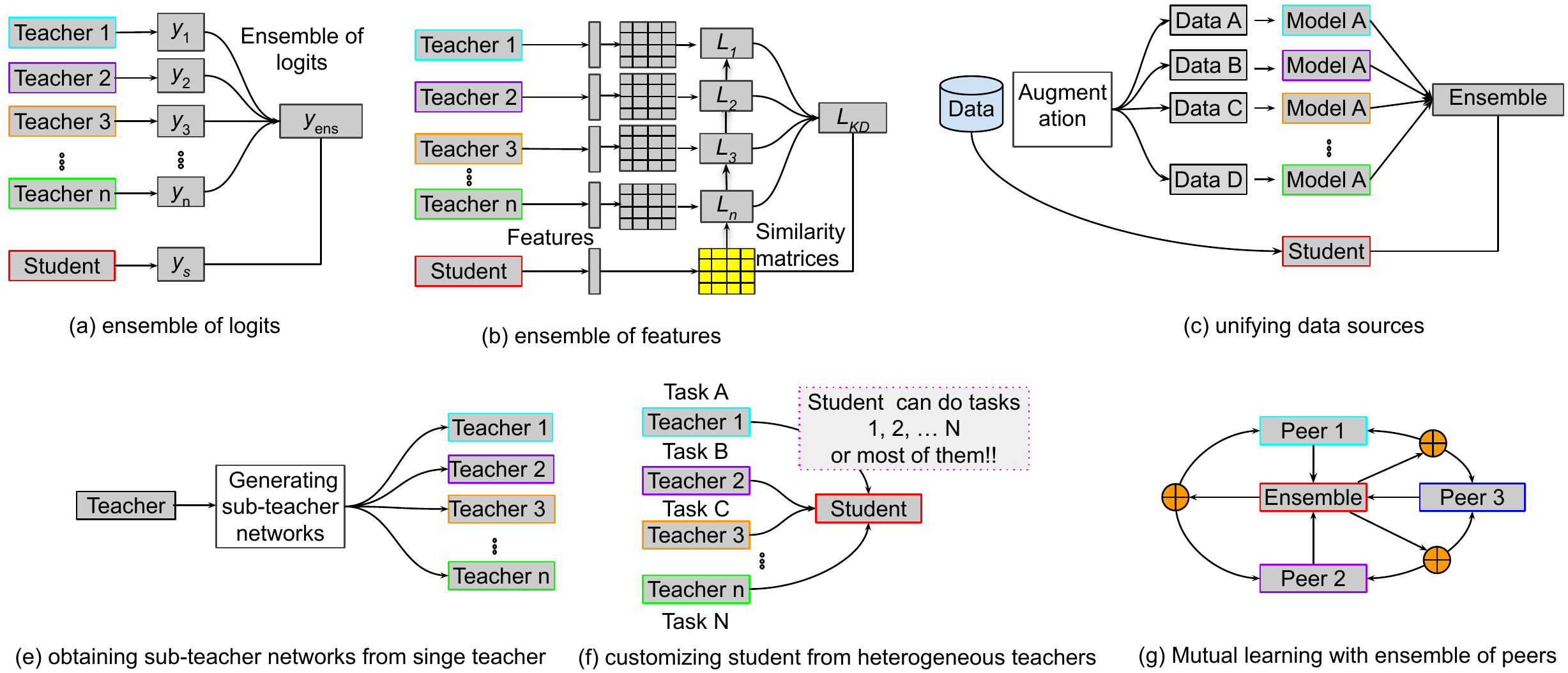}
    \vspace{-15pt}
    \caption{Graphical illustration for KD with multiple teachers. The KD methods can be categorized into six types: (a) KD from the ensemble of logits, (b) KD from the ensemble of feature representations via some similarity matrices, (c) unifying various data sources from the same network (teacher) model A to generate various teacher models, (d) obtaining hierarchical or stochastic sub-teacher networks given one teacher network; (e) training a versatile student network from multiple heterogeneous teachers, (f) online KD from diverse peers via ensemble of logits.}
    \label{fig:multiple_teachers}
    \vspace{-5pt}
\end{figure*}

\subsection{Distillation from multiple teachers}
\label{multi_teach}
\textbf{Overall insight:} \textit{The student can learn better knowledge from multiple teachers, which are more informative and instructive than a single teacher.}

While impressive progress has been achieved under
the common S-T KD paradigm, where knowledge is transferred from one high-capacity teacher network to a student network. The knowledge capacity in this setting is quite limited \cite{park2019feed}, and knowledge diversity is scarce for some special cases, such as cross-model KD \cite{zhang2018better}. To this end, some works probe learning a portable student from \textit{multiple} teachers or an ensemble of teachers. The intuition behind this can be explained in analogy with the cognitive process of human learning. In practice, a student does not solely learn from a single teacher but learn a concept of knowledge better provided with instructive guidance from multiple teachers on the same task or heterogeneous teachers on different tasks. In such a way, the student can amalgamate and assimilate various illustrations of knowledge representations from multiple teacher networks, and build a comprehensive knowledge system \cite{you2017learning, shen2019customizing, sau2016deep}. As a result, many new KD methods \cite{you2017learning, papernot2016semi, ruder2017knowledge, sau2016deep, tarvainen2017mean, furlanello2018born, zhang2018deep, zhu2018knowledge, song2018collaborative, radosavovic2018data,tan2019multilingual, vongkulbhisal2019unifying, wu2019distilled, dvornik2019diversity, yang2020model, park2019feed, lee2019stochasticity, tran2020hydra, ruiz2020distilled, wu2020multi, zhang2018better, fukuda2017efficient, mehak2018knowledge, jung2019distilling, sun2019patient, lan2018knowledge, liu2019attentive, liu2019improving, mirzadeh2019improved, he2018multi, liu2019knowledge,shen2019amalgamating, ye2019student, luo2019knowledge, shen2019customizing, anil2018large, zhou2019m2kd,xiang2020learning, gao2020residual} have been proposed.
Although these works vary in various distillation scenarios and assumptions, they share some standard characteristics that can be categorized into five types: ensemble of logits, ensemble of feature-level information, unifying data sources, and obtaining sub-teacher networks from a single teacher network, customizing student network from heterogeneous teachers and learning a student network with diverse peers, via the ensemble of logits. We now explicitly analyze each category and provide insights on how and why they are valuable for the problems.

 \subsubsection{Distillation from the ensemble of logits} Model ensemble of logits is one of the popular methods in KD from multiple teachers as shown in Fig.~\ref{fig:multiple_teachers}(a). 
In such a setting,
the student is encouraged to learn
the softened output of the assembled teachers' logits (dark knowledge)
via the cross-entropy loss as done in \cite{you2017learning, wu2020multi, dvornik2019diversity, anil2018large, tran2020hydra, furlanello2018born, jung2019distilling, lan2018knowledge, liu2019attentive, liu2019improving, mehak2018knowledge, zhou2019m2kd, mirzadeh2019improved, papernot2016semi, tan2019multilingual, tarvainen2017mean, yang2020model}, which can be generalized into: 
 \begin{equation}
 \label{ensemble_logits}
     \mathcal{L}_{Ens}^{logits}= {H}(\frac{1}{m}\sum_{i}^{m}N_{T_i}^{\tau}(x), N_S^{\tau}(x))
 \end{equation}
where $m$ is the total number of teachers, ${H}$ is the cross-entropy loss, $N_{T_i}^{\tau}$ and $N_{S}^{\tau}$ are the $i$-th teacher's and $i$-th student's logits (or softmax ouputs), and $\tau$ is the temperature. The averaged softened output serves as the incorporation of multiple teacher networks in the output layer. Minimizing Eqn.~\ref{ensemble_logits} achieves the goal of KD at this layer. Note that the averaged softened output is more objective than that of any of the individuals, because it can mitigate the unexpected bias of the softened output existing in some of the input data.

Unlike the methods as mentioned above, \cite{ruder2017knowledge,lan2018knowledge,xiang2020learning, zhang2018better,fukuda2017efficient} argue that taking the average of individual prediction may ignore the diversity and importance variety of the member teachers of an ensemble. Thus, they propose to learn the student model by imitating the summation of the teacher's predictions with a gating component. Then, Eqn.~\ref{ensemble_logits} becomes:

 \begin{equation}
 \label{ensemble_logits_gating}
     \mathcal{L}_{Ens}^{logits}= {H}(\sum_{i}^{m}g_iN_{T_i}^{\tau}(x), N_S^{\tau}(x))
 \end{equation}
where $g_i$ is the gating parameter. In \cite{ruder2017knowledge}, the $g_i$ is the normalized similarity $sim(D_{S_i}, D_T)$ of the source domain $D_{S}$ and target domain $D_T$.

\textbf{Summary:} Distilling knowledge from the ensemble of logits mainly depends on taking the average or the summation of individual teacher’s logits. Taking the average alleviates the unexpected bias, but it may ignore the diversity of individual teachers of an ensemble. The summation of the logits of each teacher can be balanced by the gating parameter $g_i$, but ways to determine better values of $g_i$ is an issue worth studying in further works.

\subsubsection{Distillation from the ensemble of features}
Distillation from the ensemble of feature representations is more flexible and advantageous than from the ensemble of logits, since it can provide more affluent and diverse cross-information to the student.
However, 
distillation from the ensemble of features \cite{park2019feed, wu2019distilled, liu2019knowledge, mehak2018knowledge, zhou2019m2kd, sun2019patient, zhang2018better} is more challenging, since each teacher's feature representation at specific layers is different from the others. Hence, transforming the features, and forming an ensemble of the teachers' feature-map-level representations becomes the key problem, as illustrated in Fig.~\ref{fig:multiple_teachers}(b). 

To address this issue, Park \etal ~\cite{park2019feed} proposed feeding the student’s feature map into some nonlinear layers (called transformers). The output is then trained to mimic the final feature map of the teacher network. In this way, the advantages of the general model ensemble and feature-based KD methods, as mentioned in Sec.~\ref{sec_feamap_dis}, can both be incorporated. The loss function is given by:

\begin{equation}
 \label{ensemble_feature_feed}
     \mathcal{L}_{Ens}^{fea}= \sum_{i}^{m}||\frac{x_{T_i}}{||x_{T_i}||_2} - \frac{{TF}_{i}(x_S)}{||{TF}_{i}(x_S)||_2}||_1
 \end{equation}
 where $x_{T_i}$ and $x_S$ are $i$-th teacher's and $i$-th student's feature maps respectively, and $TF$ is the transformer (\eg, $3\times3$ convolution layer) used for adapting the student’s feature with that of the teacher. 

 In contrast, Wu \etal ~\cite{wu2019distilled} and Liu \etal ~\cite{liu2019knowledge} proposed letting the student model to imitate the learnable transformation matrices of the teacher models. This approach is an updated version of a single teacher model \cite{tung2019similarity}. For the $i$-th teacher and student network in \cite{wu2019distilled}, the similarity between the feature maps is computed based on the Euclidean metric as:
 \begin{equation}
  \mathcal{L}_{Ens}^{fea}= \sum_{i}^{m}\alpha_i|| \log(A_S) - log (A_{T_i})||_F^2
 \end{equation}
where $\alpha_i$ is the teacher weight for controlling the contribution of the $i$-th teacher, and $\alpha_i$ should satisfy $\sum_i^{m}\alpha_i=1$. $A_S$ and $A_{T_i}$ are the similarity matrices of the student and the $i$-th teacher, respectively. These can be computed by $A_S=x_S^{\intercal} x_S$ and $A_T = x_{T_i}^{\intercal} x_{T_i}$, respectively.

\noindent\textbf{Open challenges:} Based on our review, it is evident that only a few studies propose distilling knowledge from the ensemble of feature representations. Although \cite{park2019feed, wu2020multi} proposed to let the student directly mimic the ensemble of feature maps of the teachers via either non-linear transformation or similarity matrices with weighting mechanisms, there still exist some challenges. First, how can we know which teacher’s feature representation is more reliable or more influential among the ensemble? Second, how can we determine the weighting parameter $\alpha_i$ for each student in an adaptive way? Third, instead of summing all feature information together, is there any mechanism of selecting the best feature map of one teacher from the ensemble as the representative knowledge?

\subsubsection{Distillation by unifying data sources }
Although the above mentioned KD methods using multiple teachers are good in some aspects, they assume that the target classes of all teacher and student models are the same. In addition, the dataset used for training is often scarce, and teacher models with high capacity are limited. To tackle these problems, some recent works \cite{vongkulbhisal2019unifying,wu2019distilled,gong2018teaching,radosavovic2018data, sau2016deep,xiang2020learning} propose data distillation by unifying data sources from multiple teachers, as illustrated in Fig.~\ref{fig:multiple_teachers}(c). The goal of these methods is to generate labels for the unlabeled data via various data processing approaches (\eg, data augmentation) to train a student model.

Vongkulbhisal \etal ~\cite{vongkulbhisal2019unifying} proposed to unify an ensemble of\textit{ heterogeneous classifiers} (teachers) which may be trained to classify different sets of target classes and  can share the same network architecture. To generalize distillation, a probabilistic relationship connecting the outputs of the heterogeneous classifiers with that of the unified (ensemble) classifier is proposed. Similarly, 
Wu \etal ~\cite{wu2019distilled} and Gong \etal ~\cite{gong2018teaching} also explored transferring knowledge from teacher models trained in existing data to a student model by using unlabeled data to form a decision function.
Besides, some works utilize the potential of data augmentation approaches to build multiple teacher models from a trained teacher model.
Radosavovic \etal ~\cite{radosavovic2018data} proposed a distillation method via \textit{multiple transformations} on the unlabeled data to build diverse teacher models sharing the same network structure. The technique consists of four steps. First, a single teacher model is trained on manually labeled data. Second, the trained teacher model is applied to multiple transformations of the unlabeled data. Third, the predictions on the unlabeled data are converted into an ensemble of numerous predictions. Fourth, the student model is trained on the union of the manually labeled data and the automatically labeled data. Sau \etal ~\cite{sau2016deep} proposed an approach to simulate the effect of multiple teachers by injecting noise to the training data, and perturbing the logit outputs of a teacher. In such a way, the perturbed outputs not only simulate the setting of multiple teachers, but also generate noise in the softmax layer, thus regularizing the distillation loss. 

\textbf{Summary:} Unifying data sources using data augmentation techniques and unlabeled data from a single teacher model to build up multiple sub-teacher models is also valid for training a student model. However, it requires a high-capacity teacher with more generalized target classes, which could confine the application of these techniques. In addition, the effectiveness of these techniques for some low-level vision problems should be studied further based on feature representations.

\subsubsection{From a single teacher to multiple sub-teachers}
It has been shown that students could be further improved with multiple teachers used as ensembles or used separately. However, using multiple teacher networks is resource heavy, and delays the training process. Following this, some methods \cite{he2018multi, you2017learning, wu2020multi, ruder2017knowledge, lee2019stochasticity, song2018collaborative, tran2020hydra} have been proposed to generate multiple sub-teachers from a single teacher network, as shown in Fig.~\ref{fig:multiple_teachers}(d). Lee \etal ~\cite{lee2019stochasticity} proposed stochastic blocks and skip connections to teacher networks, so that the effect of multiple teachers can be obtained in the same resource from a single teacher network. The sub-teacher networks have reliable performance, because there exists a valid path for each batch. By doing so, the student can be trained with multiple teachers throughout the training phase. Similarly, Ruiz \etal ~\cite{ruiz2020distilled} introduced hierarchical neural ensemble by employing a \textit{binary-tree} structure to share a subset of intermediate layers between different models. This scheme allows controlling the inference cost on the fly, and deciding how many branches need to be evaluated. Tran \etal ~\cite{tran2020hydra}, Song \etal ~\cite{song2018collaborative} and He \etal ~\cite{ he2018multi} introduced \textit{multi-headed architectures} to build multiple teacher networks, while amortizing the computation through a shared heavy-body network. Each head is assigned to an ensemble member, and tries to mimic the \textit{individual predictions} of the ensemble member. 

\noindent\textbf{Open challenges:} Although network ensembles using stochastic or deterministic methods can achieve the effect of multiple teachers and online KD, many uncertainties remain. Firstly, it is unclear how many teachers are sufficient for online distillation. Secondly, which structure is optimal among the ensemble of sub-teachers is unclear? Thirdly, balancing the training efficiency and accuracy of the student network is an open issue. These challenges are worth exploring in further studies.

\subsubsection{Customizing student form heterogeneous teachers}
\label{customize_student}
In many cases, well-trained deep networks (teachers) are focused on different tasks, and are optimized for different datasets. However, most studies focus on training a student by distilling knowledge from teacher networks on the same task or on the same dataset. To tackle these problems, \textit{knowledge amalgamation} has been initialized by recent works \cite{ye2019student, shen2019customizing, shen2019amalgamating, ye2019amalgamating, gao2017knowledge, rusu2015policy, dvornik2019diversity, liu2019knowledge, luo2019knowledge, zhou2019m2kd} to learn a versatile student model by distilling knowledge from the expertise of all teachers, as illustrated in Fig.~\ref{fig:multiple_teachers}(e). Shen \etal ~\cite{shen2019customizing}, Ye \etal ~\cite{ye2019student}, Luo \etal ~\cite{luo2019knowledge} and Ye \etal ~\cite{ye2019amalgamating} proposed training a student network by customizing the tasks without accessing human-labeled annotations. These methods rely on schemes such as branch-out \cite{ahmed2016network} or selective learning \cite{galvan2001selective}. The merits of these methods lie in their ability to reuse deep networks pre-trained on various datasets of diverse tasks to build a tailored student model based on the user demand. The student inherits most of the capabilities of heterogeneous teachers, and thus can perform multiple tasks simultaneously. Shen \etal ~\cite{shen2019amalgamating} and Gao \etal ~\cite{gao2017knowledge} utilized a similar methodology, but focused on same task classification, with two teachers specialized in different classification problems. In this method, the student is capable of handling comprehensive or fine-grained classification. Dvornik \etal ~\cite{dvornik2019diversity} attempted to learn a student that can predict unseen classes by distilling knowledge from teachers via few-shot learning. Rusu \etal ~\cite{rusu2015policy} proposed a multi-teacher single-student policy distillation method that can distill multiple policies of reinforcement learning agents to a single student network for sequential prediction tasks. 

\noindent\textbf{Open challenges:} Studies such as the ones mentioned above have shown considerable potential in customizing versatile student networks for various tasks. However, there are some limitations in such methods. Firstly, the student may not be compact due to the presence of branch-out structures. Secondly, current techniques mostly require teachers to share similar network structures (\eg, encoder–decoder), which confines the generalization of such methods. Thirdly, training might be complicated because some works adopt a dual-stage strategy, followed by multiple steps with fine-tuning. These challenges open scopes for future investigation on knowledge amalgamation.

\begin{table*}[t!]
\caption{A taxonomy of KD with multiple teachers. 
$L_{CE}$ is for cross-entropy loss, $L_{Ens}$ is for the KD loss between the ensemble teacher and the student, $L_{KD}$ indicates the KD loss between individual teacher and the student, $L_{KD_{fea +logits}}$ means KD loss using feature and logits, KL indicates the KL divergence loss for mutual learning, $L_{GAN}$ is for adversarial loss, MMD means mean maximum discrepancy loss, $L_{reg}$ is the regression loss, N/A means not available. \textit{Note that the losses summarized are generalized terms, which may vary in individual works}. }

\vspace{-15pt}
\small
\begin{center}
\begin{tabular}{c|c|c|c|c|c|c|c|c}
\hline
 Method & \thead{ Ensemble \\Logits} & \thead{Ensemble of \\Features} & \thead{Unifying \\data sources} & \thead{Customize \\student} & \thead{Extending\\ teacher} & \thead{Online \\KD} & \thead{Mutual \\learning} & \thead{Major Loss \\ functions} \\
\hline\hline
Anil \cite{anil2018large} &  \checkmark & \xmark & \xmark & \xmark & \xmark & \checkmark & \checkmark& $L_{CE}$+$L_{Ens}$ \\ \hline
Chen \cite{chen2019online} &\checkmark& \checkmark& \xmark & \xmark & \xmark & \checkmark & \checkmark&  $L_{CE}$+$L_{Ens}$+$L_{KD}$ \\ \hline
Dvornik \cite{dvornik2019diversity}  &\checkmark& \xmark& \xmark & \checkmark & \xmark & \checkmark & \checkmark&  $L_{CE}$+$L_{Ens}$+$L_{KD}$\\ \hline
Fukuda \cite{fukuda2017efficient}  & \checkmark& \xmark & \xmark  &\xmark & \xmark & \xmark & \xmark& $L_{CE}$+$L_{Ens}$ \\ \hline
Furlanello \cite{furlanello2018born} &\checkmark & \xmark  & \xmark & \xmark & \xmark & \xmark &  \xmark & $L_{CE}$+$L_{Ens}$ \\ \hline
He \cite{he2018multi} &\xmark &\checkmark & \xmark & \xmark & \checkmark & \xmark &\xmark & $L_1$+$L_{KD}$ \\ \hline
Jung \cite{jung2019distilling}  &\checkmark & \xmark & \xmark&  \xmark & \xmark & \xmark & \xmark & $L_{CE}$+$L_{KD}$ \\ \hline
Lan \cite{lan2018knowledge} & \checkmark & \xmark & \xmark&  \xmark  & \xmark & \checkmark & \xmark & $L_{CE}$+$L_{Ens}$ \\ \hline
Lee \cite{lee2019stochasticity} & \checkmark & \checkmark  & \xmark & \xmark & \checkmark & \checkmark & \checkmark & N/A \\ \hline 
Liu \cite{liu2019knowledge} &\xmark& \checkmark  & \xmark & \checkmark & \xmark & \checkmark & \xmark & $L_{CE}$+$L_{KD}$ \\ \hline 
Luo \cite{luo2019knowledge} &\checkmark & \checkmark  & \xmark &\checkmark& \xmark & \xmark & \xmark & $L_{CE}$+$L_{KD_{fea +logits}}$\\ \hline 
Zhou \cite{zhou2019m2kd} & \checkmark & \checkmark &\xmark & \checkmark & \xmark & \xmark & \xmark &  MMD+$L_{KD}$ \\ \hline
Mirzadeh \cite{mirzadeh2019improved} & \checkmark & \xmark &\xmark & \xmark & \xmark & \xmark & \xmark & $L_{CE}$+$L_{KD}$\\ \hline 
Papernot \cite{papernot2016semi} &\checkmark & \xmark & \checkmark & \xmark & \xmark & \checkmark & \xmark & $L_{KD}$  \\ \hline
Park \cite{park2019feed} &\xmark  & \checkmark & \xmark & \xmark & \xmark & \xmark & \xmark & $L_{CE}$+$L_{KD}$ \\ \hline 
Radosavovic \cite{radosavovic2018data} &\checkmark & \xmark & \checkmark & \xmark & \xmark & \xmark & \xmark & N/A \\ \hline 
Ruder \cite{ruder2017knowledge} &\checkmark  & \xmark & \xmark & \xmark & \xmark & \xmark & \xmark & $L_{CE}$+$L_{KD}$ \\ \hline 
Ruiz \cite{ruiz2020distilled} &\checkmark  & \xmark & \xmark & \xmark & \checkmark & \xmark & \xmark & $L_{CE}$+$L_{Ens}$ \\ \hline 
Sau \cite{sau2016deep} &\checkmark  & \xmark & \checkmark & \xmark & \xmark &\xmark & \xmark & $L_2$ (KD)  \\ \hline 
Shen \cite{shen2019amalgamating} &\checkmark  & \checkmark & \xmark & \checkmark & \xmark & \xmark & \xmark & $L_{KD}$+$L_{PL}$ \\ \hline 
Shen \cite{shen2019customizing} &\checkmark & \checkmark& \xmark & \checkmark & \xmark & \xmark & \xmark &  $L_{KD_{fea +logits}}$+$L_{reg}$\\ \hline 
Song \cite{song2018collaborative} & \checkmark & \xmark & \xmark & \xmark & \checkmark & \checkmark & \xmark & $L_{CE}$+$L_{KD}$\\\hline
Tarvaninen \cite{tarvainen2017mean} &\checkmark  & \xmark & \xmark & \xmark & \xmark & \checkmark & \xmark & $L_{KD}$ \\ \hline 
Tran \cite{tran2020hydra} &\checkmark  & \xmark & \xmark & \xmark & \checkmark & \xmark & \checkmark &  $L_{CE}$+$L_{Ens}$+KL  \\ \hline
Vongkulbhisal \cite{vongkulbhisal2019unifying} &\checkmark  & \xmark & \checkmark & \xmark & \xmark & \xmark & \xmark & $L_{CE}$+$L_{Ens}$ \\ \hline
Wu \cite{wu2019distilled} &\xmark& \checkmark & \checkmark & \xmark & \xmark & \xmark & \xmark & $L_{KD}$  \\ \hline
Wu \cite{wu2020multi} &\checkmark  & \xmark & \xmark & \xmark & \xmark & \xmark& \xmark & $L_{CE}$+$L_{Ens}$ \\ \hline 
Yang \cite{yang2020model} &\checkmark  & \xmark & \xmark& \xmark & \xmark & \xmark & \xmark & $L_{CE}$+$L_{KD}$ \\ \hline 
Ye \cite{ye2019student} &\checkmark & \checkmark  & \xmark  & \checkmark  & \xmark & \xmark & \xmark & $L_{KD}$+$L_{KD}$  \\ \hline 
You \cite{you2017learning} &\checkmark  & \checkmark & \xmark & \xmark & \xmark & \xmark & \xmark &  $L_{CE}$+$L_{KD_{fea +logits}}$\\ \hline 
Zhang \cite{zhang2018better} &\checkmark & \checkmark & \xmark & \xmark & \xmark & \xmark & \xmark & $L_{CE}$+$L_{KD_{fea +logits}}$\\ \hline 
Zhang \cite{zhang2018deep} &\checkmark  & \xmark & \xmark & \xmark & \xmark & \checkmark & \checkmark & $L_{CE}$+KL \\ \hline 
Zhu \cite{zhu2018knowledge} &\checkmark  & \xmark & \xmark & \xmark & \xmark & \checkmark & \checkmark & $L_{CE}$+KL+$L_{Ens}$ \\ \hline 
Chung \cite{chung2020featuremaplevel} & \checkmark & \checkmark & \xmark &\xmark & \xmark & \checkmark & \checkmark & $L_{CE}$+$L_{GAN}$+KL \\  \hline
Kim \cite{kim2019feature} & \checkmark & \checkmark & \xmark & \xmark & \xmark & \checkmark & \checkmark &$L_{CE}$+$L_{Ens}$+KL \\ \hline 
Hou \cite{hou2017dualnet} & \xmark & \checkmark & \xmark & \xmark & \xmark & \checkmark & \xmark & $L_{CE}$+$L_{Ens}$ \\ \hline
Xiang \cite{xiang2020learning} & \checkmark &\xmark & \checkmark & \xmark & \xmark & \xmark & \xmark & $L_{CE}$+$L_{KD}$ \\
\hline
\end{tabular}
\end{center}
\vspace{-15pt}
\label{table:kd_multipleTeachs}
\end{table*}

\subsubsection{Mutual learning with ensemble of peers}
One problem with conventional KD methods using multiple teachers is their computation cost and complexity, because they require pre-trained high-capacity teachers with two-stage (also called offline) learning. To simplify the distillation process, one-stage (online) KD methods \cite{anil2018large, zhang2018deep, lan2018knowledge, chen2019online, chung2020featuremaplevel, kim2019feature, hou2017dualnet, zhu2018knowledge, zhang2018better} have been developed, 
as shown in Fig.~\ref{fig:multiple_teachers}(f). Instead of pre-training a static teacher model, these methods train a set of student models simultaneously by making them learn from each other in a peer-teaching manner. There are some benefits of such methods. First, these approaches merge the training processes of teachers and student models, and use peer networks to provide teaching knowledge. Second, these online distilling strategies can improve the performance of models of any capacity, leading to generic applications. Third, such a peer-distillation method can sometimes outperform teacher-based two-stage KD methods. For KD with mutual learning, the distillation loss of \textit{two peers} is based on the KL divergence, which can be formulated as:
\begin{equation}
    \mathcal{L}_{Peer}^{KD}= KL(z_1, z_2) + KL(z_2, z_1)
\end{equation}
where $KL$ is the KL divergence function, and $z_1$ and $z_2$ are the predictions of peer one and peer two, respectively.

In addition, Lan \etal ~\cite{lan2018knowledge} and Chen \etal ~\cite{chen2019online} also constructed a multi-branch variant of a given target (student) network by adding auxiliary branches to create a local ensemble teacher (also called a group leader) model from all branches. Each branch was trained with a distillation loss that aligns the prediction of that branch with the teacher’s prediction. Mathematically, the distillation loss can be formulated by minimizing the KL divergence of $z_e$ (prediction of the ensemble teacher) and prediction $z_i$ of the $i$-th branch peer:
\begin{equation}
    \mathcal{L}_{Ens}^{KD}= \sum_{i=1}^mKL(z_e, z_i)
\end{equation}
where the prediction $z_e= \sum_{i=1}^m g_iz_i$. $g_i$ is the weighting score or attention-based weights \cite{chen2019online} of the $i$-th branch peer $z_i$. 

Although most of these methods only consider using logit information, some works also exploit feature information. Chung \etal ~\cite{chung2020featuremaplevel} proposed a feature-map-level distillation by employing adversarial learning (discriminators). Kim \etal ~\cite{kim2019feature} introduce a feature fusion module to form an ensemble teacher. However, the fusion is based on the concatenation of the features (output channels) from the branch peers. Moreover, Liu \etal ~\cite{liu2019knowledge} presented a knowledge flow framework which moves the knowledge from the features of multiple teacher networks to a student.

\textbf{Summary:} Compared to two-stage KD methods using pre-trained teachers, distillation from student peers has many merits. The methods are based on mutual learning of peers, and sometimes on ensembles of peers. Most studies rely on logit information; however, some works also exploit feature information via adversarial learning or feature fusion. There is room for improvement in this direction. For instance, the number of peers most optimal for KD processing is worth investigating. In addition, the possibility of using both the online and offline methods simultaneously when the teacher is available is intriguing. Reducing the computation cost without sacrificing accuracy and generalization is also an open issue. We will discuss the advantages and disadvantages of online and offline KD in Sec.~\ref{online_kd}.

\noindent\textbf{Potentials}
Table.~\ref{table:kd_multipleTeachs} summarizes the KD methods with multiple instructors. Overall, most methods rely on the ensemble of logits. However, the knowledge of feature representations has not been taken into account much. Therefore, it is possible to exploit the knowledge of the ensemble of feature representations by designing better gating mechanisms. Unifying data sources and extending teacher models are two effective methods for reducing individual teacher models; however, their performances are degraded. Thus, overcoming this issue needs more research. Customizing a versatile student is a valuable idea, but existing methods are limited by network structures, diversity, and computation costs, which must be improved in future works.


\begin{table*}[t!]
\caption{A taxonomy of data-free knowledge distillation. 
}
\vspace{-15pt}
\small
\begin{center}
\begin{tabular}{c|c|c|c|c|c|c}
\hline
 Method & \thead{Original \\ data needed} & \thead{ Metadata or \\prior info.} & \thead{Number of\\ generators} & Inputs & Discriminator & \thead{Multi-task \\ distillation}   \\
\hline\hline
Lopes \cite{lopes2017data} & \checkmark & \thead{Activations \\of all layers} & \xmark &Image shape &\xmark &\xmark\\ \hline
Bhardwaj \cite{bhardwaj2019dream} & \checkmark & \thead{Activations of \\ pooling layer}& \xmark & Image shape & \xmark & \xmark  \\ \hline
Haroush \cite{haroush2019knowledge} & \checkmark & \thead{Batch \\normalization layer} & \xmark & Image shape & \xmark & \xmark\\ \hline
Nayak \cite{nayak2019zero} & \xmark & Class similarities & \xmark  & \thead{Class label+  Number of DIs}  & \xmark & \xmark \\  \hline
Chen \cite{chen2019data} &  \xmark &\xmark& One &  Noise & Teacher & \xmark  \\ \hline
Fang \cite{fang2019data}  &\xmark& \xmark& One& Noise/images & Teacher + student &\xmark \\ \hline
Ye \cite{ye2020datafree}  & \xmark &\xmark & Three & Noise & Teachers  & \checkmark \\ \hline
Yoo \cite{yoo2019knowledge} &\xmark &\xmark  & One & Noise + class labels & Teacher & \xmark \\ \hline
Yin \cite{yin2019dreaming}  &\xmark& \xmark& One & Noise & Teacher & \xmark \\ \hline
Micaelli \cite{micaelli2019zero}  & \xmark & \xmark & One & Noise & Teacher & \xmark \\ \hline
\end{tabular}
\end{center}
\vspace{-10pt}
\label{table:datafree_comp}
\end{table*}

\section{Distillation based on data format}
\subsection{Data-free distillation}
\label{data_free}
\noindent\textbf{Overall insight:} \textit{ Can we achieve KD when the original data for the teacher or (un)labeled data for training student are not available?} 

One major limitation of most KD methods such as \cite{hinton2015distilling, park2019relational, park2019feed, romero2014fitnets} is that they assume the training samples of the original networks (teachers) or of target networks (students) to be available. However, the training dataset is sometimes unknown in real-world applications owing to privacy and transmission concerns \cite{lopes2017data}. To handle this problem, some representative data-free KD paradigms \cite{lopes2017data, nayak2019zero, ye2020datafree, chen2019data, fang2019data, bhardwaj2019dream, yin2019dreaming, haroush2019knowledge, yoo2019knowledge,micaelli2019zero, kulkarni2017knowledge} are newly developed. A taxonomy of these methods are summarized in Table.~\ref{table:datafree_comp}, and detailed technical analysis is provided as follows.

\subsubsection{Distillation based on metadata}
To the best of our knowledge,
Lopes \etal ~\cite{lopes2017data} initially proposed to reconstruct the original training dataset using only teacher model and it is \textit{metadata} recorded in the form of \textit{precomputed activation statistics}. Thus, the objective is to find the set of images whose representation best matches the one given by the teacher network. Gaussian noise is randomly passed as input to the teacher, and the gradient descent (GD) is made to minimize the difference between the metadata and the representations of noise input. To better constrain the reconstruction, the metadata of \textit{all} layers of the teacher model are used and recorded to train the student model with high accuracy. Bhardwaj \etal ~\cite{bhardwaj2019dream} demonstrated that metadata from a \textit{single} layer (average-pooling layer) using $k$-means clustering is sufficient to achieve high student accuracy. In contrast to \cite{lopes2017data, bhardwaj2019dream} requiring sampling the activations generated by real data, Haroush \etal ~\cite{haroush2019knowledge} proposed using metadata (\eg, channel-wise mean and standard deviation) from Batch Normalization (BN) \cite{ioffe2015batch} layer with synthetic samples. 
The objective of metadata-based distillation can be formulated as:
\begin{equation}
    X^* = \arg\min_{X\sim R ^{H\times W}} L(\Phi(X), \Phi_0)
\end{equation}
where $X^*$ is the image (with width $W$ and height $H$) to be found, $\Phi$ is the representation of $X$, $\Phi_0$ is the representation of the metadata, and $L$ is the loss function (\eg, $l_2$).

\subsubsection{Distillation based on class-similarities}
Nayak \etal ~\cite{nayak2019zero} argued that the metadata used in \cite{lopes2017data, bhardwaj2019dream} are actually not complete data-free approaches, since the metadata is formed using the training data itself.
They instead proposed a zero-shot KD approach, in which no data samples and no metadata information are used. In particular, the approach obtains useful prior information about the underlying data distribution in the form of \textit{class similarities} from the model parameters of the teacher. This prior information can be further utilized for crafting data samples (also called data impressions (DIs)) by modeling the output space of the teacher model as a Dirichlet distribution. The class similarity matrix, similar to \cite{tung2019similarity}, is calculated based on the softmax layer of the teacher model. The objective of the data impression $X_i^k$ can be formulated based on the cross-entropy loss:
\begin{equation}
    X_i^k =\arg\min_X L_{CE}(y_i^{k}, T(X, \theta_T, \tau))
\end{equation}
where $y_i^{k}$ is the sampled $i$-th softmax vector, and $k$ is the certain class. 

\subsubsection{Distillation using generator}
Considering the limitation of metadata and similarity-based distillation methods, some works \cite{ye2020datafree, fang2019data, chen2019data, yin2019dreaming, yoo2019knowledge, micaelli2019zero} propose novel data-free KD methods via adversarial learning \cite{goodfellow2014generative, wang2020deceiving, wang2019event}. Although the tasks and network structures vary in these methods, most are built on a common framework. That is, the pretrained teacher network is fixed as a discriminator, while a generator is designed to synthesize training samples, given various input source (\eg, noise \cite{chen2019data,yin2019dreaming, yoo2019knowledge, ye2020datafree}). However, slight differences exist in some studies. Fang \etal ~\cite{fang2019data} point out the problem of taking the teacher as the discriminator since the information of the student is ignored, and generated samples cannot be customized without the student. Thus, they take both the teacher and the student as the discriminator to reduce the discrepancy between them, while a generator is trained to generate some samples to adversarially enlarge the discrepancy. In contrast, Ye \etal ~\cite{ye2020datafree} focus more on strengthening the generator structure, and three generators are designed and subtly used. Specifically, a group-stack generator is trained to generate the images originally used for pre-training the teachers, and the intermediate activations. Then, a dual generator takes the generated image as the input, the dual part is taken as the target network (student), and regrouped for multi-label classifications. To compute the adversarial loss for both the generated image and the intermediate activations, multiple group-stack discriminators (multiple teachers) are also designed to amalgamate multi-knowledge into the generator. Yoo \etal ~\cite{yoo2019knowledge} make the generator take two inputs: a sampled class label $y$, and noise $z$. Meanwhile, a decoder is also applied to reconstruct the noise input $z'$ and class label $y'$ from the fake data $x'$ generated by the generator from the noise input $z$, and class label $y$. Thus, by minimizing the errors between $y$ and $y'$ and between $z$ and $z'$, the generator generates more reliable data. Although adversarial loss is not used in \cite{yin2019dreaming}, the generator (called DeepInversion) taking an image prior regularization term to synthesize images is modified from DeepDream \cite{mordvintsev2015inceptionism}. 

\subsubsection{Open challenges for data-free distillation}
Although data-free KD methods have shown considerable potential and new directions for KD, there still exist many challenges. First, the recovered images are unrealistic and low-resolution, which may not be utilized in some data-captious tasks (\eg, semantic segmentation). Second, training and computation of the existing methods might be complicated due to the utilization of many modules. Third, diversity and generalization of the recovered data are still limited, compared with the methods of data-driven distillation. Fourth, the effectiveness of such methods for low-level tasks (\eg, image super-resolution) needs to be studied further.   

\subsection{Distillation with a few data samples}
\label{distill_fewdata}
\textbf{Overall insight:} \textit{How to perform efficient knowledge distillation with only a small amount of training data?} 

Most KD methods with S-T structures, such as \cite{hinton2015distilling, kim2018paraphrasing, park2019feed, chung2020featuremaplevel}, are based on matching information (\eg, logits, hints) and optimizing the KD loss with the fully annotated large-scale training dataset. As a result, the training is still data-heavy and processing-inefficient. To enable efficient learning of the student while using small amount of training data, some works 
\cite{liu2019semantic, li2018few, kimura2018few, bai2019few, kulkarni2017knowledge} propose few-sample KD strategies. The technical highlight of these methods is based on generating pseudo training examples, or aligning the teacher and the student with layer-wise estimation metrics. 

\begin{figure*}[t!]
    \centering
    \includegraphics[width=\textwidth]{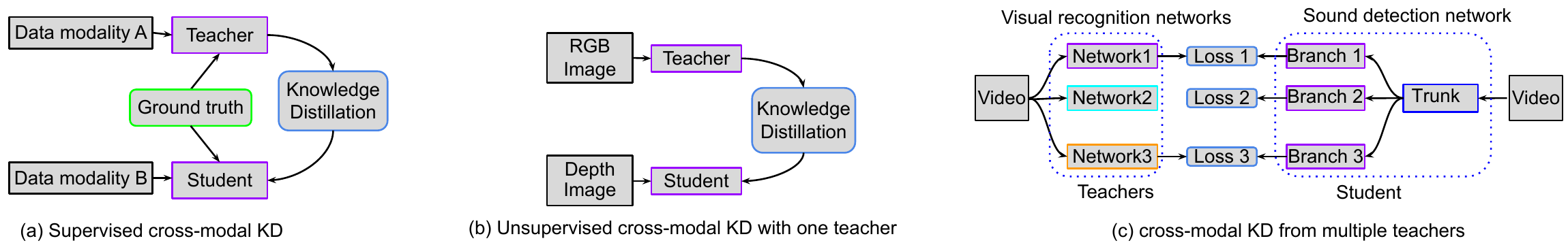}
    \caption{Graphical illustration of cross-modal KD methods. (a) supervised cross-modal KD from the teacher with one modality to the student with another modality. (b) unsupervised cross-modal KD with one teacher. (c) unsupervised cross-modal KD with multiple teachers, each of which is transferring the discriminative knowledge to the student.}
    \label{fig:cross_modal_dis}
\end{figure*}

\subsubsection{Distillation via pseudo examples}
\textbf{Insight:} \textit{If training data is insufficient, try to create pseudo examples for training the student.}

\cite{kimura2018few, liu2019semantic, kulkarni2017knowledge} focus on creating pseudo training examples when training data is scarce and leading to overfitting of the student network. Specifically, Kimura \etal ~\cite{kimura2018few} adopt the idea of inducing points \cite{snelson2006sparse} to generate pseudo training examples, which are then updated by applying adversarial examples \cite{szegedy2015going,goodfellow2014explaining}, and further optimized by an imitation loss. Liu \etal ~\cite{liu2019semantic} generate pseudo ImageNet \cite{deng2009imagenet} labels from a teacher model (trained with ImageNet), and also utilize the semantic information (\eg, words) to add a supervision signal for the student. Interestingly, Kulkarni \etal ~\cite{kulkarni2017knowledge} create a `mismatched' unlabeled stimulus (\eg, soft labels of MNIST dataset \cite{lecun1998gradient} provided by the teacher trained on CIFAR dataset \cite{krizhevsky2009learning}), which are used for augmenting a small amount of training data to train the student. 

\subsubsection{Distillation via layer-wise estimation}
\textbf{Insight:} \textit{Layer-wise distillation from the teacher network via estimating the accumulated errors on the student network can also achieve the purpose of few-example KD.}

In Bai \etal ~\cite{bai2019few} and Li \etal ~\cite{li2018few}, the teacher network is first compressed to create a student via network pruning \cite{zhu2017prune}, and layer-wise distillation losses are then applied to reduce the estimation error on given limited samples. To conduct layer-wise distillation, Li \etal ~\cite{li2018few} add a $1\times1$ layer after each pruned layer block in the student, and estimate the least-squared error to align the parameters with the student. Bai \etal ~\cite{bai2019few} employ cross distillation losses to mimic the behavior of the teacher network, given its current estimations. 

\subsubsection{Challenges and potentials}
Although KD methods with a small number of examples inspired by the techniques of data augmentation and layer-wise learning are convincing, these techniques are still confined by the structures of teacher networks. This is because most methods rely on network pruning from teacher networks to create student networks. Besides, the performance of the student is heavily dependent on the amount of the crafted pseudo labels, which may impede the effectiveness of these methods. Lastly, most works focus on generic classification tasks, and it is unclear whether these methods are effective for tasks without class labels (\eg, low-level vision tasks).

\begin{table*}[t!]
\caption{A taxonomy of cross-modal knowledge distillation methods.
}
\vspace{-15pt}
\small
\begin{center}
\begin{tabular}{c|c|c|c|c|c|c|c}
\hline
 Method & Use GT & \thead{Source \\modality}  & \thead{Target \\modality} & \thead{Number of \\teachers} & Online KD & Knowledge & \thead{Model \\compression}   \\
\hline\hline
Ayter \cite{aytar2016soundnet} & \xmark & RGB frames & Sound & Two &\xmark &Logits & \xmark \\ \hline
Su \cite{su2016adapting} & \checkmark & HR image map& LR image & One & \xmark & Soft labels & \checkmark \\ \hline
Nagrani \cite{nagrani2018learnable} & \checkmark & RGB frames & Voice & One & \checkmark & Soft labels&  \xmark \\ \hline
Nagrani \cite{nagrani2018seeing} & \checkmark & Voice/face & Face/voice  & Multiple  & \checkmark & Features &\xmark  \\  \hline
Hoffman \cite{hoffman2016cross} &  \checkmark &RDG images& Depth images &  One & \xmark & Features & \xmark \\ \hline

Afouras  \cite{afouras2019asr} &  \xmark &Audio& Video &  One & \xmark & Soft labels & \xmark  \\ \hline
Albanie \cite{albanie2018emotion}  &\xmark& Video frames& Sound & One & \xmark & Logits & \xmark\\ \hline
Gupta \cite{gupta2016cross} &\xmark & RGN images  & Depth images & One & \xmark & Soft labels & \xmark \\ \hline

Salem \cite{salem2019learning}  &\xmark& \thead{Scene \\ classification, \\object detection}& Localization & Three & \xmark & Soft labels & \xmark \\ \hline
Thoker \cite{thoker2019cross} & \xmark & RGB video & Skeleton data & One & \xmark & Logits & \xmark\\ \hline
Zhao \cite{zhao2018through} & \xmark & RGB frames & Heatmaps & One & \xmark &\thead{Confidence\\ maps}  & \xmark\\ \hline
Owens~  \cite{owens2016ambient} & \xmark & Sound  & Video frames & One & \xmark & Soft labels & \xmark\\ \hline 
Arandjelovic \cite{arandjelovic2017look} & \xmark & Video frames & Audio & One & \xmark & Features & \xmark\\ \hline
Do \cite{do2019compact} & \checkmark & \thead{Image,\\Questions,\\Answer info.} & Image questions & Three & \xmark & Logits & \checkmark\\ \hline
Aytar \cite{aytar2017see} & \checkmark & Image & \thead{Sound, \\Image, Text} & Three & \xmark & Features & \xmark\\ \hline
Kim \cite{kim2018learning} & \xmark & Sound/images & Images/sound & One & \xmark & Features & \xmark\\  \hline
Dou \cite{dou2020unpaired} & \xmark & CT images & MRI images & One & \checkmark & Logits & \checkmark\\ \hline
Hafner \cite{hafner2018cross}& \xmark & Depth images & RGB images & One & \xmark & Embeddings & \xmark\\ \hline
Gan \cite{gan2019self}  & \checkmark & Video frame & Sound & One & \xmark & \thead{Feature \\ soft labels} & \xmark \\ \hline
Perez \cite{perez2020audio} & \checkmark & \thead{RGB video\\Acoustic images} & Audio & One & \xmark & \thead{Soft labels} & \xmark \\ 
\hline
\end{tabular}
\end{center}
\vspace{-10pt}
\label{table:cross_modal_KD}
\end{table*}

\subsection{Cross-modal distillation}
\label{cross_modal_sec}
\textbf{Overall insight:} \textit{KD for cross-modal learning is typically performed with network architectures containing modal-specific representations or shared layers, utilizing the training images in correspondence of different domains.} 


One natural question we ask is if it is possible to transfer knowledge from a pre-trained teacher network for one task to a student learning another task, while the training examples are in correspondence across domains. \textit{Note that KD for cross-modal learning is essentially different from that for domain adaptation, in which data are drawn independently from different domains, but the tasks are the same.}

Compared to previously mentioned KD methods focused on transferring supervision within the same modality between the teacher and the student, cross-modal KD uses the teacher's representation as a supervision signal to train the student learning another task. In this problem setting, the student needs to rely on the visual input of the teacher to accomplish its task. Following this, many novel cross-modal KD methods \cite{aytar2016soundnet,albanie2018emotion, gupta2016cross, zhao2018through, do2019compact, thoker2019cross,salem2019learning, afouras2019asr, dou2020unpaired, hafner2018cross,su2016adapting, hoffman2016cross, owens2016ambient,arandjelovic2017look,aytar2017see,nagrani2018learnable,nagrani2018seeing, zhao2020knowledge} have been proposed. We now provide a systematic analysis of the technical details, and point the challenges and potential of cross-domain distillation.

\subsubsection{Supervised cross-modal distillation}
Using the ground truth labels for the data used in the student network is the common way of cross-modal KD, as shown in Fig.~\ref{fig:cross_modal_dis}(a). 
 Do \etal ~\cite{do2019compact}, Su \etal ~\cite{su2016adapting}, Nagrani \etal ~\cite{nagrani2018learnable}, Nagrani \etal ~\cite{nagrani2018seeing} and Hoffman \etal ~\cite{hoffman2016cross} rely on supervised learning for cross-modal transfer. Several works \cite{nagrani2018seeing,nagrani2018learnable, afouras2019asr, perez2020audio} leverage the synchronization of visual and audio information in the video data, and learn a joint embedding between the two modalities. Afouras \etal ~\cite{afouras2019asr} and Nagrani \etal ~\cite{nagrani2018seeing} transfer the voice knowledge to learn a visual detector, while Nagrani \etal ~\cite{nagrani2018learnable} utilize visual knowledge to learn a voice detector (student). In contrast, Hoffman \etal ~\cite{hoffman2016cross}, Do \etal ~\cite{do2019compact} and Su \etal ~\cite{su2016adapting} focus on different modalities in the visual domain only. In particular, Hoffman \etal ~\cite{hoffman2016cross} learn a depth network by transferring the knowledge from an RGB network, and fuse the information across modalities. This improves the object recognition performance during the test time. Su \etal ~\cite{su2016adapting} utilize the knowledge from high-quality images to learn a classifier with better generalization on low-quality image (paired).  

\subsubsection{Unsupervised cross-modal distillation} Most cross-modal KD methods exploit unsupervised learning, since the labels in target domains are hard to get. Thus, these methods are also called distillation `\textit{in the wild}'. In this setting, the knowledge from the teacher's modality provides \textit{supervision} for the student network. To this end, some works  \cite{afouras2019asr,albanie2018emotion,aytar2016soundnet,gupta2016cross,hafner2018cross,salem2019learning,thoker2019cross,zhao2018through,owens2016ambient,aytar2017see,arandjelovic2017look,dou2020unpaired,kim2018learning, zhao2020knowledge} aimed for cross-modal distillation in an unsupervised manner.   

\subsubsection{Learning from one teacher} Afouras \etal ~\cite{afouras2019asr}, Albanie \etal ~\cite{albanie2018emotion}, Gupta \etal ~\cite{gupta2016cross}, Thoker \etal ~\cite{thoker2019cross}, Zhao \etal ~\cite{zhao2018through}, Owens \etal ~\cite{owens2016ambient}, Kim \etal ~\cite{kim2018learning}, Arandjelovic \etal ~\cite{arandjelovic2017look}, Gan \etal ~\cite{gan2019self}, and Hafner \etal ~\cite{hafner2018cross} focus on distilling knowledge from one teacher (see Fig.~\ref{fig:cross_modal_dis}(b)), and mostly learn a single student network. Thoker \etal ~\cite{thoker2019cross} and Zhao \etal ~\cite{zhao2018through} learn two students. Especially, Thoker \etal ~ refer to mutual learning \cite{zhang2018deep}, where two students learn from each other based on two KL divergence losses. In addition, Zhao \etal ~\cite{zhao2018through} exploit the feature fusion strategy, similar to \cite{kim2019feature,  ke2019dual} to learn a more robust decoder. Do \etal ~\cite{do2019compact} focuses on unpaired images of two modalities, and learns a semantic segmentation network (student) using the knowledge from the other modality (teacher).  

 \subsubsection{Learning from multiple teachers} Aytar \etal ~\cite{aytar2016soundnet}, Salem \etal ~\cite{salem2019learning}, Aytar \etal ~\cite{aytar2017see} and Do \etal ~\cite{do2019compact} exploit the potential of distilling from multiple teachers as mentioned in Sec.~\ref{multi_teach}. Most methods rely on concurrent knowledge among visual, audio, and textual information, as shown in Fig.~\ref{fig:cross_modal_dis}(c). However, Salem \etal ~\cite{salem2019learning} focus on the visual modality only, where teachers learn the information of object detection, image classification, and scene categorization via a multi-task approach, and distill the knowledge to a single student. 
 
 \subsubsection{Potentials and open challenges}
 \noindent \textbf{Potentials:} Based on the analysis of the existing cross-modal KD techniques in Table.~\ref{table:cross_modal_KD}, we can see that cross-modal KD expands the generalization capability of the knowledge learned from the teacher models. Cross-domain KD has considerable potential in \textit{relieving the dependence} for a large amount of labeled data in one modality or both. In addition, cross-domain KD is more \textit{scalable}, and can be \textit{easily} applied to \textit{new} distillation tasks. Moreover, it is advantageous for learning multiple modalities of data `in the wild', since it is relatively easy to get data with one modality based on other data. In visual applications, cross-modal KD has the potential to distill knowledge among images taken from different types of cameras. For instance, one can distill knowledge from an RGB image to event streams (stacked event images from event cameras) \cite{su2016adapting, wang2020eventsr}. 
 
\noindent \textbf{Open challenges:} Since the knowledge is the transferred representations (\eg, logits, features) of teacher models, ensuring the robustness of the transferred knowledge is crucial. We hope to transfer the good representations, but negative representations do exist. Thus, it is imperative that the supervision provided by the teachers is complementary to the target modality. Moreover, existing cross-modal KD methods are highly dependent on data sources (\eg, video, images), but finding data with paired (\eg, RGB image with depth pair) or multiple modalities (class labels, bounding boxes and segmentation labels) is not always an easy task. We are compelled to ask if it is possible to come up with a way for data-free distillation or distillation with a few examples? In other words, is it possible to just learn a student model with the data from the target modality based on the knowledge of the teacher, without referencing the source modality?  

Moreover, existing cross-modal KD methods are mostly offline methods, which are computation-heavy and memory-intensive. Thus, it would be better if an online KD strategy is considered. Lastly, some works (\eg, \cite{salem2019learning,aytar2017see}) learn a student model using the knowledge from multiple teachers. However, the student is less versatile or modality-dependent. Inspired by the analysis of Sec.~\ref{customize_student}, we open a research question: Is it possible to learn a versatile student that can perform tasks from multiple modalities? 

\begin{figure*}[t!]
    \centering
    \includegraphics[width=0.9\textwidth]{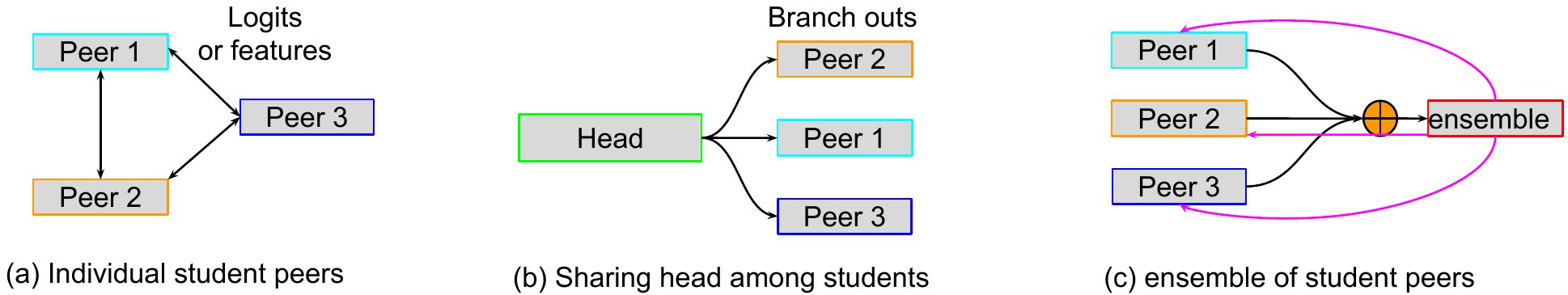}
    \caption{An illustration of online KD methods. (a) online KD with individual student peer learning from each other, (b) online KD with student peers sharing trunk (head) structure, (c) online KD by assembling the weights of each student to form a teacher or group leader.}
    \label{fig:online_distillation}
\end{figure*}

\section{Online and Teacher-free distillation}
\subsection{Online distillation}
\label{online_kd}
\textbf{Overall insight:} \textit{With the absence of a pre-trained powerful teacher, simultaneously training a group of student models by learning from peers’ predictions is an effective substitute for two-stage (offline) KD}

In this section, we provide a deeper analysis of online (one-stage) KD methods in contrast to the previously discussed offline (two-stage) KD methods. Offline KD methods often require pre-trained high-capacity teacher models to perform one-way transfer \cite{hinton2015distilling, Ahn_2019_CVPR, kim2018paraphrasing, you2017learning,fukuda2017efficient, hou2017dualnet, mullapudi2019online, gao2019multistructure, lin2019mod}. However, it is sometimes difficult to get such ‘good’ teachers, and the performance of the student gets degraded when the gap of network capacity between the teacher and the student is significant. In addition, two-stage KD requires many parameters, resulting in higher computation costs. To overcome these difficulties, some studies focus on online KD that simultaneously trains a group of student peers by learning from the peers’ predictions.
 
\subsubsection{Individual student peers}
 Zhang \etal ~\cite{zhang2018deep}, Gao \etal ~\cite{gao2019multistructure} and Anil \etal ~\cite{anil2018large} focus on online \textit{mutual learning} \cite{zhang2018deep} (also called codistilation) in which a pool of untrained student networks with the same network structure simultaneously learns the target task. In such a peer-teaching environment, each student learns the average class probabilities from the other (see Fig.~\ref{fig:online_distillation}(a)). However, Chung \etal ~\cite{chung2020featuremaplevel} also employ individual students, and additionally design a feature map-based KD loss via adversarial learning. Hou \etal ~\cite{hou2017dualnet} proposed DualNet, where two individual student classifiers were fused into one fused classifier. During training, the two student classifiers are locally optimized, while the fused classifier is globally optimized as a mutual learning method. Other methods such as \cite{mullapudi2019online, cioppa2019arthus}, focus on online video distillation by periodically updating the weights of the student, based on the output of the teacher. Although codistillation achieves parallel learning of students, \cite{zhang2018deep, anil2018large, chung2020featuremaplevel, hou2017dualnet} do not consider the ensemble of peers' information as done in other works such as \cite{chen2019online,gao2019multistructure}.  

\subsubsection{Sharing blocks among student peers} Considering the training cost of employing individual students, some works propose sharing network structures (\eg, head sharing) of the students with branches as shown in Fig.~\ref{fig:online_distillation}(b). Song \etal ~\cite{song2018collaborative} and Lan \etal ~\cite{lan2018knowledge} build the student peers on multi-branch architectures \cite{szegedy2015going}. In such a way, all structures together with the shared trunk layers (often use head layers) can construct individual student peers, and any target student peer network in the whole multi-branch can be optimized.

\subsubsection{Ensemble of student peers}
While using codistillation and multi-architectures can facilitate online distillation, knowledge from all student peers is not accessible. To this end, some studies \cite{lan2018knowledge, kim2019feature, chen2019online,lin2019mod, gao2019multistructure} proposed using the assembly of knowledge (logits information) of all student peers to build an on the fly teacher or group leader, which is in turn distilled back to all student peers to enhance student learning in a closed-loop form, as shown in Fig.~\ref{fig:online_distillation}(c). Note that in ensemble distillation, the student peers can either be independent, or share the same head structure (trunk). The ensemble distillation loss is given by Eqn.~\ref{ensemble_logits_gating} of Sec.~\ref{multi_teach}, where a gating component $g_i$ is added to balance the contribution of each student. Chen \etal~\cite{chen2019online} obtain the gating component $g_i$ based on the self-attention mechanism \cite{vaswani2017attention}.
\subsubsection{Summary and open challenges}
\textbf{Summary:} Based on the above analysis, we have determined that codistillation, multi-architectures, and ensemble learning are three main techniques for online distillation. There are some advantages of online KD compared with offline KD. Firstly, online KD does not require pre-training teachers. Secondly, online learning provides a simple but effective way to improve the learning efficiency and generalizability of the network, by training together with other student peers. Thirdly, online learning with student peers often results in better performance than offline learning.

\noindent \textbf{Open challenges:} There are some challenges in online KD. Firstly, there is a lack of theoretical analysis for why online learning is sometimes better than offline learning. Secondly, in online ensemble KD, simply aggregating students’ logits to form an ensemble teacher restrains the diversity of student peers, thus limiting the effectiveness of online learning. Thirdly, existing methods are confined problems in which ground truth (GT) labels exist (\eg, classification). However, for some problems (\eg, low-level vision problems), ways for the student peers to form effective ensemble teachers need to be exploited.	
     
\begin{figure*}[t!]
    \centering
    \includegraphics[width=\textwidth]{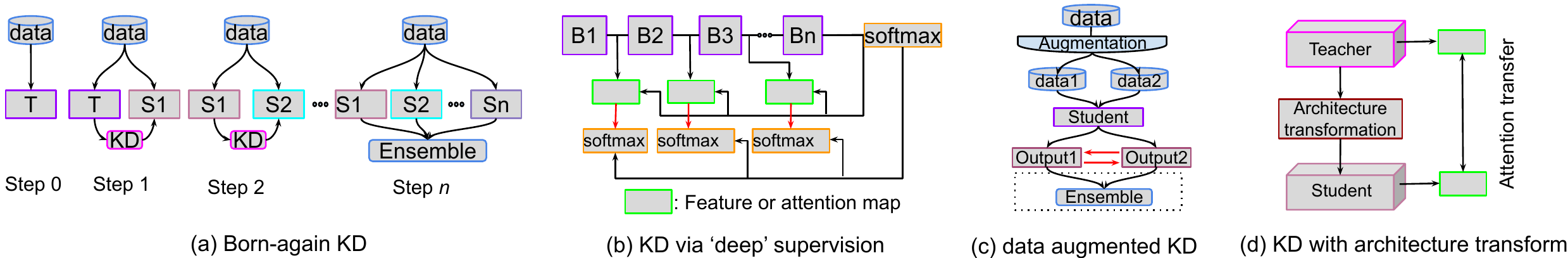}
    \caption{An illustration of self-distillation methods. (a) born-again distillation. Note that $T$ and $S_1, \cdots, S_n$ can be multi-tasks. (b) distillation via `deep' supervision where the deepest branch (B$_n$) is used to distill knowledge to shallower branches. (c) distillation via data augmentation (\eg, rotation, cropping). (d) distillation with network architecture transformation (\eg, changing convolution filters). }
    \label{fig:self_kd}
\end{figure*}     
\subsection{Teacher-free distillation}
\label{self_distillation}
\textbf{Overal insight:} \textit{Is it possible to enable the student to distill knowledge by itself to achieve plausible performance?}

The conventional KD approaches \cite{hinton2015distilling, romero2014fitnets, tung2019similarity, kim2019feature, yim2017gift} still have many setbacks to be tackled, although significant performance boost has been achieved. 
First of all, these approaches have low efficiencies, 
since student models scarcely exploit all knowledge from the teacher models. Secondly, designing and training high-capacity teacher models still face many obstacles. Thirdly, two-stage distillation requires high computation and storage costs. To tackle these challenges, several novel self-distillation frameworks \cite{mobahi2020self,xu2019data,hahn2019self,lee2019rethinking, furlanello2018born, crowley2018moonshine, zhang2019your, hou2019learning, clark2019bam, luan2019msd, yang2019training, lan2018self} have been proposed recently. The goal of self-distillation is to learn a student model by distilling knowledge in itself without referring to other models. We now provide a detailed analysis of the technical details for self-distillation.

\begin{table*}[t!]
\caption{A taxonomy of self-distillation methods. Logits and hints indicate the knowledge to be distilled. `Deep' supervision is for self-distillation from deepest branch (or layer) of the student network. One-stage KD is checking whether self-distillation is achieved in one step. \checkmark/ \xmark is for yes/no.}
\vspace{-15pt}
\small
\begin{center}
\begin{tabular}{c|c|c|c|c|c|c|c}
\hline
 Method & Logits & Hints & \thead{Data \\ augmentation} & \thead{'Deep' \\supervision} & One-stage KD & \thead{Multi-task KD} & \thead{Architecture \\transformation}   \\
\hline\hline
Clarm \cite{clark2019bam} & \checkmark & \xmark & \xmark &\xmark &\xmark &\checkmark & \xmark \\ \hline
Chowley \cite{crowley2018moonshine} & \xmark & Attention map& \xmark & \xmark & \xmark & \xmark & \checkmark \\ \hline
Furlanello \cite{furlanello2018born} & \checkmark & \xmark & \xmark & \xmark & \xmark & \xmark & \xmark \\ \hline
Hahn \cite{hahn2019self} & \checkmark & \xmark & \xmark  & \xmark  & \xmark & \xmark & \xmark \\  \hline
Hou \cite{hou2019learning} &  \xmark &Attention maps& \xmark &  \checkmark & \checkmark & \xmark & \xmark \\ \hline
Luan \cite{luan2019msd} &  \checkmark &Feature maps& \xmark &  \checkmark& \checkmark & \checkmark &\xmark  \\ \hline
Lee \cite{lee2019rethinking}  &\checkmark& \xmark& \checkmark& \xmark & \xmark & \xmark & \xmark\\ \hline
Xu \cite{xu2019data} &\checkmark & Feature maps  & \checkmark & \xmark & \checkmark & \xmark &\xmark \\ \hline
Zhang \cite{zhang2019your}  &\checkmark& Feature maps& \xmark & \checkmark & \checkmark & \xmark &\xmark \\ \hline
Yang \cite{yang2019training} & \checkmark & \xmark & \xmark & \xmark & \xmark & \xmark & \xmark\\

\hline
\end{tabular}
\end{center}
\vspace{-10pt}
\label{table:selfKD_comp}
\end{table*}

\subsubsection{Born-again distillation}
\textbf{Insight}: \textit{Sequential self-teaching of students enables them to become masters, and outperform their teachers significantly.} 

Furlanello \etal ~\cite{furlanello2018born} initializde the concept of self-distillation, in which the students are parameterized identically to their teachers, as shown in Fig.~\ref{fig:self_kd}(a). Through sequential teaching, the student is continuously updated, and at the end of the procedure, additional performance gains are achieved by an ensemble of multiple student generations. Hahn \etal ~\cite{hahn2019self} apply born-again distillation \cite{furlanello2018born} to natural language processing. Yang \etal ~\cite{yang2019training} observe that it remains unclear how S-T optimization works, and they then focus on putting strictness (adding an extra term to the standard cross-entropy loss) to the teacher model, such that the student can better learn inter-class similarity, and potentially prevent over-fitting. Instead of learning a single task, Clark \etal ~\cite{clark2019bam} extend \cite{furlanello2018born} to the multi-task setting, where single-task models are distilled sequentially to teach a multi-task model. Since the born-again distillation approach is based on the multi-stage training, it is less efficient and computation-heavy compared to the following methods. 

\subsubsection{Distillation via `deep' supervision}
\textbf{Insight:} \textit{The deeper layers (or branches) in the student model contains more useful information than those of shallower layers.}

Among the methods, Hou \etal ~\cite{hou2019learning}, Luan \etal ~\cite{luan2019msd} and Zhang \etal ~\cite{zhang2019your} propose similar approaches where the target network (student) is divided into several shallow sections (branches) according to their depths and original structures (see Fig.~\ref{fig:self_kd}(b)). As the deepest section may contain more useful and discriminative feature information than shallower sections, the deeper branches can be used to distill knowledge to the shallower branches. In contrast, in \cite{hou2019learning}, instead of directly distilling features, attention-based methods used in \cite{zagoruyko2016paying} are adopted to force shallower layers to mimic the attention maps of deeper layers. Luan \etal ~\cite{luan2019msd} make each layer branch (ResNet block) a classifier. Thus, the deepest classifier is used to distill earlier the classifiers' feature maps and logits. 

\subsubsection{Distillation based on data augmentation}
\label{dis_dataaug}
\textbf{Insight:} \textit{Data augmentation (\eg, rotation, flipping, cropping, etc) during training forces the student network to be invariant to augmentation transformations via self-distillation}. 

Although most methods focus on how to better supervise student in self-distillation,
data representations for training the student are not fully excavated and utilized. To this end, Xu \etal ~\cite{xu2019data} and Lee \etal ~\cite{lee2019rethinking} focus on self-distillation via data augmentation of the training samples, as shown in Fig.~\ref{fig:self_kd}(c). There are some advantages to such a framework. First, it is efficient and effective to optimize a single student network without branching or the assistance of other models. Second, with data-to-data self-distillation, the student learns more inherent representations for generalization. Third, the performance of the student model is significantly enhanced with relatively low computation cost and memory load. 

Xu \etal ~\cite{xu2019data} apply random mirror and cropping to the batch images from the training data. Besides, inspired by mutual learning \cite{zhang2018deep}, the last feature layers and softmax outputs of the original batch image and distorted batch images are mutually distilled via MMD loss \cite{huang2017like} and KL divergence loss, respectively. In contrast, Lee \etal ~\cite{lee2019rethinking} consider two types of data augmentation (rotation and color permutation to the same image), and the ensemble method used in \cite{lan2018knowledge, chen2019online, zhu2018knowledge} is employed to aggregate all logits of the student model to one, which is in turn is used by the student to transfer the knowledge to itself.

\subsubsection{Distillation with architecture transformation}
\textbf{Insight:} \textit{A student model can be derived by changing the convolution operators in the teacher model with any architecture change.}

In contrast with all the above-mentioned self-distillation methods, Crowley \etal ~\cite{crowley2018moonshine} proposes structure model distillation for memory reduction by replacing standard convolution blocks with cheaper convolutions, as shown in Fig.~\ref{fig:self_kd}(d). In such a way, a student model that is a simple transformation of the teacher's architecture is produced. Then, attention transfer (AT) \cite{huang2017like} is applied to align the teacher's attention map with that of the student's. 

\subsubsection{Summary and open challenges}
\textbf{Summary:} In Table.~\ref{table:selfKD_comp}, we summarize and compare different self-distillation approaches. Overall, using logits/feature information and two-stage training for self-distillation with `deep' supervision from the deepest branch are main stream. Besides, data augmentation and attention-based self-distillation approaches are promising. Lastly, it is shown that multi-task learning with self-distillation is also a valuable direction, deserving more research. 

\noindent\textbf{Open challenges:} 
There still exist many challenges to tackle. First, theoretical support laks in explaining why self-distillation works better. Mobahi \etal ~\cite{mobahi2020self} provide theoretical analysis for born-again distillation \cite{furlanello2018born} and find that self-distillation may reduce over-fitting by loop-over training, thus leading to good performance. However, it is still unclear why other self-distillation methods (\eg, online `deep' supervision \cite{luan2019msd, zhang2019your, hou2019learning}) work better.

In addition, existing methods focus on self-distillation with certain types of group-based network structures (\eg, ResNet group). Thus, the generalization and flexibility of self-distillation methods need to be probed further. Lastly, all existing methods focus on classification-based tasks, and it is not clear whether self-distillation is effective for other tasks (\eg, low-level vision tasks).

\section{Label-required/-free distillation}
\textbf{Overall Insight:} \textit{It is possible to learn a student without referring to the labels of training data?}
\vspace{-5pt}
\subsection{Label-required distillation} The success of KD relies on the assumption that labels provide the required level of semantic description for the task at hand \cite{hinton2015distilling, bucilua2006model}.  For instance, in most existing KD methods for classification-related tasks \cite{hinton2015distilling, park2019relational, szegedy2015going, park2019feed, hegde2019variational, yuan2019revisit, Ahn_2019_CVPR}, image-level labels are required for learning student network. Meanwhile, some works exploit the pseudo labels when training data are scarce. We now provide a systematic analysis for these two types of methods.   

\subsubsection{KD with original labels.} Using the ground truth labels for the data used in the student network is the common way for KD. As depicted in Eq.\ref{loss_student}, the overall loss function is composed of the student loss and the distillation loss. The student loss is heavily dependent on the ground truth label $y$. Following this fashion, main-stream methods mostly utilize the original labels and design better distillation loss terms to achieve better performances \cite{chen2019online, wang2019heterogeneous, zhou2018graph, chen2017learning, guo2018learning, huang2017like}. This convention has also been continually adopted in recent KD methods, such as online distillation \cite{chen2019online, mullapudi2019online}, teacher-free distillation \cite{gan2019self,zhang2018better, zhang2019your}, and even cross-modal learning \cite{gupta2016cross, su2016adapting, nagrani2018learnable, nagrani2018seeing, do2019compact, aytar2017see}. While using labels expand the generalization capability of knowledge for learning student network, such  approaches fail when labels are scarce or unavailable.  

\subsubsection{KD with pseudo labels.} Some works also exploit the pseudo labels. The most common methods can be discomposed into two groups. The first one aims to create noisy labels. \cite{li2017learning, xie2019self, xu2019positive, sarfraz2019noisy} propose to leverage large number of noisy labels to augment small amount of clean labels, which turns to improve the generalization and robustness of student network. The second group of methods focus on creating pseudo labels via metadata \cite{lopes2017data}, class similarities \cite{nayak2019zero} or generating labels \cite{ye2020datafree,fang2019data}, etc. 
\vspace{-5pt}
\subsection{Label-free distillation} However, in real-world applications, labels for the data used in the student network is not always easy to obtain. Hence, some attempts have been taken for label-free distillation. We now provide more detailed analysis for these methods. 

\subsubsection{KD with dark knowledge.} This has inspired some works to exploit KD without the requirement of labels. Based on our review, label-free distillation is mostly achieved in cross-modal learning, as discussed in Sec.~\ref{cross_modal_sec}. With paired modality data (\eg, video and audio), where the label of modality (\eg, video) is available, the student learns the end tasks only based on the distillation loss in Eq.\ref{loss_student} \cite{aytar2016soundnet, afouras2019asr, salem2019learning, owens2016ambient, arandjelovic2017look, hafner2018cross}. That is, in this situation, the dark knowledge  of teacher provides `supervision' for the student network.

\subsubsection{Creating meta knowledge.} Recently, a few methods \cite{ye2020datafree, fang2019data,yin2019dreaming} propose data-/label-free frameworks for KD.  The core technique is to craft samples with labels by using either feature or logits information, which are also called meta knowledge. Although these methods point out an interesting direction for KD, there still exist many challenges to achieve reasonable performance.  

\subsection{Potential and challenges.} Label-free distillation is a promising since it relieves the need for data annotation. However, the current status of research shows that there still exist many uncertainties and challenges in this direction. The major concern is how to ensure that the `supervision' provided by teacher is reliable enough. As some works interpret the knowledge as a way a label regularization \cite{yuan2019revisit} or class similarities  \cite{cheng2020explaining}, it is crucial to guarantee that the knowledge can be captured by the student. 

Another critical challenge of label-required distillation is that, in Eq.~\ref{loss_student}, the KD loss term never involves any label information although the student loss (\eg, cross-entropy loss) uses labels. As labels provide informative knowledge for the student learning, it is worthwhile to find a way to use labels for the KD loss to further improve the performance. While it is generally acknowledged that a pretrained teacher has already mastered sufficient knowledge about the label information, its predictions still have a considerable gap with the ground truth labels. Based on our literature review, there exists some difficulties to bring label information to the distillation loss. That is, to bring the label information, the teacher might need to be updated or fine-tuned, which may cause additional computation cost. However, with some recent attempts based on meta learning or continual learning, it is possible to learn the label information with only a few examples. Besides, it might be possible to learn a bootstrap representation based on the labels, as done in \cite{grill2020bootstrap}, and further incorporate the information to the KD loss.  We do believe this direction is promising in real-world applications and thus expect future research might move towards this direction.

\section{KD with novel learning metrics}

\begin{figure*}[t!]
    \centering
    \includegraphics[width=0.85\textwidth]{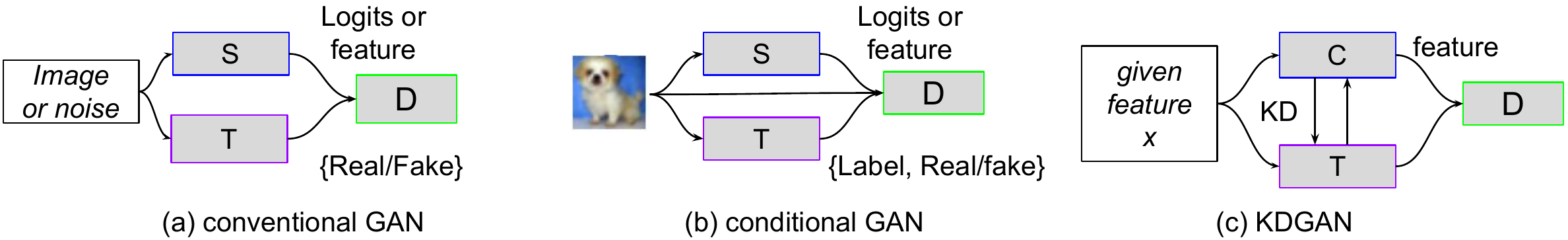}
    \caption{ An illustration of GAN-based KD methods. 
     (a) KD based on GAN \cite{goodfellow2014generative} where discriminator $D$ discerns the feature/logits of $T$ and $S$; (b) KD based on conditional GAN (CGAN) \cite{mirza2014conditional} where the input also functions as a condition to $D$; (c) KD based on TripleGAN \cite{chongxuan2017triple} where the classifier $C$, teacher $T$ and $D$ play a minmax game. }
    \label{fig:kd_based_on_GAN}
\end{figure*}

\subsection{Distillation via adversarial learning}
\label{kd_gan}
\textbf{Overall Insight:} \textit{GAN can help learn the correlation between classes and preserve the multi-modality of S-T framework, especially when student has relatively small capacity.}

In Sec.~\ref{one_teacher}, we have discussed the two most popular approaches for KD. However, the key problem is that it is difficult for the student to learn the true data distribution from the teacher, since the teacher can not perfectly model the real data distribution. Generative adversarial networks (GANs) \cite{goodfellow2014generative, chongxuan2017triple,wang2019event, wang2020deceiving, wang2020eventsr} have been proven to have potential in learning the true data distribution in image translation. To this end, recent works \cite{wang2018kdgan, xu2017training, belagiannis2018adversarial, liu2018teacher, liu2018ktan, wang2018adversarial, roheda2018cross, xu2018training, chen2019data, shen2019meal, heo2019knowledge, liu2019exploiting, goldblum2019adversarially, hong2019gan, zhai2019lifelong, wang2019minegan,gao2019adversarial,shen2019adversarial,liu2019structured, chung2020featuremaplevel,aguinaldo2019compressing, li2020gan} have tried to explore adversarial learning to improve the performance of KD. 
These works are, in fact, built on three fundamental prototypes of GANs \cite{goodfellow2014generative, li2017learning, mirza2014conditional}. Therefore, we formulate the principle of these three types of GANs, as illustrated in Fig.~\ref{fig:kd_based_on_GAN}, and analyze the existing GAN-based KD methods.

\subsubsection{A basic formulation of GANs in KD} 
The first type of GAN, as shown in Fig.~\ref{fig:kd_based_on_GAN}(a), is proposed to generate continuous data by training a generator $G$ and discriminator $D$, which penalizes the generator $G$ for producing implausible results. The generator $G$ produces synthetic examples $G(z)$ (\eg, images) from the random noise $z$ sampled using a specific distribution (\eg, normal) \cite{goodfellow2014generative}. These synthetic examples are fed to the discriminator $D$ along with the real examples sampled from the real data distribution $p(x)$. The discriminator $D$ attempts to distinguish the two inputs, and both the generator $G$ and discriminator $D$ improve their respective abilities in a minmax game until the discriminator $D$ is unable to distinguish the fake from the real. The objective function can be written as follows:
\begin{equation}
\begin{split}
     \min_G\max_D J(G,D) = \E_{x \sim p(x)}[log(D(x))] + \\
     \E_{z \sim p(z)}[log(1-D(G(z)))] 
\end{split}
\label{adv_loss}
\end{equation}
where $p_z(z)$ is the distribution of noise (\eg, uniform or normal). 

The second type of GAN for KD is built on conditional GAN (CGAN) \cite{mirza2014conditional,isola2017image, wang2019event, wang2020deceiving}, as shown in Fig.~\ref{fig:kd_based_on_GAN}(b). CGAN is trained to generate samples from a class conditional distribution $c$. The generator is replaced by useful information rather than random noise. Hence, the objective of the generator is to generate realistic data, given the conditional information. Mathematically, the objective function can be written as:
\begin{equation}
\begin{split}
     \min_G\max_D J(G,D) = \E_{x \sim p(x)}[log(D(x|c))] + \\
     \E_{z \sim p(z)}[log(1-D(G(z|c)))] 
\end{split}
\label{cgan_loss}
\end{equation}

Unlike the above-mentioned GANs, triple-GAN \cite{li2017learning} (the third type) introduces a three-player game where there is a classifier $C$, generator $G$, and discriminator $D$, as shown in Fig.~\ref{fig:kd_based_on_GAN}(c). Adversarial learning of generators and discriminators overcomes some difficulties \cite{goodfellow2014generative}, such as not having a good optimization and the failure of the generator to control the semantics of generated samples. We assume that there is a pair of data $(x,y)$ from the true distribution $p(x,y)$. After a sample $x$ is sampled from $p(x)$, $C$ assigns a pseudo label $y$ following the conditional distribution $p_c(y|x)$, that is, $C$ characterizes the conditional distribution $p_c(y|x) \approx p(y|x)$.The aim of the generator is to model the conditional distribution in the other direction $p_g(x|y) \approx p(x|y)$,  while the discriminator determines whether a pair of data (x,y) is from the true distribution $p(x,y)$.  Thus, the minmax game can be formulated as:
\begin{equation}
\begin{split}
     \min_{C,G}\max_D J(C,G,D) = \E_{(x,y) \sim p(x,y)}[log(D(x,y))] + \\
     \alpha \E_{(x,y) \sim p_c(x,y)}[log(1-D(x,y))] +\\
     (1-\alpha) \E_{(x,y)\sim p_g(x,y)}[log(1-D(G(y,z),y))]
\end{split}
\label{tripe_gan_loss}
\end{equation}
where $\alpha$ is a hyper-parameter that controls the relative importance of $C$ and $G$.

\begin{table*}[t!]
\caption{Taxonomy of KD based on adversarial learning. 
}
\small
\begin{center}
\begin{tabular}{c|c|c|c|c|c}
\hline
 Method & GAN type & Purpose & Inputs of $D$ & Number of $D$ & Online KD   \\
\hline\hline
Chen \cite{chen2019data} &  First type &Classification& Logits & One & No  \\ \hline
Belagiannis \cite{belagiannis2018adversarial}  &First type& Classification& Features & One & No \\ \hline
Liu \cite{liu2018ktan}  & First type &\thead{Classification \\ Object detection} & Features  &One & No \\ \hline
Hong \cite{hong2019gan} &First type &Object detection  &Features & Six & No \\ \hline
    Wang \cite{wang2018adversarial}  &First type& Classification& Features & One & No \\ \hline
Aguinaldo \cite{aguinaldo2019compressing}  & First type & Classification & Features&  One & No \\ \hline
Chung \cite{chung2020featuremaplevel} &  First type (LSGAN \cite{mao2017least}) & Classification & Features &  Two/Three  & Yes \\ \hline
Wang \cite{wang2019minegan} & First type(WGAN-GP \cite{gulrajani2017improved})  & Image generation  & Features & One/Multiple & Yes \\ \hline 
Chen \cite{chen2020distilling} &First/Second type& Image translation  & Features & Two & No \\ \hline 
Liu \cite{liu2019structured}  &Second type (WGAN-GP \cite{gulrajani2017improved})& Semantic segmentation& Features & One & No \\ \hline
Xu \cite{xu2017training} &Second type& Classification  & Logits & One & No \\ \hline 
Roheda \cite{roheda2018cross} &Second type  & \thead{Cross-domain \\ surveillance}  & Features & One & Yes\\ \hline 
Zhai \cite{zhai2019lifelong} &Second type (BicyleGAN \cite{zhu2017toward})  & Image translation  & Features & One & Yes\\ \hline 
Liu \cite{liu2018teacher} &Second (AC-GAN \cite{odena2017conditional})  & Image translation  &Features & One & No\\ \hline 
Wang \cite{wang2018kdgan} &Third type  & Image translation & Features & One & No \\ \hline 
Li \cite{li2020gan}   & First/Second type & Image translation & Features & One & No \\ \hline
Fang \cite{fang2019data} & First type & \thead{Classification\\ Semantic segmentation} & Logits & One & No \\ \hline
Yoo \cite{yoo2019knowledge} & Second type & Classification & Logits & One & No \\
\hline
\end{tabular}
\end{center}
\label{table:gan_kd_comp}
\end{table*}

\subsubsection{How does GAN boost KD?}
Based on the aforementioned formulation of GANs, we analyze how they are applied to boost the performance of KD with S-T learning. 

 \noindent\textbf{KD based on the conventional GAN (first type)} Chen \etal \cite{chen2019data} and Fang \etal \cite{fang2019data} focused on distilling the knowledge of \textit{logits} from teacher to student via the first type of GAN, as depicted in Fig.~\ref{fig:kd_based_on_GAN}(a). 
 \footnote{\cite{chen2019data,fang2019data} are data-free KD methods, which will be explicitly discussed in Sec. \ref{data_free}.} There are several benefits of predicting logits based on the discriminator. First, the learned loss, as described using Eqn~\ref{adv_loss}, can be effective in image translation tasks \cite{isola2017image, wang2019event, wang2020deceiving}. The second benefit is closely related to the multi-modality of the network output; therefore, it is not necessary to exactly mimic the output of one teacher network to achieve good student performance as it is usually done \cite{hinton2015distilling, romero2014fitnets}.
 However, the low-level feature alignment is missing because the discriminator only captures the high-level statistics of the teacher and student outputs (logits).

 In contrast, Belagiannis \etal ~\cite{belagiannis2018adversarial}, Liu \etal ~\cite{liu2018ktan}, Hong \etal ~\cite{hong2019gan}, Aguinaldo \etal ~\cite{aguinaldo2019compressing}, Chung \etal ~\cite{chung2020featuremaplevel}, Wang \etal ~\cite{wang2019minegan}, Wang \etal ~\cite{wang2018adversarial}, Chen \etal ~\cite{chen2020distilling}, and Li \etal ~\cite{li2020gan} aimed to distinguish whether the \textit{features} come from the teacher or student via adversarial learning, which effectively pushes the two distributions close to each other. \footnote{Note that in \cite{chung2020featuremaplevel}, least squares GAN (LSGAN) \cite{mao2017least} loss was used and in \cite{wang2019minegan}, Wasserstein GAN-gradient penalty (WGAN-GP) loss \cite{gulrajani2017improved} was used to stabilize training.} The features of the teacher and student are used as inputs to the discriminator because of their \textit{dimensionality}. The feature representations extracted from the teacher are high-level abstract information and easy for classification, which lowers the probability for the discriminator to make a mistake \cite{liu2018ktan}.  
However, the GAN training in this setting is sometimes unstable and even difficult to converge, particularly when the model capacity between the student and teacher is large. To address this problem, some regularization techniques such as dropout \cite{srivastava2014dropout} or $l_2$ or $l_1$ regularization \cite{belagiannis2018adversarial} are added to Eqn.~\ref{adv_loss} to confine the weights. 

\noindent\textbf{KD based on CGAN (second type)} 
Xu \etal ~\cite{xu2017learning} and Yoo \etal ~\cite{ yoo2019knowledge} employed CGAN \cite{mirza2014conditional} for KD, where the discriminator was trained to distinguish whether the \textit{label distribution} (logits) was from the teacher or the student. The student, which was regarded as the generator, was adversarially trained to deceive the discriminator. Liu \etal ~\cite{liu2018teacher} also exploited CGAN for compressing image generation networks. However, the discriminator predicted the class label of the teacher and student, together with an auxiliary classifier GAN \cite{odena2017conditional}. 

In contrast, Roheda \etal ~\cite{roheda2018cross}, Zhai \etal ~\cite{zhai2019lifelong}, Li \etal ~\cite{li2020gan}, Chen \etal ~\cite{chen2020distilling}, and Liu \etal ~\cite{liu2019structured} focused on discriminating the \textit{feature} space of the teacher and student in the CGAN framework. Interestingly, Chen \etal ~\cite{chen2020distilling} deployed two discriminators, namely, the teacher and student discriminators,for compressing image translation networks. To avoid model collapse, Liu \etal ~\cite{liu2019structured} used Wasserstein loss \cite{gulrajani2017improved} to stabilize training.  

\noindent\textbf{KD based on TripleGAN (third type)} In contrast to the distillation methods based on conventional GAN and CGAN , Wang \etal ~\cite{wang2018kdgan} proposed a three-player game named KDGAN, consisting of a classifier (tent), a teacher, and a discriminator (similar to the prototype in TripleGAN \cite{li2017learning}), as shown in Fig.~\ref{fig:kd_based_on_GAN}(c). The classifier and the teacher learn from each other via distillation losses, and are adversarially trained against the discriminator via the adversarial loss defined in Eqn.~\ref{tripe_gan_loss}. By simultaneously optimizing the distillation and the adversarial loss, the classifier (student) learns the true data distribution at equilibrium.

\subsubsection{Summary and open challenges}
In Table~\ref{table:gan_kd_comp}, we summarize existing GAN-based knowledge distillation methods regarding the practical applications, input features of the discriminator $D$, the number of discriminators used, and whether it is one-stage (without the need for the teacher to be trained first). In general, most methods focus on classification tasks based on the first type of GAN (conventional GAN) \cite{goodfellow2014generative} and use the features as the inputs to the discriminator $D$. Besides, it is worth noting that most methods use only one discriminator for discerning the student from the teacher. However, some works such as \cite{chung2020featuremaplevel}, \cite{wang2019minegan} and \cite{chen2020distilling} employ multiple discriminators in their KD frameworks. One can see that most methods follow a two-stage KD paradigm where the teacher is trained first, and then knowledge is transferred to the student via KD loss. In contrast, studies such as \cite{chung2020featuremaplevel, wang2019minegan, roheda2018cross, zhai2019lifelong} also exploit online (one-stage) KD, without the necessity of pre-trained teacher networks. \textit{More detailed analyses of KD methods with respect to online/two-stage distillation and image translation are described in Sec.~\ref{online_kd} and Sec.~\ref{img_trans}, respectively.}

\noindent \textbf{Open challenges:} The first challenge for GAN-based KD is the stability of training, especially when the capacity between the teachers and the students is large. Secondly, it is less intuitive whether using only logits or only features or both as inputs to the discriminator is good because there lacks theoretical support. Thirdly, the advantages of using multiple discriminators are less clear and what features in which position are suitable for training GAN also needs to be further studied.

\subsection{Distillation with graph representations} 
\label{kd_graph}
\noindent \textbf{Overall insight:} \textit{Graphs are the most typical locally connected structures that capture the features and hierarchical patterns for KD.}

Up to now, we have categorized and analyzed the most common KD methods using either logits or feature information. However, one critical issue regarding KD is data. In general, training a DNN requires embedding a high-dimensional dataset to facilitate data analysis. Thus, the optimal goal of training a teacher model is not only to transform the training dataset into a low-dimensional space, but also to analyze the intra-data relations \cite{lee2019graph,liu2019knowledgegraph}. However, most KD methods do not consider such relations. Here, we introduce the definitions of the basic concepts of graph embedding and knowledge graph based on \cite{cai2018comprehensive,hamilton2017representation}. We provide an analysis of existing graph-based KD methods, and discuss new perspectives about KD.

\noindent \subsubsection{Notation and definition}
\begin{defn}
A \textbf{graph} can be depicted as $\mathcal{G} = (V, E)$, where $v \in V$ is a node and $e \in E$ is an edge. A graph $\mathcal{G}$ is associated with a node type mapping function $F_v$: $V \to \mathcal{T}^v$, and an edge type mapping function $F_e$: $E \to \mathcal{T}^e $.
\end{defn}
Here, $\mathcal{T}^v$ and $\mathcal{T}^e$ denote the node types and edge types, respectively. For any $v_i \in V$, there exists a particular mapping type: $F_v(v_i) \in \mathcal{T}^v$. Similar mapping comes to any $e_{ij} \in E$, which is mapped as $F_e(e_{ij}) \in \mathcal{T}^e$, where $i$ and $j$ indicate the $i$-th and $j$-th nodes. 

\begin{defn}
A \textbf{homogeneous graph (directed graph)}, depicted as $\mathcal{G}_{hom} = (V, E)$, is a type of graph in which $|\mathcal{T}^v| = |\mathcal{T}^e|=1$. All nodes and edges in this graph embedding are of one type.
\end{defn}

\begin{figure}[t!]
    \centering
    \includegraphics[width=\columnwidth]{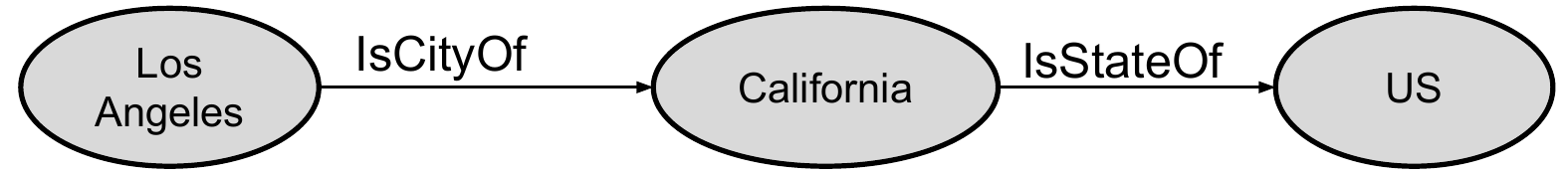}
    \caption{An example of knowledge graph.}
    \label{fig:knoweldge_graph}
\end{figure}

\begin{defn}
A \textbf{knowledge graph}, defined as $\mathcal{G}_{kn} = (V, E)$, is an instance of a directed heterogeneous graph whose nodes are \textit{entities}, and edges are subject-property-object triplets. Each edge has the form: head entity, relation, tail entity, denoted as $<h,r,t>$, indicating a relationship from a head $h$ to a tail $t$. 
\end{defn}
 $h, t \in V$ are entities and $r \in E$ is the relation. Hereby, we note $<h, r, t>$ as a triplet for knowledge graph. An example is shown in Fig.~\ref{fig:knoweldge_graph}. The knowledge graph includes two triplets $<h, r, t>$: $< Los Angeles, IsCityOf, California>$ and $<California, isStateOf, US>$. 

\begin{table}[t!] 
 \centering
 \caption{A summary of notations used in Sec.~\ref{kd_graph}.}
 \begin{tabular}[width=\columnwidth]{l|c}
 \hline 
  Notations & Descriptions\\
\hline\hline
 $|\cdot|$ & The cardinally of a set \\ \hline
 $\mathcal{G}= (V,E)$ & Graph $\mathcal{G}$ with a set of node $V$ and set of edge $E$ \\ \hline

 $v_i$, $e_{ij}$ & A node $v_i \in V$ and an edge $e_{ij}$ linking $v_i$ and $v_j$ \\ \hline
  $\textbf{x}_{v_i}$, $\textbf{x}_{e[v_i]}$ & Features of $v_i$ and features of edges of $v_i$  \\ \hline
  $\textbf{h}_{ne[v_i]}$, $\textbf{x}_{ne[v_i]}$ & Features of states and of neighboring nodes of $v_i$ \\ \hline
 $F_v(v_i)$, $F_e(e_{ij})$ & Mapping of node type $v_i$ and edge type $e_{ij}$ \\ \hline
 $\mathcal{T}^v$, $\mathcal{T}^e$ & The set of node types and set of edge types \\ \hline
 $<h, r, t>$ & Head, relation and tail in knowledge graph  \\ \hline
 $N$ & Number of nodes in the graph  \\ \hline
 $\textbf{h}_{v_i}$ & Hidden state of $i$-th node $v$  \\ \hline
 $f_t$, $f_o$ & local transition and output functions  \\ \hline
 $F_t$, $F_o$ & global transition and output functions  \\ \hline
 $\textbf{H}$, $\textbf{O}$, $\textbf{X}$ & Stack of all hidden states, outputs, features \\ \hline
 $\textbf{H}^{t}$ & Hidden state of $t$-th iteration of $\textbf{H}$  \\ \hline

\hline
 \end{tabular}
\label{tab:comp_table1}
\end{table}

\noindent \textbf{Graph neural networks.} A graph neural network (GNN) is a type of DNN that operates directly on the graph structure. A typical application is about node classification \cite{scarselli2008graph}. In the node classification problem, the $i$-th node $v_i$ is characterized by its feature $x_{v_i}$, and ground truth $t_{v_i}$. Thus, given a labeled graph $\mathcal{G}$, the goal is to leverage the labeled nodes to predict the unlabeled ones. It learns to represent each node with a $d$ dimensional vector state $h_{v_i}$ containing the information of its neighborhood. Specifically speaking, $h_{v_i}$ can be mathematically described as \cite{zhou2018graph}: 
\begin{equation}
    \textbf{h}_{v_i} = f_t(\textbf{x}_{v_i}, \textbf{x}_{co[v_i]}, \textbf{h}_{ne[v_i]}, \textbf{x}_{ne[v_i]})
\end{equation}
\begin{equation}
    \textbf{o}_{v_i} = f_o(\textbf{h}_{v_i}, \textbf{x}_{v_i})
\end{equation}
where $\textbf{x}_{co[v_i]}$ denotes the feature of the edges connected with $v_i$, $\textbf{h}_{ne[v_i]}$ denotes the embedding of the neighboring nodes of $\textbf{v}_i$, and $\textbf{x}_{ne[v_i]}$ denotes the features of the neighboring nodes of $v_i$. The function $f_t$ is a transition function that projects these inputs onto a $d$-dimensional space, and $f_o$ is the local output function that produces the output. Note that $f_t$ and $f_o$ can be interpreted as the feedforward neural networks. If we denote $\textbf{H}$, $\textbf{O}$, $\textbf{X}$ and $\textbf{X}_N$ as the concatenation of the outputs of stacking all the states, all the outputs, all the features, and all the node features, respectively, then $\textbf{H}$ and $\textbf{O}$ can be formulated as: 
\begin{equation}
    \textbf{H}= F_t(\textbf{H}, \textbf{X})
    \label{global_general}
\end{equation}
\begin{equation}
    \textbf{O}= F_o(\textbf{H}, \textbf{X})
\end{equation}
where $F_t$ is the global transition function and $F_o$ is the global output function. Note that $F_t$ and $F_o$ are the stacked functions of $f_t$ and $f_o$, respectively, in all nodes $V$ in the graph. 

Since we are aiming to get a unique solution for $\textbf{h}_{v_i}$, in \cite{scarselli2008graph, zhou2018graph}, a neighborhood aggregation algorithm is applied, such that:

\begin{equation}
    \textbf{H}^{t+1}= F_t(\textbf{H}^t, \textbf{X})
    \label{globale_state}
\end{equation}
where $\textbf{H}^t$ denotes $t$-th iteration of $\textbf{H}$. 

Given any initial state $\textbf{H}(0)$, $\textbf{H}^{t+1}$ in Eqn.~\ref{globale_state} convergences exponentially to the solution in Eqn.~\ref{global_general}. Based on the framework, $f_t$ and $f_o$ can be optimized via supervised loss when the target information $t_v^i$ is known:
\begin{equation}
    \mathcal{L} = \sum_{i=1}^N(t_v^i - o_v^i)
\end{equation}
where $N$ is the total number of supervised nodes in the graph.

\begin{figure*}[t!]
    \centering
    \includegraphics[width=\textwidth]{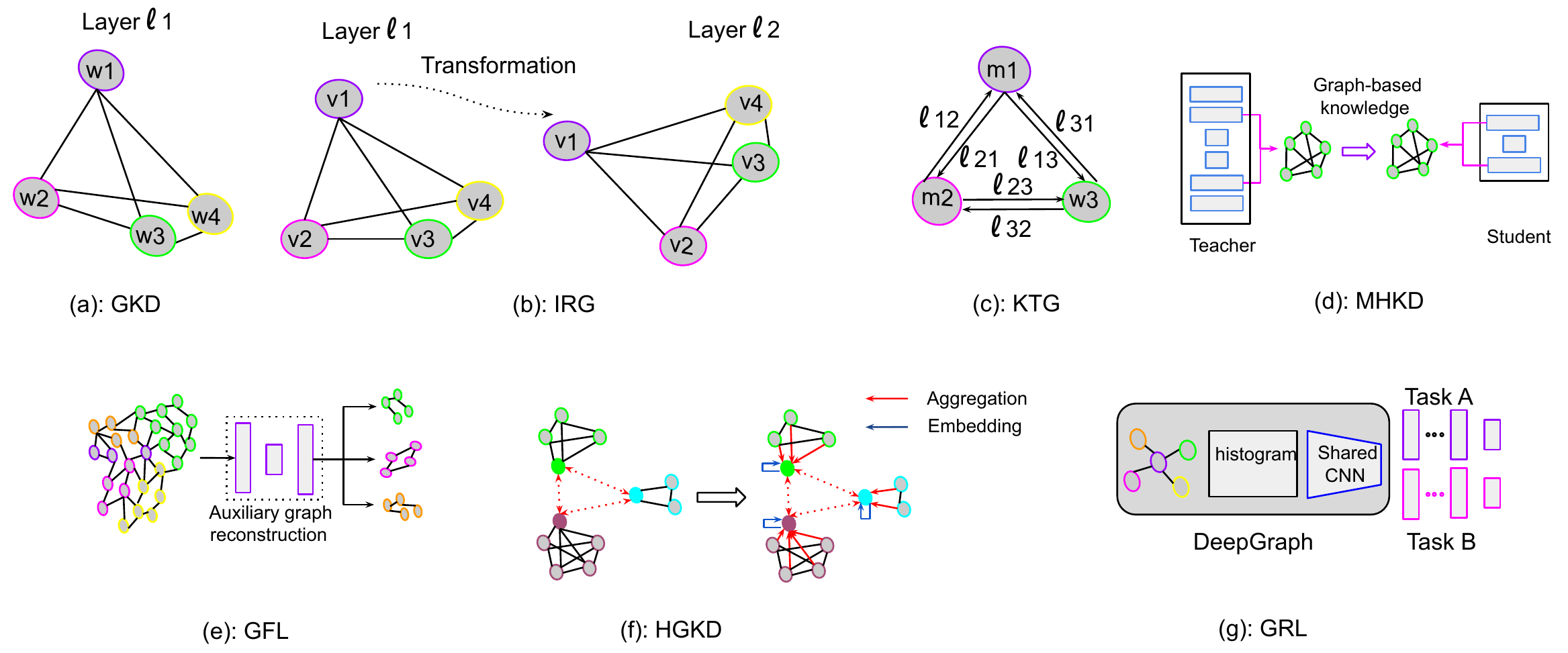}
    \vspace{-15pt}
    \caption{A graphical illustration of graph-based KD methods. GKD \cite{lassance2019deep}, IRG \cite{liu2019knowledgegraph}, KTG \cite{minami2019knowledge}, MHKD \cite{lee2019graph} all focus on graph-based knowledge distillation for model compression. GFL \cite{yao2019graph} and HGKD \cite{wang2019heterogeneous} aim to improve semi-supervised node classification via graph-based knowledge transfer, whereas GRL \cite{ma2019graph} exploits graph-based knowledge for multi-task learning.}
    \label{fig:gkd}
\end{figure*}

\subsubsection{Graph-based distillation}
Based on the above explanation regarding the fundamentals of graph representations and GNN, we now delve into the existing graph-based distillation techniques. To our knowledge, Liu \etal ~\cite{liu2011cross} first introduced a graph-modeling approach for visual recognition task in videos. Videos are action models modeled initially as bag of visual words (BoVW), which is sensitive to visual changes. However, some higher-level features are shared across views, and enable connecting the action models of different views. To better capture the relationship between two vocabularies, they construct a bipartite graph $\mathcal{G} = (V, E)$ to partition them into visual-word clusters. Note that $V$ is the union of vocabularies $V_1$ and $V_2$, and $E$ are the weights attached to nodes. In this way, knowledge from BoVW can be transferred to visual-word clusters, which are more discriminative in the presence of view changes. Luo \etal ~\cite{luo2018graph} consider incorporating rich, privileged information from a large-scale multimodal dataset in the source domain, and improve the learning in the target domain where training data and modalities are scarce. Regarding using S-T structures for KD, to date, there are several works such as \cite{liu2019knowledgegraph,lee2019graph,ma2019graph, lassance2019deep, wang2019heterogeneous,minami2019knowledge, yao2019graph, yang2020distillating}.

GKD \cite{lassance2019deep} and IRG \cite{liu2019knowledgegraph} consider the geometry of the perspective feature spaces by reducing intra-class variations, which allow for dimension-agnostic transfer of knowledge. This perspective is the opposite of Liu \etal ~\cite{liu2019knowledgegraph} and RKD \cite{park2019relational}. Specifically, instead of directly exploring the mutual relation between data points in students and teachers, GKD \cite{lassance2019deep} regards this relation as a geometry of data space (see Fig.~\ref{fig:gkd}(a)). Given a batch of inputs $\textbf{X}$, we can compute the inner representation $\textbf{X}_l^S=[\textbf{x}_l^S, \textbf{x} \in \textbf{X}$ and $\textbf{X}_l^T = [\textbf{x}_l^T, \textbf{x} \in \textbf{X}$ at layer $l$ ($l \in \Lambda$) of the teacher and student networks. Using cosine similarity metric, these representations can be used to build a $k$-nearest neighbor similarity graph for the teacher $\mathcal{G}_l^T(\textbf{X})=<\textbf{X}_l^T, \textbf{W}_l^T>$, and for the student $\mathcal{G}_l^S(\textbf{X})=<\textbf{X}_l^S, \textbf{W}_l^S>$. Note that $\textbf{W}_l^T$ and $\textbf{W}_l^S$ represent the edge weights, which represent the similarity between the $i$-th and $j$-th elements of $\textbf{X}_l^T$ and $\textbf{X}_l^S$. Based on graph representation for both the teacher and the student, the KD loss in Eqn.~\ref{fea_dis_loss} can be updated as follows: 
\begin{equation}
    \mathcal{L}= \sum_{l \in \Lambda} D\left(\mathcal{G}_l^S(\textbf{X}), \mathcal{G}_l^T(\textbf{X})\right)
\label{GKD_loss}
\end{equation}
where the distance metric $D$ is based on $L_2$ distance.

\begin{table*}[t!]
\caption{A summary of KD methods via graph representations. 
}
\vspace{-17pt}
\small
\begin{center}
\begin{tabular}[width=\textwidth]{c|c|c|c|c|c}
\hline
 Method & Purpose & Graph type & Knowledge type & Distance metric & Graph embedding   \\
\hline\hline
GKD \cite{lassance2019deep} &Model compression& Heterogeneous graph & Layer-wise feature & $L_2$ & GSP \cite{ortega2018graph}  \\ \hline
IRG \cite{liu2019knowledgegraph} &Model compression& Knowledge graph & Middle layers& $L_2$ & Instance relations\\ \hline
MHKD \cite{lee2019graph} &Model compression & Knowledge graph & Middle layers & KL & SVD \cite{lee2018self} + Attention  \\ \hline
KTG \cite{minami2019knowledge} & Model compression & Directed graph  & Network model & $L_1$ + KP loss & -- \\ \hline
GFL \cite{yao2019graph} &Few-shot learning& GNN & Class of nodes & Frobenius norm & HGR \cite{ying2018hierarchical}  \\ \hline
HGKD \cite{wang2019heterogeneous} &Few-shot learning&GNN & Class of nodes& Wasserstein& GraphSAGE \cite{hamilton2017inductive}  \\ \hline
GRL \cite{ma2019graph} & Multi-task leaning & GNN & Class of nodes & Cross-entropy & HKS \cite{li2016deepgraph} \\ \hline
Yang \cite{yang2020distillating} & Model compression & GNN & Topological info. & KL & Attention \\ \hline

\hline
\end{tabular}
\end{center}
\label{table:fea_comp}
\end{table*}
IRG \cite{liu2019knowledgegraph} essentially is similar to GKD \cite{lassance2019deep} in the construction of the graph, however, IRG also takes into account the instance of graph transformations. The aim of introducing feature space transformation across layers is because there may be too tight or dense constraints and descriptions fitting on the teacher’s instance features at intermediate layers. The transformation of the instance relation graph is composed of vertex transformation and edge transformation
from the $l_1l$-th layer to $l_2$-th layer, as shown in Fig.~\ref{fig:gkd} (b). Thus the loss in Eqn.~\ref{GKD_loss} can be extended to:
\begin{equation}
\begin{split}
      \mathcal{L}= \sum_{l \in \Lambda} D_1(\mathcal{G}_l^S(\textbf{X}), \mathcal{G}_l^T(\textbf{X})) +  \\ 
      D_2 \left((\Theta_T(\mathcal{G}_l^S(\textbf{X})), \Theta_S(\mathcal{G}_l^T(\textbf{X}))\right)
\end{split}
\end{equation}
where $\Theta_T$ and $\Theta_S$ are the transformation functions for the teacher and the student, respectively, and $D_1$ and $D_2$ are the distance metrics for instance relation and instance translation.
 
MHKD \cite{lee2019graph} is a method that enables distilling data based knowledge from a teacher network to a graph using an attention network (see Fig.~\ref{fig:gkd}(d)). Like IRG \cite{liu2019knowledge}, feature transformation is also considered to capture the intra-data relations. The KD loss is based on the KL-divergence loss using the embedded graphs from the teacher and the student. KTG \cite{minami2019knowledge} also exploits graph representation; however, it focuses on a different perspective of KD. The knowledge transfer graph provides a unified view of KD, and has the potential to represent diverse knowledge patterns. Interestingly, each node in the graph represents the direction of knowledge transfer. On each edge, a loss function is defined for transferring knowledge between two nodes linked by each edge. Thus, combining different loss functions can represent collaborative knowledge learning with pair-wise knowledge transfer. Fig.~\ref{fig:gkd}(c) shows the knowledge graph of diverse collaborative distillation with three nodes, where $L_{s,t}$ represents the loss function used for the training node.

In addition, GFL \cite{yao2019graph}, HGKT \cite{wang2019heterogeneous}, GRL \cite{ma2019graph} and MHGD \cite{lee2019graph} all resort to GNN for the purpose of KD. HGKT and GFL focus on transferring knowledge from seen classes to unseen classes in few-shot learning \cite{kipf2016semi,sung2018learning}. GFL \cite{yao2019graph} leverages the knowledge learned by the auxiliary graphs to improve semi-supervised node classification in the target graph. As shown in Fig.~\ref{fig:gkd}(e), GFL learns the representation of a whole graph, and ensures the transfer of a similarly structured knowledge. Auxiliary graph reconstruction is achieved by using a graph autoencoder. HGTK aims to build a heterogeneous graph focusing on transferring intra-class and inter-class knowledge simultaneously. Inspired by modeling class distribution in adversarial learning \cite{goodfellow2014generative, wang2020deceiving, wang2018kdgan, heo2019knowledge, xu2017training, haidar2019textkd}, in which instances with the same class are expected to have the same distribution, the knowledge is transferred from seen classes to new unseen classes based on learned aggregation and embedding functions, as shown in Fig.~\ref{fig:gkd} (f). GRL \cite{ma2019graph} builds a multi-task KD method for representation learning based on DeepGraph \cite{li2016deepgraph}.This knowledge is based on GNN, and maps raw graphs to metric values. The learned graph metrics are then used as auxiliary tasks, and the knowledge of the network is distilled into graph representations (see Fig.~\ref{fig:gkd}(g)). The graph representation structure is learned via a CNN by feeding the graph descriptor to it. We denote pairs of graph and graph-level labels as $\{(G_i, y _i)\}_{i=1}^N$, where $G_i \in \mathcal{G}$, $y_i \in \mathcal{Y}$, and $\mathcal{G}$, $\mathcal{Y}$ are the cardinally of all possible graphs and labels respectively. Then, the loss for learning the model parameters are described as:
\begin{equation}
    \mathcal{L} = \E [D(y_i, f(G_i;\theta))]
\end{equation}
where $\theta$ are the model parameters.

\noindent \textbf{Open challenges}
Graph representations are of significant importance for tackling KD problems because they better capture the hierarchical patterns in locally connected structures. However, there are some challenges. Firstly, graph representations are difficult to generalize because they are limited to structured data or specific types of data. Secondly, it is challenging to measure graph distances appropriately, since existing distance measure (\eg, $l_2$) may not fit well. Thirdly, layer-wise distillation is difficult to achieve in graph KD, because graph representation models and network structures in such cases are limited.

\subsection {KD for semi-/self-supervised learning}
\textbf{Overall insight:} \textit{KD with S-T learning aims to learn a rich representation by training a model with a large number of unlabeled datasets, and limited amount of labeled data.}

Semi-supervised learning usually handles the problem of over-fitting due to the lack of high-quality labels of training data. To this end, most methods apply S-T learning that assumes a dual role as a teacher and a student. The student model aims to learn the given data as before, and the teacher learns from the noisy data and generates predicted targets. These are then transferred to the student model via consistency cost. In self-supervised learning, the student itself generates knowledge to be learned via various approaches, and the knowledge is then transferred by the student to itself via distillation losses. We now provide a detailed analysis of the technical details of the existing methods. 

\subsubsection{Semi-supervised learning} The baseline S-T frameworks for semi-supervised learning was initialized by Laine \etal ~\cite{laine2016temporal} and Tarvainen \etal ~\cite{tarvainen2017mean}, as illustrated in Fig.~\ref{fig:overview_fig}(b).
The student and the teacher models have the same structures, and the teacher learns from noise and transfers knowledge to the student via consistency cost. Interestingly, in \cite{tarvainen2017mean}, the teacher's weights are updated using the earth moving average (EMA) of the student's weights. Inspired by \cite{tarvainen2017mean}, Luo \etal ~\cite{luo2018smooth}, Zhang \etal ~\cite{zhang2019pairwise}, French \etal ~\cite{french2017self}, Choi \etal ~\cite{choi2019self}, Cai \etal ~\cite{cai2019exploring} and Xu \etal ~\cite{xu2019self} all employ similar frameworks where the teacher's weights are updated using exponential moving average (EMA) of the student. However, Ke \etal \cite{ke2019dual} mention that using a coupled EMA teacher is not sufficient for the student, since the degree of coupling increases as the training goes on. To tackle this problem, the teacher is replaced with another student, and two students are optimized individually during training while a stabilization constraint is provided for knowledge exchange (similar to mutual learning \cite{zhang2018deep}). 

Instead of taking independent weights between the teacher and the student, Hailat \etal ~\cite{hailat2018teacher} employ weight-sharing, in which the last two fully connected layers of the teacher and the student are kept independent. The teacher model plays the role of teaching the student, stabilizing the overall model, and attempting to clean the noisy labels in the training dataset. In contrast, Gong \etal ~\cite{gong2018teaching} and Xie \etal ~\cite{xie2019self} follow the conventional distillation strategy proposed by \cite{hinton2015distilling}, where a pretrained teacher is introduced to generate learnable knowledge using unlabeled data, and utilizes it as privileged knowledge to teach the student on labeled data. However, during learning of the student, Xie \etal \ inject noise (\eg, dropout) to the student such that it learns better than the teacher. 
Papernot \etal ~\cite{papernot2016semi} propose to distill from multiple teachers (an ensemble of teachers) on a disjoint subset of sensitive data (augmented with noise) and to aggregate the knowledge of teachers to guide the student on query data. 

\subsubsection{Self-supervised learning} Distilling knowledge for self-supervised learning aims to preserve the learned representation for the student itself, as depicted in Fig.~\ref{fig:overview_fig}(c). Using pseudo labels is the most common approach, as done in \cite{lee2018self,noroozi2018boosting}. Specifically, Lee \etal ~\cite{lee2018self} adopt self-supervised learning for KD, which not only ensures the transferred knowledge does not vanish, but also provides an additional performance improvement. In contrast, Noroozi \etal ~\cite{noroozi2018boosting} propose to transfer knowledge by reducing the learned representation (from a pretrained teacher model) to \textit{pseudo-labels} (via clustering) on the unlabeled dataset,
which are then utilized to learn a smaller student network. Another approach is based on data augmentation (\eg, rotation, cropping, color permutation) \cite{lee2019rethinking, xu2019data,xu2020knowledge}, which has been mentioned in Sec.~\ref{dis_dataaug}.  In contrast to making the `positive' and `negative' (augmented) examples, BYOL \cite{grill2020bootstrap} directly bootstraps the representations with two neural networks, referred to as online and target networks, that interact and learn from each other. This spirit is somehow similar to mutual learning \cite{zhang2018deep}; however, BYOL trains its online network to predict the target network’s representation of another augmented view of the same image. The promising performance of BYOL might point out a new direction of  KD with self-supervised learning via representation boostrap rather than using negative examples.

\subsubsection{Potentials and open challenges}
Based on the technical analysis for the KD methods in semi-/self-supervised learning, it is noticeable that online distillation is the mainstream. However, there are several challenges. First, as pointed by \cite{ke2019dual}, using EMA for updating teacher's weights might lead to less optimal learning of knowledge. Second, no methods attempt to exploit the rich feature knowledge from teacher models. Third, data augmentation methods in these distillation methods are less effective compared to those proposed in Sec.~\ref{self_distillation}, in which the advantages of adversarial learning are distinctive. Fourth, the representations of knowledge in these methods are limited and less effective. BYOL \cite{grill2020bootstrap} opens a door for representation boost, and there exists a potential to further bind this idea with KD in the further research. Moreover, it has the potential to exploit a better-structured data representation approach, such as in GNNs. 
With these challenges, the future directions of KD for semi-/self-supervised learning could gain inspirations from exploiting feature knowledge and more sophisticated data augmentation methods together with more robust representation approaches.

\subsection{Few-shot learning}
\textbf{Insight:}\textit{Is it possible possible to learn an effective student model to classify unseen classes (query sets) by distilling knowledge from a teacher model with the support set?}

In contrast to the methods discussed in Sec.~\ref{distill_fewdata} focusing on distillation with a few samples for training a student network (without learning to generalize to new classes), this section stresses on analyzing the technical details of few-shot learning with KD. Few-shot learning is to classify new data, having seen from only a few training examples. Few-shot learning itself is a meta-learning problem in which the DNN learns how to learn to classify, given a set of training tasks, and evaluate using a set of test tasks. Here, the goal is to discriminate between $N$ classes with $K$ examples of each (so-called $N$-way-$K$-shot classification). In this setting, these training examples are known as the \textit{support set}. In addition, there are further examples of the same classes, known as a \textit{query set}. The approaches for learning prior knowledge of a few-shot are usually based on \textit{three} types: prior knowledge about similarity, prior knowledge about learning procedure, and prior knowledge about data. 
We now analyze the KD methods for few-shot learning \cite{dvornik2019diversity,flennerhag2018transferring,park2019relational,liu2019semantic,jin2019learning} that have been recently proposed.

\noindent \textbf{Prior knowledge about similarity:} Park \etal     ~\cite{park2019relational} propose distance-wise and angle-wise distillation losses. The aim is to penalize the structural differences in relation to the learned representations between the teacher and the student for few-shot learning.

\noindent \textbf{Prior knowledge about learning procedure:} \cite{flennerhag2018transferring,jin2019learning} tackle the second type of prior knowledge, namely learning procedure. To be specific, Flennerhag \etal ~\cite{flennerhag2018transferring} focuses on transferring knowledge across the learning process, in which the information from previous tasks is distilled to facilitate the learning on new tasks. However, Jin \etal ~\cite{jin2019learning} address the problem of learning a meta-learner that can automatically learn what knowledge to transfer from the source network to where in the target network.

\noindent \textbf{Prior knowledge about data:} Dvornik \etal ~\cite{dvornik2019diversity} and Liu \etal ~\cite{liu2019semantic} address the third type of prior knowledge, namely data variance. To be specific, in \cite{dvornik2019diversity}, en ensemble of several teacher networks is elaborated to leverage the variance of the classifiers and encouraged to cooperate while encouraging the diversity of prediction. However, in \cite{liu2019semantic}, the goal is to preserve the knowledge of the teacher (\eg, intra-class relationship) learned at the pretraining stage by generating pseudo labels for training samples in the fine-tuning set. 

\subsubsection{What's challenging?}
Based on our analysis, the existing techniques actually expose crucial challenges. First, the overall performance of KD-based few-shot learning is convincing, but the power of meta-learning is somehow degraded or exempted. Second, transferring knowledge from multi-source networks is a potential, but identifying what to learn and where to transfer is heavily based on the meta-learner, and selecting which teacher to learn is computation-complex. Third, all approaches focus on a task-specific distillation, but the performance drops as the domain shifts. Thus, future works may focus more on handling these problems.

\subsection{Incremental Learning}
\textbf{Overall insight:} \textit{KD for incremental learning mainly deals with two challenges: maintaining the performance on old classes, and balancing between old and new classes.} 

Incremental learning investigates learning the new knowledge continuously to update the model's knowledge, while maintaining the existing knowledge \cite{wu2019large}. Many attempts \cite{hou2019learning, michieli2019knowledge,castro2018end,wu2019large, zhou2019m2kd,shmelkov2017incremental, zhai2019lifelong} have been made to utilize KD in addressing the challenge of maintaining the old knowledge. Based on the number of teacher networks used for distillation, these methods can be categorized into two types: distillation from a single teacher and distillation from multiple teachers.

\subsubsection{Distillation from a single teacher}
Shmelkov \etal ~\cite{shmelkov2017incremental}, Wu \etal ~\cite{wu2019large}, Michieli \etal ~\cite{michieli2019knowledge} and Hou \etal ~\cite{hou2019learning} focus on learning student networks for new classes, by distilling knowledge (logits information) from pretrained teachers on old-class data. 
Although these methods vary in tasks and distillation process, they follow similar S-T structures. Usually, the pretrained model is taken as the teacher, and the same network or a different network is employed to adapt for new classes. Michieli \etal \ exploit the intermediate feature representations and transfer them to the student.    

\subsubsection{Distillation from multiple teachers}
Castro \etal ~\cite{castro2018end}, Zhou \etal ~\cite{zhou2019m2kd} and Ammar \etal ~\cite{ammar2015autonomous} concentrate on learning an incremental model with multiple teachers. Specifically, Castro \etal \ share the same feature extractor between teachers and the student. The teachers contain old classes, and their logits are used for distillation and classification. Interestingly, Zhou \etal \ propose a multi-model and multi-level KD strategy in which all previous model snapshots are leveraged to learn the last model (student). This approach is similar to born-again KD methods, as mentioned in Sec. \ref{self_distillation}, where the student model at the last step, and is updated using the assembled knowledge from all previous steps. However, the assembled knowledge also depends on the intermediate feature representations. Ammar \etal \ develop a cross-domain incremental RL framework, in which the transferable knowledge is shared and projected to different task domains of the task-specific student peers.

\subsubsection{Open challenges}
The existing methods rely on multi-step training (offline). However, it will be more significant if the online (one-step) distillation approaches can be utilized to improve the learning efficiency and performance. Moreover, existing methods require accessing the previous data to avoid ambiguities between the update steps. However, the possibility of data-free distillation methods remains open. Furthermore, existing methods only tackle the incremental learning of new classes in the same data domain, but it will be fruitful if cross-domain distillation methods can be applied in this direction.

\subsection{Reinforcement learning}
\textbf{Overall insight:} \textit{KD in reinforcement learning is to encourage polices (such as students) in the ensemble to learn from the best policies (such as teachers), thus enabling rapid improvement and continuous optimization.}

Reinforcement learning (RL) is a learning problem that trains a policy to interact with the environment in way that yields maximal reward. To use the best policy to guide other policies, KD has been employed in  \cite{ashok2017n2n,liu2019knowledge,flennerhag2018transferring,burda2018exploration,hong2020periodic,xue2020transfer,lin2017collaborative, rusu2015policy}. Based on the specialties of these methods, we divide them into three categories, and provide an explicit analysis. \textit{We assume familiarity with basics of RL, and skip the definitions of Deep Q-network and A3C.}

\subsubsection{Collaborative distillation}
Xue \etal ~\cite{xue2020transfer}, Hong \etal ~\cite{hong2020periodic}, and Lin \etal ~\cite{lin2017collaborative} focus on collaborative distillation, which is similar to mutual learning \cite{zhang2018deep}. In Xue \etal, the agents teach each other based on the reinforcement rule, and teaching occurs between the value function of the agents (students and teachers). Note that the knowledge is provided by a group of student peers periodically, and assembled to enhance the learning speed and stability as in \cite{hong2020periodic}. However, Hong \etal \cite{hong2020periodic} periodically distill the best-performing policy to the rest of the ensemble. Lin \etal \ stress on collaborative learning among heterogeneous learning agents, and incorporate the knowledge into online training. 

\subsubsection{Model compression with RL-based distillation}
Ashok \etal ~\cite{ashok2017n2n} tackle the problem of model compression via RL. The method takes a larger teacher network, and outputs a compressed student network derived from it. In particular, two recurrent policy networks are employed to aggressively remove layers from the teacher network, and to carefully reduce the size of each remaining layer. The learned student network is evaluated by a reward, which is a score based on the accuracy and compression of the teacher. 

\subsubsection{Random network distillation}
Burda \etal ~\cite{burda2018exploration} focus on a different perspective where the prediction problem is randomly generated. The approach involves two networks: the target (student) network, which is fixed and randomly initialized, and a predictor (teacher) network trained on the data collected by the agent. With the knowledge distilled from the predictor, the target network tends to have lower prediction errors. Rusu \etal ~\cite{rusu2015policy} also apply random initialization for the target network. However, they focus more on online learning action policies, which can be either single-task or muti-task.  

\subsubsection{Potentials of RL-based KD}
We have analyzed existing RL-based KD methods in detail. Especially, we notice that model compression via RL-based KD is promising due to its extraordinary merits. First, RL-based KD better addresses the problem of \textit{scalability} of network models. This is similar to neural architecture search (NAS). Moreover, the reward functions in RL-based KD \textit{better balance} the accuracy-size trade-off. It is also possible to transfer knowledge from a \textit{smaller model to a larger model}, which is a distinctive advantage over other KD methods. 

\section{Applications for visual intelligence}
\subsection{Semantic and motion segmentation}
\textbf{Insight:} \textit{Semantic segmentation is a structured problem, and structure information (\eg, spatial context structures) needs to be taken into account when distilling knowledge for semantic segmentation networks.}

Semantic segmentation is a special classification problem that predicts the category label in a pixel-wise manner. As existing the state-of-the-art (SOTA) methods such as fully convolutional
networks \cite{long2015fully} have large model sizes and high computation costs, some methods 
\cite{dou2020unpaired, liu2019structured, michieli2019knowledge, shan2019distilling, he2019knowledge,chen2018road,xie2018improving, fang2019data, mullapudi2019online} have been proposed to train lightweight networks via KD. Although these methods vary in their learning methods, most of them share the same distillation frameworks. Particularly, Xie \etal ~\cite{xie2018improving}, Shan \etal ~\cite{shan2019distilling}, and Michieli \etal ~\cite{michieli2019knowledge} focused on pixel-wise, feature-based distillation methods. Moreover, Liu ~\etal \cite{liu2019structured} and He \etal ~\cite{he2019knowledge} both exploited affinity-based distillation strategy using intermediate features. Liu \etal~ also employed pixel-wise and holistic KD losses via adversarial learning. In contrast, Dou \etal ~\cite{dou2020unpaired} focused on unpaired multi-modal segmentation, and proposed an online KD method via mutual learning \cite{zhang2018deep}. Chen \etal ~\cite{chen2018road} proposed a target-guided KD approach to learn the real image style by training the student to imitate a teacher trained with real images. Mullapudi \etal ~\cite{mullapudi2019online} trained a compact video segmentation model via online distillation, in which a teacher's output was used as a learning target to adapt the student and select the next frame for supervision. 

\subsection{KD for visual detection and tracking}
\textbf{Insight:} \textit{Challenges such as regression, region proposals, and less voluminous labels must be considered when distilling visual detectors.}

Visual detection is a crucial high-level task in computer vision. Speed and accuracy are two key factors for visual detectors. KD is a potential choice to achieve sped-up and lightweight network models. However, applying distillation methods to detection is more challenging than applying classification methods. First, detection performance degrades seriously after compression. Second, detection classes are not equally important, and special considerations for distillation have to be taken into account. Third, domain and data generalization has to be considered for a distilled detector. To overcome these challenges, several impressive KD methods \cite{chen2017learning,wang2019distilling,hong2019gan,hao2019end,chen2019new,chen2019learning,tang2019learning,jin2019learning,saputra2019distilling,jin2020uncertainty,lee2019teaching,xu2019training,kruthiventi2017low,ge2018low,luo2016face,feng2019triplet} have been proposed for compressing visual detection networks. We categorize these methods according to their specialties (\eg, pedestrian detection). 
 
\subsubsection{Generic object detection}  \cite{chen2017learning,hong2019gan,chen2019new,hao2019end,jin2020uncertainty,tang2019learning,wang2019distilling, felix2020squeezed, liu2019teacher} aimed to learn lightweight object detectors with KD. Among these works, Chen \etal ~\cite{chen2019new} and Hao \etal ~\cite{hao2019end} highlighted learning a class-incremental student detector by following the generic KD framework (from a pretrained teacher). However, novel object detection losses were adopted as strong impetus for learning new classes. These losses handled classification results, location results, the detected region of interest, and all intermediate region proposals. Moreover, Chen \etal ~\cite{chen2017learning} learned a student detector by distilling knowledge from the intermediate layer, logits, and regressor of the teacher, in contrast to \cite{wang2019distilling}, in which only the intermediate layer of the teacher was utilized based on fine-grained imitation masks to identify informative locations. Jin \etal  ~\cite{jin2020uncertainty}, Tang \etal ~\cite{tang2019learning}, and Hong \etal \cite{hong2019gan} exploited multiple intermediate layers as useful knowledge. Jin \etal~ designed an uncertainty-aware distillation loss to learn the multiple-shot features from the teacher network. However, Hong \etal and Tang \etal \ were based on one-stage KD (online) via adversarial learning and semi-supervised learning, respectively. In contrast, Liu \etal ~\cite{liu2019teacher} combined single S-T learning and mutual learning of students for learning lightweight tracking networks.  

\subsubsection{Pedestrian detection} While pedestrian detection is based on generic object detection, various sizes and aspect ratios of pedestrians under extreme illumination conditions are challenges. To learn an effective lightweight detector, Chen \etal ~\cite{chen2019learning} suggested using the unified hierarchical knowledge via multiple intermediate supervisions, in which not only the feature pyramid (from low-level to high-level features) and region features, but also the logits information were distilled. Kruthiventi \etal ~\cite{kruthiventi2017low} learned an effective student detector in challenging illumination conditions by extracting dark knowledge (both RGB and thermal-like hint features) from a multi-modal teacher network.

\subsubsection{Face detection} Ge \etal ~\cite{ge2018low} and Karlekar \etal ~\cite{karlekar2019deep} compressed face detectors to recognize low-resolution faces via selective KD (last hidden layer) from teachers which were initialized to recognize high-resolution faces. In contrast, Jin \etal ~\cite{jin2019learning}, Luo \etal ~\cite{luo2016face}, and Feng \etal ~\cite{feng2019triplet} used single type of image. Jin \etal focused on compressing face detectors by using the supervisory signal from the classification maps of teacher models and regression maps of the ground truth. They identify that it is better to learn a classification map of a larger model than that of smaller models. Feng \etal presented a triplet KD method to transfer knowledge from a teacher model to a student model, in which a triplet of samples, the anchor image, positive image, and negative image, was used. The purpose of the triplet loss was to minimize the feature similarity between the anchor and positive images, while maximizing that between the anchor and negative images. Luo \etal addressed the importance of neurons at the higher hidden layer of the teacher, and a neuron selection method was applied to select neurons that were crucial for teaching the student. Dong \etal ~\cite{dong2019teacher} concentrated on the interaction between the teacher and the students. Two students learned to generate pseudo facial landmark labels, which were filtered and selected as the qualified knowledge by the teacher. 

\subsubsection{Vehicle detection and driving learning} Lee \etal ~\cite{lee2019teaching}, Saputra \etal ~\cite{saputra2019distilling}, and Xu \etal ~\cite{xu2017training} focused more on detection tasks for autonomous driving. In particular, Lee \etal focused on compressing a vehicle maker classification system based on cascaded CNNs (teacher) into a single CNN structure (student). The proposed distillation method used the feature map as the transfer medium, and the teacher and student were trained in parallel (online distillation). Although the detection task was different, Xu \etal ~ build a binary weight Yolo vehicle detector by mincing the feature maps of the teacher network from easy tasks to difficult ones progressively.
Zhao \etal ~\cite{zhao2019lates} exploited an S-T framework to encourage the student to learn the teacher's sufficient and invariant representation knowledge (based on semantic segmentation) for driving. 

\subsubsection{Pose detection} 
Distilling human pose detectors has several challenges. First, lightweight detectors have to deal with arbitrary person images/videos to determine joint locations with unconstrained human appearances. Second, the detectors must be robust in viewing conditions and background noises. Third, the detectors should have fast inference speeds, and be memory-efficient. To this end, \cite{zhang2019fast,martinez2019efficient,thoker2019cross,hwang2020lightweight,xu2020integral,wang2019distill, nie2019dynamic} formulated various distillation methods. Zhang \etal ~\cite{zhang2019fast} achieved effective knowledge transfer by distilling the joint confidence maps from a pre-trained teacher model, whereas Huang \etal ~\cite{hwang2020lightweight} exploited the heat map and location map of a pretrained teacher as the knowledge to be distilled. Furthermore, Xu \etal ~\cite{xu2020integral}, Thoker \etal ~\cite{thoker2019cross}, and Martinez \etal ~\cite{martinez2019efficient} focused on multi-person pose estimation. Thoker \etal~ addressed cross-modality distillation problems, in which a novel framework based on mutual learning \cite{zhang2018deep} of two students supervised by one teacher was initialized. Xu \etal ~\cite{xu2020integral} learned the integral knowledge, namely, the feature, logits, and structured information via a discriminator under the standard S-T framework. Martinez \etal ~\cite{martinez2019efficient} trained the student to mimic the confidence maps, feature maps, and inner-stage predictions of a pre-trained teacher with depth images. Wang \etal ~\cite{wang2019distill} trained a 3D pose estimation network by distilling knowledge from non-rigid structure from motion using only 2D landmark annotations. In contrast, Nie \etal ~\cite{nie2019dynamic} introduced an online KD in which the pose kernels in videos were distilled by leveraging the temporal cues from the previous frame in a one-shot learning manner.  

\subsection{Domain adaptation} 
\textbf{Insight:} \textit{Is it possible to distill knowledge of a teacher in one domain to a student in another domain?}

Domain adaptation (DA) addresses the problem of learning a target domain with the help of a different but related source domain \cite{ao2017fast}. Since Lopez \etal ~\cite{lopez2015unifying} and Gupta \etal ~\cite{gupta2016cross} initially proposed the technique of transferring knowledge between images from different modalities (called generalized distillation), it is natural to ask if this novel technique be used to address the problem of DA. The challenge of DA usually comes with transferring knowledge from the source model (usually with labels) to the target domain with unlabeled data. To address the problem, several KD methods based on S-T frameworks \cite{ao2017fast,hoffman2017cycada,meng2019domain,xu2019self,choi2019self,chen2019crdoco,tsai2018learning,cai2019exploring,deng2019cluster} have been proposed recently. Although these methods are focused on diverse tasks, technically, they can be categorized into two types: unsupervised and semi-supervised DA via KD. 

\subsubsection{Semi-supervised DA} 
French \etal ~\cite{french2017self}, Choi \etal ~\cite{choi2019self}, Cai \etal ~\cite{cai2019exploring}, Xu \etal ~\cite{xu2019self}, and Cho \etal ~\cite{cho2019large} proposed similar S-T frameworks for semantic segmentation and object detection. These frameworks were the updated methods of Mean-Teacher \cite{tarvainen2017mean}, which is based on self-ensemble of the student networks (teacher and student models have the same structure). Note that the weights of the teacher models in these methods are the EMAs of the weights of the student models. In contrast, Choi \etal added a target-guided generator to produce augmented images instead of stochastic augmentation, as in \cite{xu2019self, french2017self,cai2019exploring}. Cai \etal \ also exploited the feature knowledge from the teacher model, and applied region-level and intra-graph consistency losses instead of the mean square error loss. 

In contrast, Ao \etal ~\cite{ao2017fast} proposed a generalized distillation DA method by applying the generalized distillation information \cite{lopez2015unifying} to multiple teachers to generate soft labels, which were then used to supervise the student model (this framework is similar to online KD from multiple teachers as mentioned in Sec.~\ref{multi_teach}). Cho \etal ~\cite{cho2019large} proposed an S-T learning framework, in which a smaller depth prediction network was trained based on the supervision of the auxiliary information (ensemble of multiple depth predictions) obtained from a larger stereo matching network (teacher). 

\subsubsection{Unsupervised DA} Some methods such as \cite{chen2019crdoco, hoffman2017cycada} distill the knowledge from the source domain to the target domain based on adversarial learning \cite{goodfellow2014generative} and image translation \cite{isola2017image, wang2020deceiving, wang2019event}. Technically, images in the source domain are translated to images in the target domain as data augmentation, and cross-domain consistency losses are adopted to force the teacher and student models to produce consistent predictions. Tsai \etal ~\cite{tsai2018learning} and Deng \etal ~\cite{deng2019cluster} focused on aligning the feature similarities between teacher and student models, compared with Meng \etal ~\cite{meng2019domain}, who focused on aligning softmax outputs. 

\subsection{Depth and scene flow estimation} 
\textbf{Insight:} \textit{The challenges for distilling depth and flow estimation tasks come with transferring the knowledge of data and labels.}

Depth and optical flow estimations are low-level vision tasks aiming to estimate the 3D structure and motion of the scene. There are several challenges. First, in contrast to other tasks (\eg, semantic segmentation), depth and flow estimations do not have class labels. Thus, applying existing KD techniques directly may not work well.  
Moreover, learning a lightweight student model usually requires a large amount of labeled data to achieve robust generalization capability. However, acquiring these data is very costly. 

To address these challenges, Guo \etal 
~\cite{guo2018learning}, Pilzer \etal ~\cite{pilzer2019refine}, and Tosi \etal ~\cite{tosi2019learning} proposed distillation-based approaches to learn monocular depth estimation. These methods were focused on handling the second challenge, namely, data distillation. Specifically, Pilzer \etal ~\cite{pilzer2019refine} proposed an unsupervised distillation approach, where the left image was translated to the right via the image translation framework \cite{isola2017image, wang2020deceiving}. The inconsistencies between the left and right images were used to improve depth estimation, which was finally used to improve the student network via KD. In contrast, Guo \etal~ and Tosi \etal~ focused on cross-domain KD, which aimed to distill the \textit{proxy} labels obtained from the stereo network (teacher) to learn a student depth estimation network. Choi \etal ~\cite{cho2019large} learned a student network for monocular depth inference by distilling the knowledge of depth predictions from a \textit{stereo} teacher network via the data ensemble strategy.

Liu \etal ~\cite{liu2019ddflow} and Aleotti \etal ~\cite{aleotti2019learning} proposed data-distillation methods for scene flow estimation. Liu \etal \ distilled reliable predictions from a teacher network with unlabeled data, and used these predictions (for non-occluded pixels) as annotations to guide a student network to learn the optical flow. They proposed to leverage on the knowledge learned by the teacher networks specialized in stereo to distill proxy annotations, which is similar to the KD method for depth estimation in \cite{guo2018learning, tosi2019learning}. Tosi \etal ~\cite{tosi2020distilled} learned a compact network for predicting holistic scene understanding tasks including depth, optical flow, and motion segmentation, based on distillation of proxy semantic labels and semantic-aware self-distillation of optical information. 

\subsection{Image translation}
\label{img_trans}
\textbf{Insight}:\textit{ Distilling GAN frameworks for image translation has to consider three factors: large number of parameters of the generators, no ground truth labels for training data, and complex framework (both generator and discriminator)}.

Attempts were made in several works to compress GANs for image translation with KD. Aguinaldo \etal ~\cite{aguinaldo2019compressing} focused on unconditional GANs, and proposed to learn a smaller student generator by distilling knowledge from the generated images of a larger teacher generator using mean squared error (MSE). However, the knowledge incorporated in the teacher discriminator was not investigated. In contrast, Chen \etal ~\cite{chen2020distilling} and Li \etal ~\cite{li2020gan} focused on conditional GANs, and exploited the knowledge from the teacher discriminator. Specifically, Chen \etal included a student discriminator to measure the distances between real images and images generated by the student and teacher generators. The student GAN was then trained under the supervision of the teacher GAN. In particular, Li \etal ~\cite{li2020gan} adopted the discriminator of the teacher as the student discriminator, and fine-tuned the discriminator together with the compressed generator, which was automatically found with significantly lower computation cost and fewer parameters, by using NAS. In contrast, Wang \etal ~\cite{wang2020collaborative} focused on compressing encoder-decoder based neural style transfer network via collaborative distillation (between the encoder and its decoder), where the student was restricted to learn the linear embedding of the teacher's output. 

\subsection{KD for Video understanding}
\subsubsection{Video classification and recognition}
Bhardwaj \etal ~\cite{bhardwaj2019efficient} and Wang \etal ~\cite{wang2019progressive} employed the general S-T learning framework for video classification. The student was trained with processing only a few frames of the video, and produced a representation similar to that of the teacher. Gan \etal ~\cite{gan2016you} focused on video concept learning for action recognition and event detection by using web videos and images. The learned knowledge from teacher network (Lead network) was used to filter out the noisy images. These were then used to fine-tune the teacher network to obtain a student network (Exceeding network). Gan \etal ~\cite{gan2018geometry} explored geometry as a new type of practical auxiliary knowledge for self-supervised learning of video representations. Fu \etal ~\cite{fu2019ultrafast}  propose focused on video attention prediction by leveraging both spatial and temporal knowledge. Farhadi \etal ~\cite{farhadi2019tkd} distill the temporal knowledge from a teacher model over the selected video frames to a student model. 

\subsubsection{Video captioning}
\cite{zhang2020object, pan2020spatio} exploited the potential of graph-based S-T learning for image captioning. Specifically, Zhang \etal ~\cite{zhang2020object} leveraged the object-level information (teacher) to learn the scene feature representation (student) via a spatio-temporal graph. Pan \etal ~\cite{pan2020spatio} highlighted the importance of the relational graph connecting for all the objects in the video, and forced the caption model to learn the abundant linguistic language via teacher-recommended learning. 

\section{Discussions}
In this section, we discuss some fundamental questions and challenges that are crucial for better understanding and improving KD. 

\subsection{Are bigger models better teachers?}
The early assumption and idea behind KD are that soft labels (probabilities) from a trained teacher reflect more about the distribution of data than the ground truth labels \cite{hinton2015distilling}. If this is true, then it is expected that as the teacher becomes more robust, the knowledge (soft labels) provided by the teacher would be more reliable and better capture the distribution of classes. That is, a more robust teacher provides constructive knowledge and supervision to the student. Thus, the intuitive approach for learning a more accurate student is to employ a bigger and more robust teacher. However, based on the experimental results in \cite{cho2019efficacy}, it is found out that a bigger and more robust model does not always make a better teacher. As the teacher's capacity grows, the student's accuracy rises to some extent, and then begins to drop. We summarize two crucial reasons behind the lack of theoretical support for KD, based on \cite{cho2019efficacy, phuong2019distillation}.
\begin{itemize}
    \item The student is able to follow the teacher, but it cannot absorb useful knowledge from the teacher. This indicates that there is a mismatch between the KD losses and accuracy evaluation methods. As pointed in \cite{phuong2019distillation}, the optimization method used could have a large impact on the distillation risk. Thus, optimization methods might be crucial for significant KD to the student. 
    \item Another reason comes from when the student is unable to follow the teacher due to the large model capacity between the teacher and the student. It is stated in \cite{heo2019comprehensive, hinton2015distilling} that the S-T similarity is highly related to how well the student can mimic the teacher. If the student is similar to the teacher, it will produce outputs similar to the teacher.
\end{itemize}

Intermediate feature representations are also effective knowledge that can be used to learn the student \cite{romero2014fitnets, kim2018paraphrasing}. The common approach for feature-based distillation is to transfer the features into a type of representation that the student can easily learn. In such a case, are bigger models are better teachers? As pointed in \cite{romero2014fitnets}, feature-based distillation is better than the distillation of soft labels, and deeper students perform better than shallower ones. In addition, the performance of the student increases upon increasing the number of layers (feature representations) \cite{kim2018paraphrasing}. However, when the student is fixed, a bigger teacher does not always teach a better student. When the similarity between the teacher and student is relatively high, the student tends to achieve plausible results.

\subsection{Is a pretrained teacher important?}
While most works focus on learning a smaller student based on the pretrained teacher, the distillation is not always efficient and effective. When the model capacity between the teacher and the student is large, it is hard for the student to follow the teacher, thus inducing the difficulty of optimization. Is a pretrained teacher important for learning a compact student with plausible performance? \cite{zhang2018deep, lan2018knowledge} propose learning from student peers, each of which has the same model complicity. The greatest advantage of this distillation approach is efficiency, since the pretraining of a high capacity teacher is exempted. Instead of teaching, the student peers learn to cooperate with each other to obtain an optimal learning solution. Surprisingly, learning without the teacher even enables improving the performance. The question of why learning without the teacher is better has been studied in \cite{tarvainen2017mean}. Their results indicate that the compact student may have a less chance of overfitting. Moreover, \cite{cho2019efficacy} suggests that early stopping of training on ImageNet \cite{deng2009imagenet} achieves better performance. The ensemble of students pool their collective predictions, thus helping to converge at a more robust minima as pointed in \cite{zhang2018deep}.

\subsection{Is born-again self-distillation better?}
Born-again network \cite{furlanello2018born} is the initial self-distillation method in which the student is trained sequentially, and the later step is supervised by the earlier generation. At the end of the procedure, all the student generations are assembled together to get an additional gain. So is self-distillation in the \textit{generations} better? \cite{cho2019efficacy} finds that network architecture heavily determines the success of KD in generations. Although the ensemble of the student models from all the generations outperforms a single model trained from scratch, the ensemble does not outperform an ensemble of an equal number of models trained from scratch. 

Instead, recent works \cite{zhang2019your, xu2019data, mobahi2020self} shift the focus from sequential self-distillation (multiple stages) to the one-stage (online) manner. The student distills knowledge to itself without resorting to the teacher and heavy computation. These methods show more efficiency, less computation costs, and higher accuracy. The reason for such better performance has been pointed out in \cite{zhang2019your, mobahi2020self}. They have figured out that that online self-distillation can help student models converge to flat minima. Moreover, self-distillation prevents student models from the `vanishing gradient’ problem. Lastly, self-distillation helps to extract more discriminative features. In summary, online self-distillation shows significant advantages than sequential distillation methods and is more generalizable. 

\subsection{Single teacher vs multiple teachers}
It is noticeable that recent distillation methods turn to exploit the potential of learning from multiple teachers. Is learning from multiple teachers really better than learning from a single teacher? To answer this question, \cite{you2017learning} intuitively identified that the student can fuse different predictions from multiple teachers to establish its own comprehensive understanding of the knowledge. The intuition behind this is that by unifying the knowledge from the ensemble of teachers, the relative similarity relationship among teachers is maintained. This provides a more integrated dark knowledge for the student. Similar to mutual learning \cite{zhang2018deep, lan2018knowledge}, the ensemble of teachers collects the individual predictions (knowledge) together, thus converging at minima that are more robust. Lastly, learning from multiple teachers relieves training difficulties such as vanishing gradient problems.

\subsection{Is data-free distillation effective enough?}
In the absence of training data, some novel methods \cite{chen2019data, ye2020datafree, lopes2017data, yoo2019knowledge} have been proposed to achieve plausible results. A theoretical explanation for why such methods are robust enough for learning a portable student has yet to be proposed. These methods are focused only on classification, and the generalization capability of such methods is still low. Most works employ generators to generate the `latent' images from noise via adversarial learning \cite{goodfellow2014explaining, wang2020deceiving}, but such methods are relatively hard to train and computationally expensive. 

\subsection{Logits vs features}
The knowledge defined in existing KD methods is from three aspects: logits, feature maps (intermediate layers), and both. However, it is still unclear which one of these represents better knowledge. While works such as \cite{romero2014fitnets, kim2018paraphrasing, huang2017like, heo2019comprehensive, tung2019similarity} focus on better interpretation of feature representations and claim that features might contain richer information; some other works \cite{hinton2015distilling, zhang2018deep, nayak2019zero, wen2019preparing} mention that softened labels (logits) could represent each sample by a class distribution, and a student can easily learn the intra-class variations. However, it is noticeable that KD via logits has obvious drawbacks. First, its effectiveness is limited to the softmax loss function, and it relies on the number of classes (cannot be applied to low-level vision tasks). Secondly, when the capacity between the teacher and the student is big, it is hard for the student to follow the teacher's class probabilities \cite{cho2019efficacy}. Moreover, as studied in \cite{tung2019similarity}, semantically similar inputs tend to elicit similar activation patterns in teacher networks, indicating that the similarity-preserving knowledge from intermediate features express not only the representation space, but also the activations of object category (similar to class distributions). Thus, we can clearly see that features provide more affluent knowledge than logits, and generalize better to the problems without class labels. 

\subsection{Interpretability of KD} In Sec.~\ref{theoretical_analysis}, we provided a theoretical analysis of KD based on the information maximization theory.  It is commonly acknowledged that the teacher model's dark knowledge provides the privileged information on similarity of class categories to improve the learning of students \cite{hinton2015distilling, bucilua2006model}. However, why KD works is also an important question.  There are some methods that explore the principles of KD from the view of label smoothing \cite{yuan2019revisit}, visual concepts \cite{cheng2020explaining}, category similarity \cite{hinton2015distilling}, etc. Specifically, \cite{yuan2019revisit} found that KD is a learned label smoothing regularization (LSR), and LSR is an ad-hoc KD. Even a poorly-trained teacher can improve the student's performance, and the weak student could improve the teacher. However, the findings in \cite{yuan2019revisit} only focus on classification-related tasks,  and these intriguing results do not apply to the tasks without labels \cite{li2020gan, chen2020distilling}.
In contrast, \cite{cheng2020explaining} claims that KD makes DNN learn more task-related visual concepts and discard task-irreverent concepts to learn discriminative features. From a general perspective, the quantification of visual concepts in \cite{cheng2020explaining} provides a more intuitive interpretation for the success of KD. However, there exists a strong need that more intensive research needs to be done in this direction.  

\subsection{Network architecture vs effectiveness of KD.} It has been demonstrated that distillation position  has a significant impact on the effectiveness of KD \cite{cho2019efficacy, heo2019comprehensive}. Most methods demonstrate this by deploying the same network for both teacher and student. However, many fail to transfer across very different teacher and student architectures. Recently, \cite{tian2019contrastive} found that \cite{yim2017gift, ba2014deep, romero2014fitnets} perform poorly even on very similar student and teacher architectures. \cite{yuan2019revisit} also reported an intriguing finding that a poorly trained teacher also can improve the student's performance. It is thus quite imperative to excavate how network architecture affects the effectiveness of KD and why KD fails to work when the network architectures of student and teacher are different.  

\section{New outlooks and perspectives}
In this section, we provide some ideas and discuss future directions of knowledge distillation. We take the latest deep learning methods (\eg, neural architecture search (NAS), graph neural network (GNN)), novel non-Euclidean distances (\eg, hypersphere), better feature representation approaches, and potential vision applications, such as 360$^\circ$ vision \cite{lee2019spherephd} and event-based vision \cite{wang2020eventsr} into account. 
\subsection{Potential of NAS}
In recent years, NAS has become a popular topic in deep learning.
NAS has the potential of automating the design of neural networks. Therefore, it can be efficient for searching more compact student models. In such a way, 
NAS can be incorporated with KD for model compression. 
This has been recently demonstrated for GAN compression \cite{li2020gan,bashivan2019teacher}. It is shown to be effective for finding efficient student model from the teacher with lower computation costs and fewer parameters. It turns out that NAS improves the compression ratio and accelerates the KD process. A similar approach is taken by \cite{ashok2017n2n} learning to remove layers of teacher network based on reinforcement learning (RL). Thus, we propose that NAS with RL can be a good direction of KD for model compression. This might significantly relieve the complexity and enhance the learning efficiency of existing methods, in which the student is manually designed based on the teacher. 

\subsection{Potential of GNN}
Although GNN has brought progress in KD learning under the S-T framework, some challenges remain. This is because most methods rely on finding structured data on which graph-based algorithms can be applied. \cite{liu2019knowledge} considers the instance features and instance relationships as instance graphs, and \cite{ma2019graph} builds an input graph representation for multi-task knowledge distillation. However, in knowledge distillation, there exists \textit{non-structural} knowledge in addition to the structural knowledge (\eg, training data, logits, intermediate features, and outputs of teacher), and it is necessary to construct a flexible knowledge graph to tackle the \textit{non-structural} distillation process. 

\subsection{Non-Euclidean distillation measure}
Existing KD losses are mostly dependent on Euclidean loss (\eg, $l_1$), and have their own limitations. \cite{derezinski2014limits} has shown that algorithms that regularize with Euclidean distance, (\eg MSE loss) are easily confused by random features. The difficulty arises when the model capacity between the teacher and the student is large. Besides, $l_2$ regularization does not penalize small weights enough. Inspired by a recent work \cite{park2019sphere} on GAN training, we propose that it is useful to exploit the information of higher-order statistics of data in non-Euclidean spaces (\eg, hypersphere). This is because geometric constraints induced by the non-Euclidean distance might make training more stable, thus improving the efficiency of KD.  

\subsection{Better feature representations}
Existing methods that focus on KD with multiple teachers show potential for handling cross-domain problems or other problems where the ground truth is not available. However, the ensemble of feature representations \cite{park2019feed, chung2020featuremaplevel, hou2017dualnet} is still challenging in some aspects. One critical challenge is fusing the feature representations and balancing each of them with robust gating mechanisms. Manually assigning weights to each component may hurt the diversity and flexibility of individual feature representations, thus impairing the effectiveness of ensemble knowledge. One possible solution is attention gates, as demonstrated in some detection tasks \cite{li2019attention, schlemper2019attention}. The aim of this approach is to highlight the important feature dimensions and prune feature responses to preserve only the activations relevant to the specific task. Another approach is inspired by the gating mechanism used in long short-term memory(LSTM) \cite{hochreiter1997long,zhao2016regional}. That is, this gate unit in KD is elaborately designed to remember features across different image regions and to control the pass of each region feature as a whole by their contribution to the task (\eg, classification) with the weight of importance.  

\subsection{A more constructive theoretical analysis}
While KD shows impressive performance improvements in many tasks, the intuition behind it is still unclear. Recently, \cite{cho2019efficacy} explained conventional KD \cite{hinton2015distilling} using linear models, and \cite{hegde2019variational, Ahn_2019_CVPR, tian2019contrastive} focus on explaining feature-based KD. Mobahi \etal
\cite{mobahi2020self} provides theoretical analysis for self-distillation. However, the mechanism behind data-free KD and KD from multiple teachers is still unknown. Therefore, further theoretical studies on explaining the principles of these methods should be undertaken.

\subsection{Potentials for special vision problems}
While existing KD techniques are mostly developed based on vision problems (\eg, classification), they rarely exploit some special vision fields such as 360$^\circ$ vision \cite{lee2019spherephd} and event-based vision \cite{wang2020eventsr, wang2019event, mostafavilearning}. The biggest challenge for both these vision fields is the lack of labeled data, and learning in these needs a special change of inputs for neural networks. Thus, the potential of KD, particularly cross-modal KD, for these two fields is promising. By distilling knowledge from the teacher trained with RGB images or frames to the student network specialized in learning to predict 360$^\circ$ images or stacked event images, it not only handles the problem of lack of data, but also achieves desirable results in the prediction tasks. 

\subsection{Integration of vision, speech and NLP.} 
As a potential, it is promising to apply KD to the integrated learning problems of vision, speech, and NLP. Although recent attempts of cross-modal KD \cite{albanie2018emotion, zhao2018through, thoker2019cross, owens2016ambient} focus on transferring the  knowledge from one modality (\eg, video) to the other (\eg, sound) on the end tasks, it is still challenging to learn the end tasks for the integration of the three modalities. The major challenge may come from collecting paired data of three modalities; however, it is possible to apply GAN or representation learning methods to unsupervised cross-modal KD for learning effective end tasks. 

\section{Conclusion}
This review of KD and S-T learning has covered major technical details and applications for visual intelligence. We provide a formal definition of the problem, and introduce the taxonomy methods for existing KD approaches. Drawing connections among these approaches, we identify a new active area of research that is likely to create new methods that take advantage of the strengths of each paradigm. Each taxonomy of the KD methods shows the current technical status regarding its advantages and disadvantages. Based on explicit analyses, we then discuss methods to overcome the challenges, and break the bottlenecks by exploiting new deep learning methods, new KD losses, and new vision application fields.   

\ifCLASSOPTIONcompsoc
  \section*{Acknowledgments}
\else
  \section*{Acknowledgment}
\fi

This work was supported by the National Research
Foundation of Korea(NRF) grant funded by the Korea government(MSIT) (NRF-2018R1A2B3008640) and Institute of Information \& Communications Technology Planning \& Evaluation(IITP) grant funded by Korea government(MSIT) (No.2020-0-00440, Development of Artificial Intelligence Technology that Continuously Improves Itself as the Situation Changes in the Real World).

\ifCLASSOPTIONcaptionsoff
  \newpage
\fi

{\small 
\bibliographystyle{IEEEtran}
\bibliography{ref}
}


\vspace{-40pt}
\begin{IEEEbiography}[{\includegraphics[width=1in,height=1.25in,clip,keepaspectratio]{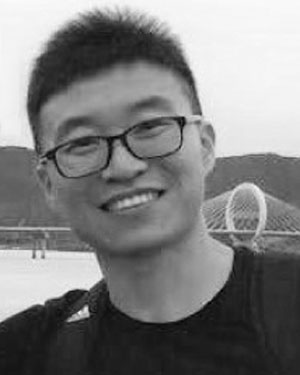}}] {Lin Wang} is a Ph.D. student in Visual Intelligence Lab., Dept. of Mechanical Engineering, Korea Advanced Institute of Science and Technology (KAIST). His research interests include deep learning (especially adversarial learning, transfer learning, semi-/self-supervised learning), event camera-based vision, low-level vision (image super-solution and deblurring, etc.) and computer vision for VR/AR.
\end{IEEEbiography}

\vspace{-40pt}
\begin{IEEEbiography}[{\includegraphics[width=1in,height=1.25in,clip,keepaspectratio]{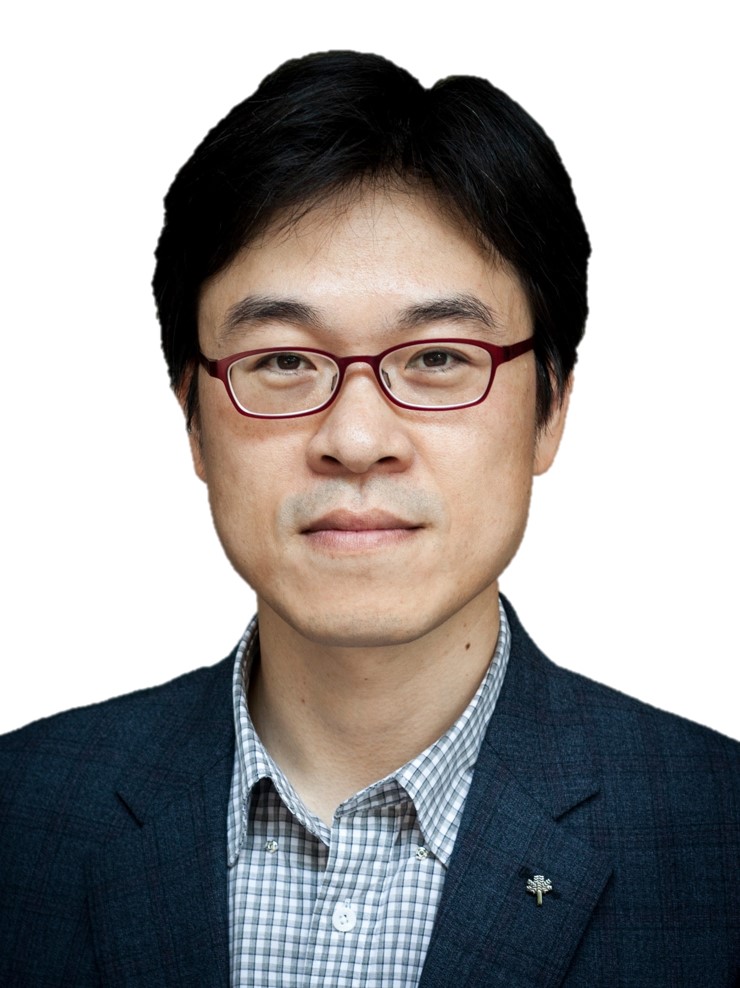}}]{Kuk-Jin Yoon}
received the B.S., M.S., and Ph.D. degrees in electrical engineering and computer science from the Korea Advanced Institute of Science and Technology in 1998, 2000, and 2006, respectively.  He is now an Associate Professor at the Department of Mechanical Engineering, Korea Advanced Institute of Science and Technology (KAIST), South Korea, leading the Visual Intelligence Laboratory. Before joining KAIST, he was a Post-Doctoral Fellow in the PERCEPTION Team, INRIA, Grenoble, France, from 2006 to 2008, and was an Assistant/Associate Professor at the School of Electrical Engineering and Computer Science, Gwangju Institute of Science and Technology, South Korea, from 2008 to 2018. His research interests include various topics in computer vision such as multi-view stereo, visual object tracking, SLAM and structure-from-motion, 360 camera and event-camera-base vision, sensor fusion.
\end{IEEEbiography}
\end{document}